\documentclass[12pt,a4paper,twoside]{scrreprt}
\usepackage[inner=2.5cm,outer=3cm,top=2cm,bottom=1.5cm,includeheadfoot]{geometry}

\usepackage{float}
\usepackage{subfigure}
\usepackage[T1]{fontenc}

\usepackage{listings}
\usepackage{fancyhdr}
\usepackage{fancyvrb}
\usepackage{color}
\usepackage{float}
\usepackage[english]{babel}
\usepackage[T1]{fontenc}
\usepackage[latin1]{inputenc}

\usepackage{graphicx}
\usepackage{wrapfig}
\usepackage{amsfonts} 
\usepackage{amssymb}
\usepackage{amsmath}
\usepackage{amsthm}
\usepackage{array}
\usepackage{calrsfs}
\usepackage[showall]{psfrag}
\usepackage{url}
\numberwithin{equation}{section}
\newcommand{\R}{{\mathbb R}} 
\newcommand{\Q}{{\mathbb Q}} 
 
\newcommand{\N}{{\mathbb N}}
\newcommand{\Z}{{\mathbb Z}}

\newcommand{\EndProof}{{\begin{flushright}\vspace{-2mm}$\Box$\end{flushright}}}
  \newtheorem{theo}{Theorem}[section]
  \newtheorem{sen}[theo]{Theorem}
  
  \newtheorem{lem}[theo]{Lemma}
  \newtheorem{defn}[theo]{Definition}
  
  \newtheorem{dat}[theo]{Definition and Theorem}

\newcommand{\be}{\begin{equation}}
\newcommand{\ee}{\end{equation}}

\newcommand{\e}{\,\mathrm e}
\newcommand{\oomega}{\overline{\omega}}

\usepackage[bf]{caption}

\begin{document}

\begin{titlepage}
\begin{center}
\large{Ludwig-Maximilians-Universit\"{a}t M\"{u}nchen\\
Mathematisches Institut}
\vspace{2.5cm}
\begin{center} \LARGE{\scshape{Diplomarbeit}\\
\vspace{0.7cm}
\textbf{Approximation of probability density functions on the Euclidean group parametrized by dual quaternions}}\\
\vspace{1.7cm}
\begin{center}
\includegraphics[]{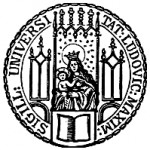}
\end{center}
\vspace{1cm}
\end{center}
\large{Autorin: Muriel Lang\\
Themenstellung und Betreuung:\\ Dr. Vitali Wachtel, Dr. Wendelin Feiten}\\
\vspace{0.8cm}
Eingereicht am 25.01.2011
\end{center}
\end{titlepage}
\thispagestyle{empty}\newpage


\vspace*{7cm}
\begin{abstract}
\begin{center}
\textbf{Abstract}
\end{center}
Perception is fundamental to many robot application areas especially in service robotics. Our aim is to perceive and model an unprepared kitchen scenario with many objects. We start with the perception of a single target object. The modeling relies especially on fusing and merging of weak information from the sensors of the robot in order to localize objects. This requires the representation of various probability distributions of pose in $S_3\times\R^3$ as orientation and position have to be localized. In this thesis I present a framework for probabilistic modeling of poses in $S_3\times\R^3$ that represents a large class of probability distributions and provides among others the operations of the fusion and the merge of estimates. Further it offers the propagation of uncertain information data. I work out why we choose to represent the orientation part of a pose by a unit quaternion. The translation part is described either by a 3-dimensional vector or by a purely imaginary quaternion. This depends on whether we define the probability density function or whether we want to represent a transformation which consists of a rotation and a translation by a dual quaternion. 
A basic probability density function over the poses is defined by a tangent point on the hypersphere and a 6-dimensional Gaussian distribution. The hypersphere is embedded to the $\R^4$ which is representing a unit quaternions whereas the Gaussian is defined over the product of the tangent space of the sphere and of the space of translations. The projection of this Gaussian to the hypersphere induces a distribution over poses in $S_3\times\R^3$.
The set of mixtures of projected Gaussians can approximate the probability density functions that arise in our application. Moreover it is closed under the operations introduced in this framework and allows for an efficient implementation.
\setcounter{page}{1}
\end{abstract}
\newpage
\thispagestyle{empty}
\vspace*{5cm}
\Large\textbf{Acknowledgments}\\
\vspace*{2cm}
\normalsize 

Especially I want to thank the participants of the "Diplomandenseminar" for the helpful remarks they gave to me throughout the whole period I was at Siemens and to Thilo Grundmann for the pictures.\\
Further thanks to Dr. Wendelin Feiten, Dr. Gisbert Lawitzky and Dr. Vitali Wachtel for the good mentoring and to Darren Murph for allowing me to use the pictures of the web page www.Engadget.com.\\

My family, Judith Gampe, Thomas Haugg, Florian Sch\"{a}fer, Simeon Schmied and all the others that were so patient with me I want to thank for their support.
\newpage
\tableofcontents
\newpage


\chapter{Introduction}\label{intro}

\section{Motivation}

\begin{figure}[bht]
\begin{center}
 \includegraphics[width=1\textwidth]{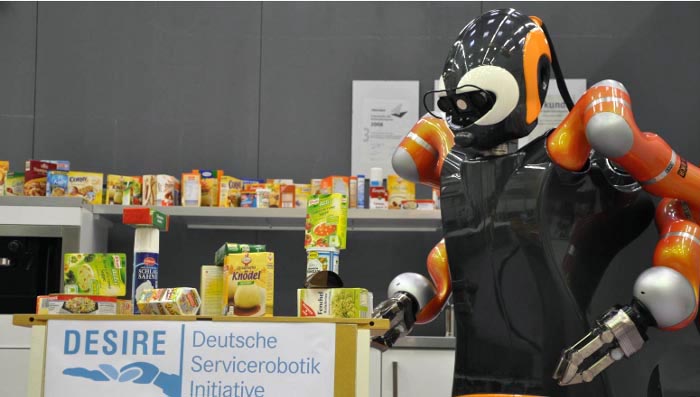}
 \end{center}
\caption[DESIRE robot observing]{The figure shows the robot of the DESIRE project (which stands for 'Deutsche Servicerobotic Initiative') observing a kitchen scenario.}
 \label{desire}
\end{figure}

Imagine a robot that has the task to get a specified object from a table. Several objects are arranged on the table like in an arbitrary kitchen scenario. This kind of task is fundamental for many applications of service robotics like mobile manipulation.\\
The robot knows the pose of the table and notices there is something on top of it. Height and position of the table usually are given to the robot to receive better results in experiments. For the algorithm it is not necessary to know the pose of the table. It could also be estimated. That something is placed on top of it, the robot can see through its stereo camera system. Often a 3D sensor like a laser scanner or ultra sonic sound sensor also is part of the robot's sensor system.\\
Robots are not able to identify unknown objects through the process established in our framework. They just can recognize known objects. Therefore a database of 3D models of objects is implemented to the robot. The robot makes several independent attempts to determine orientation and position of the target object exact enough to be able to grasp it. These localization attempts might be with different methods, from different points of view or under different conditions like brightness and shadows for instance. Thus we get plenty of more or less suitable information data which the robot needs to handle in an appropriate way. The robot repeats making independent attempts to localize the object until the resulting pose is determined \textit{well enough} i.e. the uncertainty is below a given threshold and the robot can pick up the object with a failure probability below a resulting value $\varepsilon>0$.\\

In this context rotation and orientation are often used synonymously as well as translation and position. This comes from the fact that the orientation of any object results from the rotation that moves it to this alignment and likewise the position results from the translation. Together orientation and position describe the \textbf{pose} of an object as a pose is the result of any transformation consisting of rotation and translation.\\
Furthermore I won't handle object classes in this work. I suppose that the objects are part of the robot's database and that it recognizes them.\\

\section{Problem Statement}

\begin{figure}[bht]
\begin{center}
 \includegraphics[width=1\textwidth]{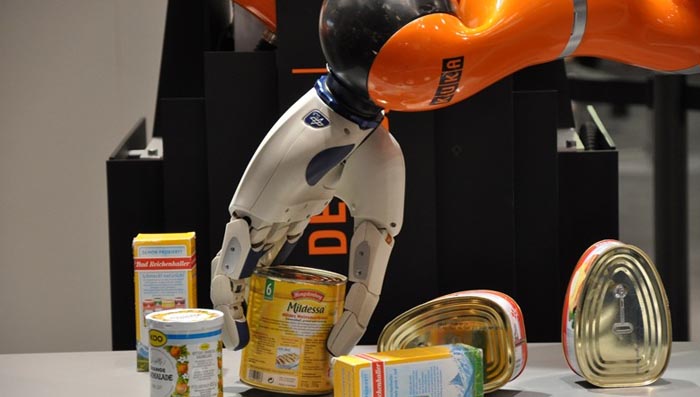}
 \end{center}
\caption[DESIRE robot grasping]{The picture shows the robot of the DESIRE project succeeding to grasp the target object after having identified it on the CeBIT in March 2009.}
 \label{desire2}
\end{figure}

For the whole present work I assume that the data association problem is solved. This means that we know which object the data describes we receive from the individual localization attempts. I claim this to reduce the uncertainty and thus to simplify the situation we have to deal with. We work with a single target object and its observation data.\\
After having reduced the complexity of the task that far we have to estimate the state of the target object. As in a kitchen scenario like illustrated in figure \ref{desire}, three dimensions for each, the rotation and translation can occur, we have to deal with a 6-dimensional space, namely the special Euclidean group, which will be introduced in \ref{pose}. Such a state of the target object in the special Euclidean group is called object pose and thus I will refer to pose estimation instead of state estimation. \\
To be able to estimate a pose at first the representation of the pose and parametrization of rotation and translation have to be defined. In section \ref{poseRep} I introduce several candidates for the pose representation and then select the most suitable one for our topic. After analyzing the problem from an algebraic point of view some probability theory has to be introduced. To estimate the pose of a target object, its probability distribution has to be known. Section \ref{distribution} is concerned with some distribution functions on the special Euclidean group which can either have a parameterized density or consist of a particle set. Finally I justify in section \ref{MoPGs} my decision to choose mixtures of projected Gaussians as preferable distribution function to estimate the pose of the target object and introduce the algorithms for basic operations. One of them is the fusing of two mixtures i.e. deriving a common estimate from two independent estimates, another one is to merge components in a mixture. \\
In chapter \ref{Approx} which is the body of this work I handle approximations of mixtures of projected Gaussians. In detail I give an upper bound for the approximation error that occurs on dropping summands of a mixture in section \ref{omit}. Further I explain the difficulties that arise on applying the operations to fuse and to merge to mixtures of projected Gaussians. If different information data shall be applied at the same time a weighting factor has to be introduced to evaluate the compatibility of the single mixture elements. Section \ref{fuse} is concerned with that whereas \ref{merge} is concerned with the merge of a mixture of projected Gaussians. Merging a mixture means reducing the number of summands through iteratively merging the two least dissimilar summands of the mixture. Therefore a kind of dissimilarity measure basing on an appropriate distance measure has to be defined.\\
I pull the analysis of distance and convergence measures out to the beginning of chapter \ref{Approx} as I need it for severals aspects of approximations. It is desired to keep the exactness that the robot is able to grasp the target object with a high probability despite any approximation of the mixture. To study the accuracy of the approximation of a pose estimation I examine the failure probability that occurs on trying to grasp an object. Thus the introduction of a grasp criterion given in section \ref{gripper} is required. The criterion also base on distance measures between the pose of the gripper and the estimated pose of the target object.\\
We require from the approximation of a mixture of projected Gaussians to optimize the necessary computational effort with minimized loss of accuracy of the pose estimation. It is known that the mixture describing the pose estimation after evaluating more and more data from the localization attempts converges to the mixture describing the true object's pose. I study further convergence criteria that hold for approximations of mixtures of projected Gaussians and examine the properties that are passed on from the original mixture to the approximated one in section \ref{prop}.\\
The last section \ref{AA} of this chapter is concerned with approximation algorithms. Given a mixture of projected Gaussian that describes the probability distribution of the pose of a target object, but might be complicated to calculate or might contain parameters which are unknown to us. Then a second mixture using a reduced number of base elements should be fitted to the first mixture. I explain the expectation maximization algorithm \ref{EM} and the Monte Carlo algorithm \ref{MC} which are two different ways how this fit can be done.\\
The practical application of the results from chapter \ref{Approx} is introduced in chapter \ref{Sensor}. As already mentioned the robot makes several localization attempts to determine the pose of the target object. Then it estimates the uncertainty of each of these tries and fuses and merges the weighted measurements according to an evaluation algorithm. With this evaluation algorithm the uncertainty shall be reduced until the distribution of the estimated pose is peaked enough so that the grasp criterion is fulfilled with failure probability smaller that a given threshold. In chapter \ref{Sensor} I explain how the robot draws conclusions from single 3D points it detects on the surface of the object, about the pose of the target object itself where the 3D points are so-called point features. The only features I treat in the framework are SIFT features what stands for scale invariant feature transform, even though the framework also would allow the handling of other features like edge detection for instance. In section \ref{code} I give a documentation of the code I wrote to program a framework for the precise estimation of an object's pose and in section \ref{example} I demonstrate how the framework works. In the outlook \ref{outlook} a picture \ref{desire2} shows the result that can be achieved by a similar approach of pose estimation to the one in this work even though I also have to remark that some aspects are left over to be treated in future work.

\section{Related Work}

The representation of rigid motions and especially of orientation in three dimensions is a central issue in various disciplines of arts, science and engineering. Rotation matrix, Euler angles, Rodrigues vector and unit quaternions are the most popular representations of a rotation in three dimensions. Rotation matrices have many parameters, Euler angles are not invariant under transforms and have singularities and Rodrigues vectors do not allow for an easy composition algorithm. Stuelpnagel \cite{stuelpnagel} points out that unit quaternions are a suitable representation of rotations on the hypersphere $S_3$ with few parameters, but does not provide probability distributions. Choe \cite{choe} represents the probability distribution of rotations via a projected Gaussian on a tangent space. He only deals with concentrated distributions and does not take translations into account. Goddard and Abidi \cite{goddard, abidi} use dual quaternions for motion tracking. In their observations the correlation between rotation and translation is captured also. The probability distribution over the parameters of the state model is a unimodal normal distribution. If the initial estimate is sufficiently certain and if the information that shall be fused to the estimate is sufficiently well focused this is an appropriate model. As can be seen in \cite{kavan} from Kavan et al. dual quaternions provide a closed form for the composition of rigid motions, similar to the transform matrix in homogeneous coordinates. Antone \cite{antone} suggests to use the Bingham distribution in order to represent weak information even though he does not give a practical algorithm for fusion of information or propagation of uncertain information. By now it is known that propagated uncertain information only can be approximated by Bingham distributions. Further Love \cite{love} states that the renormalization of the Bingham distribution is computationally expensive. Glover \cite{Jared} also works with a mixture of Bingham distributions and recommends to create a precomputed lookup table of approximations of the normalizing constant using standard floating point arithmetic. Mardia et al. \cite{mardia} use a mixture of bivariate von Mises distributions. They fit the mixture model to a data set using the expectation maximization algorithm because this allows for modeling widely spread distributions. Translations are not treated by them. To propagate the covariance matrix of a random variable through a nonlinear function, the Jacobian matrix is used in general. Kraft et al. \cite{kraft} use therefore an unscented Kalman Filter \cite{KalmanFilter}. This technique would have to be extended to the mixture distributions.\\

\chapter{Pose Estimation}

In robotic perception the six dimensional pose of the object of interest has to be estimated. The input information is weak as it comes from imperfect sensors. These uncertainties arise from not exactly tuned 3D sensors or cameras and from the joints in the robot's body. Moreover the three dimensional models which the robot uses to recognize the objects are not perfect. Altogether one has to deal with uncertain information which is modeled by widely spread probability density functions ($pdfs$) to estimate position and orientation in the six dimensional space. In order to be able to represent and process this weak information, the density functions are formulated on the bases of suitable parametric representations of the pose.\\

\section{Pose Representation}\label{poseRep}

\subsection{Representation of Orientation}

The more critical part in the pose representation is the rotation. There are several requirements concerning the parametrization of the rotation which we wish to be fulfilled the best. 
\begin{itemize}
\item{For each position there should be just one representation to avoid wrong choice of representation.}
\item{To minimize the computational effort we desire the rotation to be represented by few parameters. If a representation uses more than the  minimal number of three parameters, some additional condition needs to be satisfied to reduce the number of independent parameters to three. After calculations or estimations of parameters these conditions may have to be re-established in the \textit{intuitively best possible way} although we do not formally define what this is. We call it 'renormalization' and we desire this step to be easy to perform.}
\item{There should be an easy way to derive the parameters of the composed rotation from the parameters of two input rotations of the composition. The composability of rigid motions ins needed for instance if sensor data taken from two different sensor system poses shall be fused in a common coordinate system.}
\item{We wish the rotation to be a differentiable function of the parameters to assure smoothness. At least it should be continuous.}
\item{Finally a desired characteristic of the parameterization is to be area and distance preserving under rotation and translation. This is important when we deal with probability density functions over rigid motions.}
\end{itemize}
A short overview about the parameterizations of orientations is given in \cite{OrientationRepresentation}.


\subsubsection{Rotation Matrices}

These are the probably most wide spread representations, especially in the homogeneous coordinate formulation.\\
A rotation matrix $R$ is constrained to be orthogonal $R^\top\cdot R = R\cdot R^\top = I$ and the restriction $\det(R) = 1$ is to avoid it to be a reflection.\\
The set of all rotation matrices forms a group, known as the rotation group or the special orthogonal group $SO(n)$. It is a subset of the orthogonal group $O(n)$, which includes reflections and consists of all orthogonal matrices with determinant $1$ or $-1$.\\

In $\R^3$ the matrix for a rotation by an angle $\theta$ about the axis $v$ where $v = (x, y, z)^\top$ a unit vector, is given by:
$$R = 
\begin{bmatrix} 
\cos\theta+x^2 \left(1-\cos\theta\right) & x y\left(1-\cos\theta\right) - z\sin\theta & x z \left(1-\cos\theta\right) + y\sin\theta\\
y x\left(1-\cos\theta\right)+ z\sin\theta & \cos\theta + y^2\left(1-\cos\theta\right) & y z \left(1-\cos\theta\right) - x\sin\theta\\
z x\left(1-\cos\theta\right)- y\sin\theta & z y\left(1-\cos\theta\right) + x\sin\theta & \cos\theta + z^2\left(1-\cos\theta\right) 
\end{bmatrix}$$

\textit{What characteristics do rotation matrices have?}
\begin{itemize}
\item{The representation is unique.}
\item{It is well known and there are wide spread applications.}
\item{It consists of too many parameters. The constraints arise from the need of nine values for the rotation matrix to represent three independent variables of a 3D rotation.}
\item{The renormalization is difficult.}
\item{The rotation matrix is differentiable with resect to its parameters and preserves area and distance.}\\
\end{itemize}

\subsubsection{Euler Angles}

The Euler angles are three angels $\Psi,\, \Theta$ and $\Phi$ which describe rotations around specified axes in $\R^3$ usually. Together they define a transformation between two coordinate systems. 
There are different possibilities to choose the rotation axes. One of the most common ones is the following convention:
\begin{enumerate}
\item Rotate the coordinate system at first about $\Psi$ around the $z$-axis. Then new coordinate axes $x'$ and $y'$ are obtained.
\item Then rotate it about the angel $\Theta$ around the new $x$-axis $x'$. The new coordinate axes which are obtained after the rotation are denoted with $y''$ and $z''$.
\item Finally rotate the system about $\Phi$ around the new $z$-axis $z''$.\\
\end{enumerate}

Thus the whole rotation is obtained by the rotation matrix $R_{zx'z''}$:
$$ R_{zx'z''} = \begin{pmatrix} \cos \Psi & - \sin \Psi & 0 \\ \sin \Psi & \cos \Psi & 0 \\ 0 & 0 & 1 \end{pmatrix} \begin{pmatrix} 1 & 0 & 0 \\ 0 & \cos \Theta & - \sin \Theta \\ 0 & \sin \Theta & \cos \Theta \end{pmatrix} \begin{pmatrix} \cos \Phi & - \sin \Phi & 0 \\ \sin \Phi & \cos \Phi & 0 \\ 0 & 0 & 1 \end{pmatrix}$$
$$= \begin{pmatrix} \cos \Psi \cos \Phi - \sin \Psi \cos \Theta \sin \Phi & - \cos \Psi \sin \Phi - \sin \Psi \cos \Theta \cos \Phi & \sin \Psi \sin \Theta \\ \sin \Psi \cos \Phi + \cos \Psi \cos \Theta \sin \Phi & \cos \Psi \cos \Theta \cos \Phi - \sin \Psi \sin \Phi & - \cos \Psi \sin \Theta \\ \sin \Theta \sin \Phi & \sin \Theta \cos \Phi & \cos \Theta \end{pmatrix}$$\\

Euler angles and translation vector are natural for robotics, for instance where the angles are the motor positions like in the wrist, and common for representation of small angle ranges of the $SO(3)$ like for cars and ships. But there are twelve different choices for the rotation axes which is much room for confusion.\\

\textit{Characteristics of the Euler angles:}
\begin{itemize}
\item{Minimal number of parameters, namely three.}
\item{The composition is not straight forward and there is no easy algorithm for finding the Euler angles of a composition given the Euler angles of two individual rotations.}
\item{The parameterization is periodic with $2\pi$ and dependent on the choice of axes, moreover the Euler angles are not invariant under transformations and have singularities.}
\item{Gimbal Lock (loss of one degree of freedom in a 3D space that occurs when the axes of two of the three gimbals are driven into a parallel configuration)}\\
\end{itemize}

\subsubsection{Rodrigues Vector}


Rodrigues found the quaternions three years before Hamilton and derived the Rodrigues rotation formula from them.
This rotation formula describes an algorithm to rotate a vector in the 3-dimensional Euclidean space, given an axis $\hat{\mathbf{e}}$ and angle $\theta$ of rotation. The paper he wrote appeared in \textit{Annales de Gergonne} in 1840 \cite{Rodrigues}.\\

\begin{defn}
Let $v = (v_1,v_2,v_3)^\top$ be a vector in $\mathbb{R}^3$ and $\hat{\mathbf{e}} = (e_x\ e_y\ e_z)^\top$ the 3D unit vector describing an axis of rotation about which we want to rotate $v$ by the angle $\theta$.\\
The Rodrigues formula is defined as: 
$$v_\mathrm{rot} = v \cos\theta + (\hat{\mathbf{e}} \times v)\sin\theta + \hat{\mathbf{e}} (\hat{\mathbf{e}} \cdot v) (1 - \cos\theta)$$ 
And in matrix notation: 
$$v_{\mathrm{rot}} = v \cos\theta + \begin{pmatrix}0&-e_z&e_y\\e_z&0&-e_x\\-e_y&e_x&0\\ \end{pmatrix} \begin{pmatrix}v_1\\v_2\\v_3\\ \end{pmatrix}\sin\theta + \hat{\mathbf{e}} \cdot \hat{\mathbf{e}}^\top \cdot v (1 - \cos\theta)$$
\end{defn}
In the definition $\hat{\mathbf{e}} \times v$ means the cross product of the two vectors $\hat{\mathbf{e}}$ and $v$. It is defined as $(|\hat{\mathbf{e}}|\cdot|v|\sin\alpha)\cdot n$ where $0\leq\alpha\leq 180^\circ$ is the smallest angle between the vectors and $n$ is the normal vector of the plane containing $\hat{\mathbf{e}}$ and $v$.\\

Rodrigues parameters  can be expressed in terms of Euler axis and angle as $u = \hat{\mathbf{e}}\cdot \tan(\theta/2)$, a non-normalized 3D vector. The direction of $u$ specifies the axis, and its magnitude is $\tan(\theta/2)$. Thus the angle $\theta$ of rotation is given by $\|u\| = \tan(\theta/2)$.\\
Since $\hat{\mathbf{e}} \in S_3$ and $-\hat{\mathbf{e}} \in S_3$ define the same rotation, each orientation is uniquely determined by a point on the unit hemisphere of $S_3$.\\

\textit{Characteristics of rotation by Rodrigues vectors:}
\begin{itemize}
\item{They have the minimal number of three parameters.}
\item{There is no or at least no easy composition algorithm.}
\item{The parametrization is periodic with $2\pi$ and computations are not efficient.}\\
\end{itemize}


\subsubsection{Unit Quaternions}


Quaternions are a generalization of the complex numbers to the $\mathbb{R}^4$ as can be seen in \cite{Quat}. Instead of one imaginary unit, they have three, $i,\, j$ and $k$. With real coefficients $a,\, b,\, c,\, d$ a quaternion $q$ is defined as 
$$q :=a+i\cdot b + j\cdot c + k\cdot d$$
or $[a,b,c,d]$ in vector notation. 
\begin{defn} The set $\mathbb{H} := \{q =a+i\cdot b + j\cdot c + k\cdot d: a,\, b,\, c,\, d \in \mathbb{R}\}$, named after Hamilton, is the \textbf{skew field of quaternions} with component wise summation and quaternionic multiplication which are defined in the following and the neutral elements:
$$0 = 0+i\cdot0+j\cdot0+k\cdot0 \quad \mathrm{and}\quad 1 = 1+i\cdot0+j\cdot0+k\cdot0$$
Further it has the properties:
$$\begin{array}{lcl}
i\cdot j &=&k\\j\cdot k&=&i\\k\cdot i&=&j\\ i\cdot j\cdot k &=& i^2 = j^2= k^2= -1
\end{array}$$\\
\end{defn}

Component wise summation of quaternions:
$$\begin{array}{lcl}
q_1 + q_2 &=& (a+i\cdot b + j\cdot c + k\cdot d)+ (e+i\cdot f + j\cdot g + k\cdot h)\\
&=& a+e+i \cdot(b+f)+j \cdot (c+g)+k\cdot (d+h)
\end{array}$$\\
Multiplication of a quaternion with a scalar $\lambda \in \mathbb{R}$ is defined as:
$$\begin{array}{lcl}
\lambda\cdot q &=& \lambda \cdot (a + i\cdot b + j\cdot c + k\cdot d)\\
&=& \lambda\cdot a +\lambda\cdot j\cdot b +\lambda\cdot j\cdot c +\lambda\cdot k\cdot d
\end{array}$$\\
Quaternionic multiplication follows from the properties of imaginary units: 
$$\begin{array}{lcl}
q_1 * q_2 &=&(a+i\cdot b + j\cdot c + k\cdot d)* (e+i\cdot f + j\cdot g + k\cdot h)\\
&=& (ae - bf - cg - dh) + i\cdot (af + be + ch - dg) \\
&+& j\cdot (ag - bh + ce + df) + k \cdot (ah + bg - cf + de)
\end{array}$$\\
For the multiplication of quaternions the associative and the distributive law hold, but not the commutative law. Easy calculations show that $\mathbb{H}$ is an associative and not commutative algebra see \cite{DualQuaternions} and 1 is the identity element of $\mathbb{H}$.\\
Like for complex numbers, there is a conjugate quaternion: $\overline{q} := a - i\cdot b - j\cdot c -k\cdot d$ and the norm of a quaternion is the 2-norm in $\mathbb{R}^4$: $\|q\| =\sqrt{a^2+b^2+c^2+d^2}= \sqrt{q * \overline{q}}$.\\
Every non-zero quaternion has an inverse: $q^{-1} = \frac{1}{(||q||)^2}\overline{q}$ and for any two quaternions $q_1$ and $q_2$ we have the formula $\overline{(q_1*q_2)} = \overline{q_2} *\overline{q_1}$.\\

We can restrict the quaternion to be imaginary what means without real part:
\begin{defn}
The set $\mathbb{H}_{Im} := \{q =0+i\cdot b + j\cdot c + k\cdot d:  b,\, c,\, d \in \mathbb{R}\}$ is called the set of imaginary quaternions.\\
\end{defn}

The 3-sphere $S_3  \subset    \mathbb{H}$ in quaternionic calculus is similar to the unit
circle $S_1 \subset      \mathbb{C}$ in complex calculus. In fact: 
$$S_3=\{q \in \mathbb{H}: \|q\| = 1\}$$
$\|q\| = 1$ defines exactly the \textbf{unit quaternions} which obey the following constraint: 
$$a^2+b^2 + c^2 + d^2 = 1$$\\
In terms of the Euler axis $\hat{\mathbf{e}} = [e_x\ e_y\ e_z]^\top$ and angle $\theta$ the elements of the quaternion in vector notion can be expressed as follows:
$$\begin{array}{lcl}
a &=& \cos(\theta/2)\\ b &=&  e_x\sin(\theta/2)\\ c &=& e_y\sin(\theta/2)\\d &=&e_z\sin(\theta/2) \end{array}$$

Thus the rotation can be represented with the quaternion $q = [\cos(\frac{\theta}{2}), \hat{\mathbf{e}}\cdot \sin(\frac{\theta}{2})]$. To rotate a point in $\mathbb{R}^3$ we identify it with the quaternion $p = [0,\, b,\, c,\, d]$. Then the rotation about $q$ can be applied as: $$q*p*\overline{q}$$ 
The composition of rotations $q_1$ and $q_2$ easily is the product of them: 
$$q_1*(q_2*p*\overline{q_2})*\overline{q_1} = (q_1*q_2)*p*(\overline{q_1*q_2})$$
From $q*p*\overline{q} = (-q)*p*\overline{(-q)}$ the antipodal symmetry of the quaternions can be seen.\\

Comparing the performance of rotation with quaternions and matrices gives the following result:\\
Rotating a point, is easier done by matrix operation than with quaternions, but chaining operations like composition of rotations is less costly with quaternions. Matrix multiplication needs 27 operations in opposition to quaternion multiplication that just needs 16. For summation and subtraction 18 operation for matrices and just 12 for quaternions are needed. Thus the computation with quaternions is less expensive.\\

\textit{In summary the characteristics of rotation representation by unit quaternions are:}
\begin{itemize}
\item{Unit quaternions have few parameters, four instead of the minimal number three.}
\item{They are unique except for antipodal symmetry.}
\item{The rotation from one state to another on the great circle gives a smooth movement avoiding unnatural angular moves that occur for Euler angles.}
\item{Computations are highly efficient.}
\item{They are easy to deal with.}\\
\end{itemize}

\subsubsection{Pluecker line coordinates and Complex mapping}
These are rotation representations which I just want to name for completeness. They don't fulfill the desired requirements and thus I won't handle them any further.\\

\subsection{Pose} \label{pose}

Pose representation is equivalent to representation of rigid motion. The group of rigid motions in $\mathbb{R}^n$ is sometimes called \textbf{special Euclidean group} $SE(n)$.\\

\subsubsection{Euclidean Group}

A transformation is said to be rigid if it preserves \textbf{relative distances} what means it is angle and distance preserving.
\begin{itemize}
\item{The composition of rigid transformations is rigid.}
\item{The inverse of a rigid transformations is rigid.}
\end{itemize}
The subgroup of rigid transformations which additionally is orientation preserving, is called the \textbf{group of rigid motions} and just contains rotations and translations.\\

\begin{defn}
$E(n) := \mathbb{R}^n \times O(n)$ is the $n$-dimensional Euclidean Group, where $O(n)$ is the $n$-dimensional orthogonal group. 
\end{defn}
$E(n)$ is the symmetry group of $n$-dimensional Euclidean space and consists of bijective, distance and angle preserving affine transformations.
\begin{defn}
$SE(n) := \mathbb{R}^n \times SO(n)$ is the $n$-dimensional special Euclidean Group, where $SO(n)$ is the $n$-dimensional special orthogonal group and $S_n$ is the $n$-sphere. 
\end{defn}
$SE(n)$ is a subgroup of $E(n)$ which just contains direct isometries. Direct means orientation preserving. $SE(n)$ is also called the subgroup of rigid motions.\\

The 3-dimensional special Euclidean Group $SE(3) = \mathbb{R}^3 \times SO(3)$ thus represents all possible rotations and translations in the three dimensional Euclidean space. \\

\subsubsection{Dual Quaternions} \label{DQ}


\begin{defn}
The ring of the dual quaternions with the dual unit $\epsilon$, which has the property $\epsilon^2 = 0$, is defined as: 
$$\mathbb{H}_D = \{dq\ |\ dq=q_1 + \epsilon\cdot q_2 \,\&\,  q_1, q_2 \in \mathbb{H}\}$$
\end{defn}
This ring can also be written as $\{a_D+i\cdot b_D + j\cdot c_D + k\cdot d_D : a_D,\, b_D,\,c_D,\,d_D\, \, \mathrm{dual\; numbers}\}$ see \cite{DualQuaternions}. The dual numbers $\{a_D, b_D,c_D,d_D\}$ are called components of a dual quaternion then. Just as the quaternions, the dual quaternions have the basis $\{1,i,j,k\}$ of the 4-dimensional linear space over the dual numbers. \\

Summation of dual quaternions is component wise: 
$$\begin{array}{lcl}
 dq_1 + dq_2 &=& (q_{1,1} + \epsilon \cdot q_{1,2}) + (q_{2,1} + \epsilon \cdot q_{2,2})\\
  &=& (q_{1,1} +q_{2,1}) + \epsilon \cdot (q_{1,2} + q_{2,2})\\
\end{array}$$\\
Multiplication of a dual quaternion with a scalar $\lambda \in \mathbb{R}$ is defined as:
$$\begin{array}{lcl}
 \lambda \cdot dq &=& \lambda \cdot (q_1 + \epsilon \cdot q_2)\\
  &=& \lambda \cdot q_1 + \lambda \cdot \epsilon \cdot q_2 \\
\end{array}$$\\
The product of two dual quaternions is defined as:
$$\begin{array}{lcl}
 dq_1 **\, dq_2 &=& (q_{1,1} + \epsilon \cdot q_{1,2}) **\, (q_{2,1} + \epsilon \cdot q_{2,2})\\
  &=& (q_{1,1} *q_{2,1}) + \epsilon \cdot (q_{1,2} * q_{2,1} + q_{1,1} * q_{2,2})\\
\end{array}$$\\
As for quaternions the associative and the distributive law hold, but not the commutative law.\\

For dual quaternions there are three different conjugates:
\begin{itemize}
 \item \textit{Conjugation of the quaternions:} $\overline{dq} := \overline{q_1} + \epsilon\cdot \overline{q_2}$ \quad $\forall \, dq \in \mathbb{H}_D$
 \item \textit{Dual conjugation:} $dq^{\epsilon} := q_1 - \epsilon\cdot q_2 $ \quad $ \forall \, dq \in \mathbb{H}_D$
\item \textit{Total conjugation:} $\overline{dq}^{\epsilon} := \overline{q_1} - \epsilon\cdot \overline{q_2}$ \quad $ \forall \, dq \in \mathbb{H}_D$
\end{itemize}

For the quaternion conjugate, the definition of  dual quaternion multiplication yields $dq_1$ and $dq_2$ that $\overline{dq_1**\, dq_2} = \overline{dq_2} **\, \overline{dq_1}$. The 2-norm of a dual quaternion is given by $\|dq\| := \sqrt{dq**\,\overline{dq}}$ and the inverse of a dual quaternion is $dq^{-1} = \frac{\overline{dq}}{\|dq\|^2}$.
In all three cases the quaternion conjugate is meant.\\

Dual quaternions can be used for the representation of pose in the three dimensional Euclidean Group. The quaternion $q_r$ representing the rotation is chosen to lay on the unit sphere $S_3$. To represent the translation $(t_1,t_2,t_3)^\top \in \mathbb{R}^3$ let $q_t := [0,t_1,t_2,t_3]$ be a second quaternion. Thus the dual quaternion $$dq := q_r + \epsilon \frac{1}{2} \cdot q_t*q_r$$ 
represents the  transformation in $S_3 \times \mathbb{R}^3$. \\
Any point $p = (u,v,w)^\top$ can be embedded to $\mathbb{H}_D$ by the dual quaternion $p_d = [1,0,0,0]+\epsilon \cdot [0,u,v,w]$. The transformation of this point about $dq$ is then $dq**\,p_d**\,\overline{dq}$.\\
This pose representation contains the important property that the composition of motions or of a pose followed by a motion is represented easily by the product of dual quaternions:
$$\begin{array}{lcl}
   p_{new} &=& dq_2 **\, dq_1**\,p_{old} **\,\overline{dq_1} **\, \overline{dq_2} \\
           &=& dq_2 **\, dq_1**\,p_{old} **\,\overline{dq_2 ** \,dq_1}\\
\end{array}$$\\

The rotation and translation a dual quaternion describes can also be expressed in terms of a rotation matrix $R$ and a translation vector $t$ by the formula:
$$\begin{pmatrix} R & t \\ 0 & 1 \end{pmatrix} \cdot \begin{pmatrix} p \\ 1 \end{pmatrix} $$ 
which is equivalent to the transformation of the point $p$, corresponding to $\begin{pmatrix} p \\ 1 \end{pmatrix}$, by the dual quaternion $dq = [q_r,\frac{1}{2}q_t*q_r]$ in ordinary form: 
$$dq **\, p**\,\overline{dq}$$
Remember the restriction that $q_r \in S_3$ and $q_t$ is an imaginary quaternion.\\
The equivalence is proven by the fact that rigid motion is equivalent to the one by rotation matrix and translation vector.\\
Let $dq_1$ and $dq_2$ be two dual quaternions with $\|q_{r_1}\| = \|q_{r_2}\|= 1 $ and  $\operatorname{Re}q_{t_1} = \operatorname{Re} q_{t_2}= 0 $
$$dq_1 = q_{r_1} +\epsilon q_{d_1} = q_{r_1} +\epsilon \frac{1}{2}\cdot q_{t_1}*q_{r_1}$$
$$dq_2 = q_{r_2} +\epsilon q_{d_2} = q_{r_2} +\epsilon \frac{1}{2}\cdot q_{t_2}*q_{r_2}$$
which represent the following transformations:
\begin{itemize}
 \item $q_{r_1}$ corresponds to the rotation matrix $R_1$
 \item $q_{r_2}$ corresponds to $R_2$
 \item $q_{t_1}$ corresponds to the translation vector $t_1$
 \item $q_{t_2}$ corresponds to $t_2$
\end{itemize}
Then the rigid motion 
$\begin{pmatrix} R_3 & t_3 \\ 0 & 1 \end{pmatrix} = \begin{pmatrix} R_2 & t_2 \\ 0 & 1 \end{pmatrix}\cdot \begin{pmatrix} R_1 & t_1 \\ 0 & 1 \end{pmatrix}$ is the same as:
\begin{flalign}
   dq_3 &= dq_2**dq_1\nonumber\\
	&= (q_{r_2} +\epsilon q_{d_2})**(q_{r_1} +\epsilon q_{d_1}) = q_{r_2}*q_{r_1}+\epsilon(q_{r_2}*q_{d_1}+q_{d_2}*q_{r_1})\nonumber\\
        &= q_{r_2}*q_{r_1}+\epsilon \frac{1}{2}\cdot(q_{r_2}*q_{t_1}*q_{r_1} + q_{t_2}*q_{r_2}*q_{r_1})\nonumber\\
        &= q_{r_2}*q_{r_1} +\epsilon \frac{1}{2} \cdot(q_{r_2}*q_{t_1}*\overline{q_{r_2}} + q_{t_2})*q_{r_2}*q_{r_1}\nonumber\\
        &= q_{r_3} +\epsilon q_{d_3}\nonumber
\end{flalign}
with $q_{r_3} = q_{r_2}*q_{r_1}$ and $q_{d_3} = \frac{1}{2}\cdot q_{t_3}*q_{r_3}$ where $q_{t_3} = q_{r_2}*q_{t_1}*\overline{q_{r_2}} + q_{t_2}$.\\
By matrix multiplication we get:
$$\begin{pmatrix} R_3 & t_3 \\ 0 & 1 \end{pmatrix} = \begin{pmatrix} R_2*R_1 & R_2t_1+t_2 \\ 0 & 1 \end{pmatrix}$$
Thus $R_3 = R_2*R_1$ and $t_3= R_2t_1+t_2$.\\

\section{Distribution Functions}\label{distribution}


To estimate the pose of an object or equivalently a feature of the object in the special Euclidean group $SE(3)$ we have to choose a probability density function ($pdf$). Most of the $pdf$s depend on some parameters that describe the function but they can be particle based as well. The case of a particle based description of the distribution will be handled in \ref{sample}. We want the $pdf$s to satisfy several characteristics:
\begin{itemize}
 \item The density function has to be independent from the coordinate system. Then a coordinate change causes just a change of the arguments of the $pdf$ but not a change of the structure of the parameters.
 \item The fusion of two probability density informations shall be supported. This is needed for maximum likelihood estimation for instance. 
\item The uncertain information of an objects pose or the relation of joints in a robots arm where each link has pose uncertainty with respect to the previous link for instance shall be propagated.
 \item The representation of the $pdf$ shall use a reasonably small set of parameters, much fewer parameters than needed for a particle set. Thus computations can be done efficiently.
\end{itemize}

In the following I will introduce some candidates for the probability distributions and their density functions.\\

\subsection{Projected Gaussian} \label{PG}

An intuitively good choice on the translation part is the multivariate normal distribution $\mathcal{N}(\mu,\Sigma)$. Now we would like to have something similar on the hypersphere $S_3$ embedded to $\mathbb{R}^4$ because that would make it easy to deal with correlations between rotation and translation.\\
An obvious approach for the rotation is the projection of a three dimensional Gaussian distribution from the tangent space to the $S_3$. We will do this by central projection (i.e. the center of the projection is the midpoint of the 3-dimensional unit sphere in $\mathbb{R}^4$ and the intersections of $S_3$ with the straight line through any point on the tangent space and the center of projection get the value of the normal distribution of the corresponding point on the three dimensional tangent space). \\

\begin{defn} \label{D1}
Let $S_3$ be the 3-sphere and $q_0$ be an arbitrary point on $S_3$. Further, let $TS_{q_0}\sim \mathbb{R}^3$ be the 3-dimensional tangent space to $S_3$ at the point $q_0$, with a local coordinate system that has the tangent point as origin. Now, let $\mathcal{N}(\mu,\Sigma)$ be a multivariate normal distribution on $TS_{q_0}$ which has $p_{TS}$ as corresponding probability density function. \\
Then the central projection 
$$\Pi_{q_0}: TS_{q_0}\longrightarrow S_3$$
provides a density function $$p_{S_3}(x):=\frac{1}{C}\cdot p_{TS}\left(\Pi_{q_0}^{-1}(x)\right)$$ on $S_3$, with $C=\int_{S_3}  p_{TS}\left(\Pi_{q_0}^{-1}(x)\right) \, \mathrm{d}x$. 
\end{defn}

\begin{figure}[bht]
 \centering
 \includegraphics[width=0.5\textwidth]{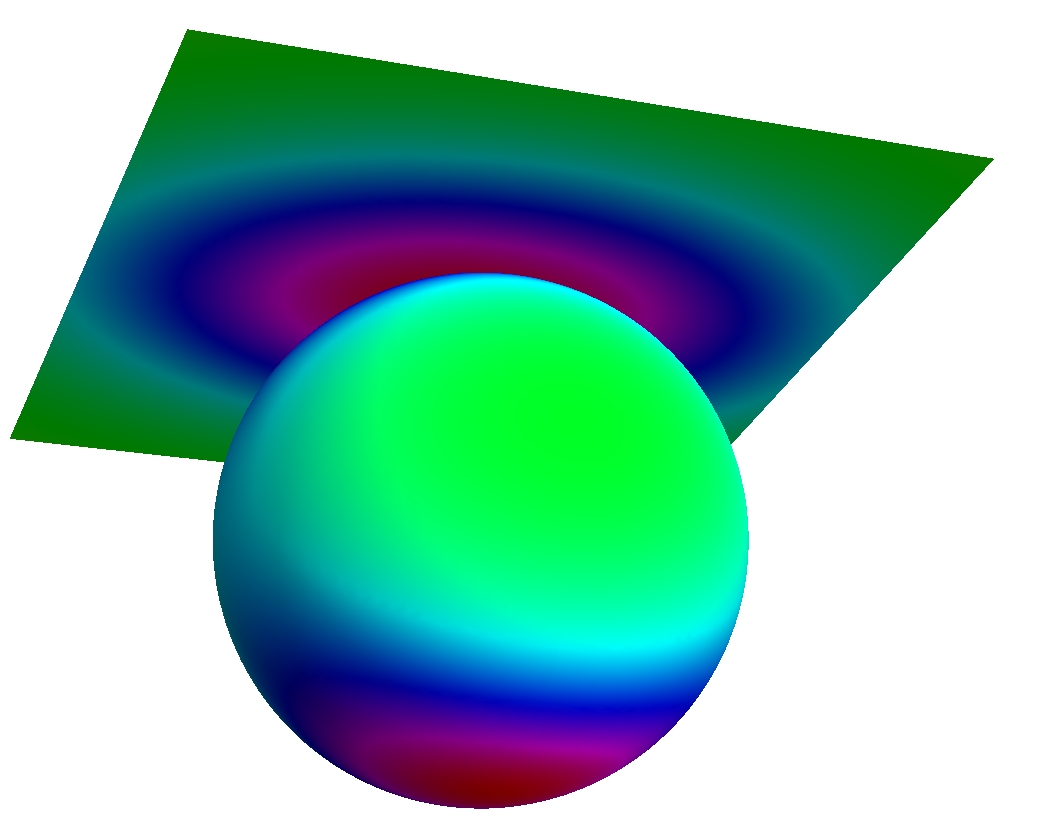}
\caption[Central projection]{The figure visualizes the central projection of a Gaussian distribution from the tangential plane to a unit sphere $S_2$.}
 \label{projection}
\end{figure}
 
As there are always two antipodal projected points on the sphere which represent the same point in the tangent space, this captures correctly the topology of the quaternion rotation space.\\

To define a \textbf{base element} in the special Euclidean group $SE(3)$ a tangent point on $S_3\subset \R^4$, a 6D mean vector $\mu$ and a $6 \times6$ covariance matrix $\Sigma$ are required. In $\R^4$ the quaternions offer a canonical way to create a non-vanishing continuous tangential vector field on the unit sphere $S_3$. Thus we can define a basis $B$ of the 3-dimensional tangent space $TS_{q_0}$ in $\mathbb{R}^4$ at the tangent point $q_0$. Further we complete $B$ to be a basis $B_0$ of $\mathbb{R}^4$ by concatenating as first vector the tangent space normal, which is the tangent point $q_0$. \\

\textit{How do we get the orthogonal vectors $q_1,\,q_2$ and $q_3$ of the basis $B$?}\\
Rotations in $\R^4$ can be represented by pairs of unit quaternions $q_l$, $q_r$, so that the rotated quaternion is given by $\mathrm{rot}(q) = q_l*q*\overline{q_r}$. Selecting $q_r = e_1$ and $q_l = q_0$, the canonical basis of $\R^4$ is rotated to the tangent point $q_0= [c_1,c_2,c_3,c_4]$ in quaternion writing. Thus the other vectors of the basis $B$ can be calculated by
$$q_i = q_0 * e_{i+1} * \overline{e_1} = q_0* e_{i+1}\quad \mathrm{for} \ i = 1,\,2,\,3$$
where $e_1$ and $e_{i+1}$ are the following unit quaternions $e_1 = \overline{e_1} = [1,0,0,0]$, $e_2 = [0,1,0,0]$, $e_3 = [0,0,1,0]$ and $e_4 = [0,0,0,1]$.\\
Than the basis is given by the following matrix:
$$B_0 = \left( q_0,\,\underbrace{q_1,\,q_2,\,q_3}_{=B}\right) = \left( \begin{pmatrix} 
 c_1 \\ c_2\\ c_3\\  c_4 \end{pmatrix}
\begin{pmatrix} 
    -c_2 & -c_3 & -c_4 \\
    c_1 & -c_4 & c_3\\
     c_4 & c_1 & -c_2\\
    -c_3 & c_2 & c_1\\ \end{pmatrix} \right)
= \begin{pmatrix} 
    c_1 & -c_2 & -c_3 & -c_4 \\
    c_2 & c_1 & -c_4 & c_3\\
    c_3 & c_4 & c_1 & -c_2\\
    c_4 & -c_3 & c_2 & c_1
\end{pmatrix}$$ \\
Anyway, the basis of the tangent space can be created randomly in all dimensions through orthogonalization by the Gram-Schmidt process followed by normalization.\\

Let us come back to the problem to define a base element in the $SE(3)$. I want it to be similar to a Gaussian distribution. Thus I will refer to the distribution function consisting of a Gaussian distribution for the rotation part which can be projected to the $S_3$ by central projection as introduced in \ref{D1} and another Gaussian distribution for the translation part of the rigid motion with \textit{base element}. \\
Subsuming the requirements, a base element with specified tangent point $q_0$ to the hypersphere $S_3$ and a basis $B_0$ of the tangent space $TS(q_0,B_0)$ is defined as:
$$\mathcal{N}\left(TS(q_0,B_0),\mu ,\Sigma \right)$$
As mentioned above in case of four dimensions the basis can be skipped, as we know then the canonical way of constructing the basis out of the tangent point.\\
The set of projected probability distributions with the projected density of the normal distribution as density function $p_{S_3}$ on the sphere $S_3$ is called the set of projected Gaussians (short $\mathrm{PG}$). The subset of $\mathrm{PG}$ for which $\mu =0$ in the corresponding normal distribution on the tangent space $TS_{q_0}$ is denoted as $\mathrm{PG}_0$. \\
Note that points $r^\bot \in S_3$ that are orthogonal to $q_0$ are not in the image of the central projection, and by consequence not in the domain of the inverse central projection. Since for each normal distribution the density goes to 0 as the argument goes to infinity, 0 is a continuous completion and we define: $p_S\left(r^\bot\right):=0$\\

If the probability density of a given pose shall be evaluated, the vector consisting of the first three entries of the mean vector, which is the mean of a Gaussian kernel in the tangent space, has to be projected to the 3-sphere by central projection as it represents a rotation. Thus the mean becomes a four dimensional vector with length 1 that represents the mean of the rotations on $S_3$. The last three entries just remain the way they are and correspond to the mean vector of the Gaussian distribution of the translations in $\mathbb{R}^3$. To introduce intuitively values to the covariance matrix it can be written in for of a block matrix:
$$\Sigma = \left( \begin{array}{cc}
\mathrm{covMatRotation} & 0 \\
0 & \mathrm{covMatTranslation} \\
\end{array}\right)$$
where 
$$\mathrm{covMatRotation} = \left(
\begin{array}{ccc}
     var(X_1) & cov(X_1,X_2) &  cov(X_1,X_3) \\
    cov(X_2,X_1) & var(X_2) &  cov(X_2,X_3) \\
    cov(X_3,X_1) & cov(X_3,X_2) &   var(X_3) \\
\end{array}
\right) $$
is the covariance matrix on the tangent space that has to be projected to $S_3$ to represent the covariance of the rotations in three dimensions and $\mathrm{covMatTranslation}$ is the covariance matrix on the other three dimensions that represent the translations.\\
Thus mean $\mu$ and covariance matrix $\Sigma$ together define a six dimensional Gauss kernel with density:
$$p(x) = \frac{1}{\sqrt{\det{(2 \pi \Sigma)} }} \exp\!\Big( {-\tfrac{1}{2}}(x-\mu)^\top\Sigma^{-1}(x-\mu) \Big)$$\\
A projected Gaussian on a 3-sphere needs to be normalized to 1 to define a probability density as mentioned already in the definition \ref{D1}. The normal renormalization constant  $\sqrt{\det{(2 \pi \Sigma)}}$ is not sufficient. Now I will explain how the correction weight for the parameterization in the integration to obtain the renormalization factor is calculated. The closed form of the renormalization factor $1/C$ itself involves confluent hypergeometric functions of a matrix argument and is quite complicated to calculate.\\

\begin{wrapfigure}{l}{7cm}
\centering
  \includegraphics[width=0.4\textwidth]{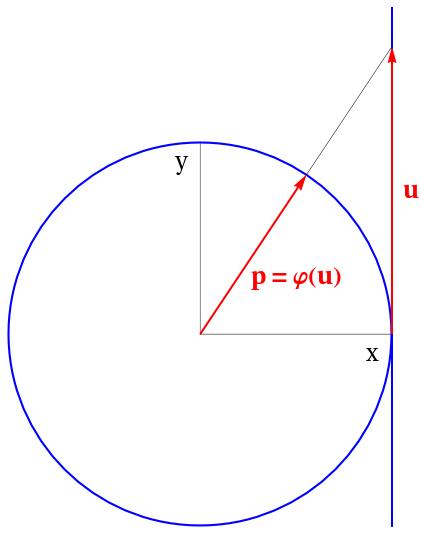}
  \caption[Parameterization of the sphere]{ }
\label{plot1}
\end{wrapfigure}

To calculate the surface integral we have to integrate in the directions of the coordinate axes with the Jacobian matrix as a factor for stretching the volume elements. We get this statement from the substitution rule for multiple variables:
\begin{theo}\label{proj}
 Let $U$ be an open set in $\R^n$ and $\varphi:U \to \R^n$ an injective differentiable function with continuous partial derivatives. The Jacobian $J_\varphi$ of $\varphi$ is nonzero for every $\mathbf{u}\in U$. Then for any compactly supported, continuous function $f$ with values in $\R$ and with support contained in $\varphi(U)$ it holds that:
 $$\int_{\varphi(U)} f(q)\,\mathrm{d}q = \int_U f(\varphi(\mathbf{u})) \left|\det(\operatorname{D}\varphi)(\mathbf{u})\right| \,\mathrm{d} \mathbf{u}$$
\end{theo}


In our case $U = \R^3$, $\mathbf{u} = (u,v,w)^\top$, $f = 1$ and $\varphi$ has to be defined as the following parametrization of the hemisphere around the $x$-axis of the hypersphere. See also figure \ref{plot1}
$$\begin{array}{lcl}
   \varphi : \R^3 &\to& S_3\\ 
   \begin{pmatrix} u\\v\\w\end{pmatrix}&\mapsto& \frac{1}{\sqrt{1+u^2+v^2+w^2}}\cdot \begin{pmatrix} 1\\u\\v\\w\end{pmatrix}
  \end{array}$$
It is sufficient to integrate over a hemisphere as the density of the projected Gaussian is antipodally symmetric and thus the overall integral easily is twice the integral over one half sphere.\\
The equation from theorem \ref{proj} reduces to:
$$\int_{S_3} 1\,\mathrm{d}q = \int_{\R^2} \sqrt{\det(J_\varphi^\top\cdot J_\varphi)}(\mathbf{u}) \,\mathrm{d}\mathbf{u}$$
where we replace $\left|\det(\operatorname{D}\varphi)(\mathbf{u})\right|$ by $\sqrt{\det(\operatorname{D}\varphi^\top\cdot \operatorname{D}\varphi)}(\mathbf{u})$ as the Jacobi matrix is not symmetric. \\
For $(J_\varphi)(\mathbf{u})$ we get:
$$\frac{1}{\left(1+u^2+v^2+w^2\right)^{\frac{3}{2}}}\cdot\begin{pmatrix}
    1+v^2+w^2 & -uv & -uw \\
   -uv & 1+u^2+w^2 &-vw \\
   -uw & -vw & 1+u^2+v^2 \\
   -u & -v & -w
\end{pmatrix}$$

and $(J_\varphi)^\top(\mathbf{u})$ is:
$$\frac{1}{\left(1+u^2+v^2+w^2\right)^{\frac{3}{2}}}\cdot\begin{pmatrix}
1+v^2+w^2 & -uv & -uw & -u\\
-uv & 1+u^2+w^2 & -vw & -v\\
-uw & -vw & 1+u^2+v^2 & -w
\end{pmatrix}$$\\

Now we can calculate $\sqrt{\det(J_\varphi^\top\cdot J_\varphi)}(\mathbf{u})= 1/(1+u^2+v^2+w^2)^2$ and thus obtain $\frac{1}{(1+u^2+v^2+w^2)^2}$ as correction weight for the parameterization in the integration.\\
As the calculation of a closed form of the renormalization factor $1/C = \int_{S_3} p_{S_3}(x)\mathrm{d}x$ is too complicated and very costly in calculation time we do this numerically with Monte Carlo integration which I will introduce in \ref{MC}. \\

An important property of projected Gaussians is the transformation invariance of its pose density. That means the density is independent from the coordinate system. I will show that the density at a certain pose in the euclidean space equals the density of the moved pose by dual quaternions. Every pose and motion in $SE(3)$ can be represented by a base element and the corresponding dual quaternion can be extracted from it. More explications are given in \ref{code} and the concrete calculations are given in the appendix \ref{A1}.\\

\begin{defn} 
Define the embedding $\mathcal{E}_Q$ of the $\mathbb{R}^4$ into the quaternions $\mathbb{H}$:
$$ \mathcal{E}_Q : \mathbb{R}^4 \to \mathbb{H}$$
$$(a,b,c,d)^\top \mapsto a+ib+jc+kd$$
\end{defn}
This function can be used to embed any 4-dimensional vector to the quaternions, but from now on, we will just embed unit vectors on the $S_3\subset \mathbb{R}^4$ to the quaternions by $\mathcal{E}_Q$.
\begin{defn} 
Furthermore define the embedding $\tilde{\mathcal{E}}_Q$ of the $\mathbb{R}^3$ into the imaginary quaternions $\mathbb{H}_{Im}$:
$$\tilde{\mathcal{E}}_Q : \mathbb{R}^3 \to \mathbb{H}_{Im}$$
$$(b,c,d)^\top \mapsto 0+ib+jc+kd$$
\end{defn}
The vectors in $\mathbb{R}^3$ that shall be embedded are not necessarily unit vectors.\\
Then it follows that there is an embedding of  $\mathbb{R}^4\times \mathbb{R}^3$ into the dual quaternions $\mathbb{H}_{DQ}$.
\begin{defn} 
Let $v_1=(a,b,c,d)^\top \in \mathbb{R}^4$ be a vector with length one, i.e. $\|v_1\| = \sqrt{a^2+b^2+c^2+d^2} = 1$ and $v_2 \in \mathbb{R}^3$. Then $q_r = \mathcal{E}_Q(v_1)$ and $q_t = \tilde{\mathcal{E}}_Q(v_2)$. Thus we can define 
$$\mathcal{E}_{DQ}:\mathbb{R}^4\times \mathbb{R}^3\to \mathbb{H}_{DQ}$$
$$ (v_1,v_2)\mapsto q_r+q_d = q_r + \epsilon\frac{1}{2}\cdot q_t * q_r = \epsilon\frac{1}{2}\cdot \tilde{\mathcal{E}}_Q(v_2)* \mathcal{E}_Q(v_1)$$\\
\end{defn}

Now the aim is to prove that the density at the original pose is equal to the moved density with respect to the new pose. \\

\textit{Proof}:\\
Let $dq_0 \in \mathbb{H}_D$ be a dual quaternion describing a starting pose in the Euclidean group. 
$$dq_0 = [q_{r0},q_{d0}] = [q_{r0},1/2\cdot q_{t0}*q_{r0}]$$
with base $B_0 = [b_1,b_2,b_3,b_4]$ where $b_1 = (a_0,b_0,c_0,d_0)^\top$ has the same entries as the tangent point $q_{r0} = [a_0,b_0,c_0,d_0]$. Then we obtain the base:
$$B_0 = \begin{pmatrix} 
a_0 & -b_0 & -c_0& -d_0\\
b_0 & a_0 & -d_0& c_0\\
c_0 & d_0 & a_0& -b_0\\
d_0 & -c_0 & b_0& a_0\\
\end{pmatrix}$$

Define $dq_c \in \mathbb{H}_D$ a constant rigid motion: $$dq_c = [q_{rc},q_{dc}] = [q_{rc},1/2\cdot q_{tc}*q_{rc}]$$
Then $dq_c$ applied to $dq_0$ defines the resulting pose $dq_1$:\\
$$\begin{array}{lcl}
dq_1 &=& dq_c **\, dq_0 \\
&=& [q_{rc},q_{dc}]**\,[q_{r0},q_{d0}] \\
&=& [q_{rc}*q_{r0},q_{rc}*q_{d0}+q_{dc}*q_{r0}] \\
&=& [q_{rc}*q_{r0},1/2 \cdot q_{rc}*q_{t0}*q_{r0}+1/2 \cdot q_{tc}*q_{rc}*q_{r0}] \\
&=& [q_{rc}*q_{r0},1/2 \cdot(q_{rc}*q_{t0}*\overline{q}_{rc}+q_{tc})*q_{rc}*q_{r0}]\\
\end{array}$$
where the conjugate $\overline{q}_{rc} = q_{rc}^{-1}$ as $q_{rc}$ is a unit quaternion.\\

Now let $rp = (u,v,w)^\top \in \mathbb{R}^3$ be a point in the tangent space of the tangent point $q_{r0}$. Then $q_{rp} = \mathcal{E}_Q(\frac{1}{\sqrt{1+u^2+v^2+w^2}}\cdot B_0\cdot (1,u,v,w)^\top)$ is the corresponding quaternion to $rp$ projected to the unit sphere $S_3$.
The point $tp$ defined as $(x,y,z)^\top \in \mathbb{R}^3$ represents a translation in the 3-dimensional space. $q_{tp} =\tilde{ \mathcal{E}}_Q((x,y,z)^\top) = [0,x,y,z]$ is the embedded point $tp$ in the imaginary quaternions $\mathbb{H}_{Im}$.
Then $dq_p = [q_{rp},q_{dp}] = [q_{rp},1/2\cdot q_{tp}*q_{rp}]$ has a certain density in the system of the original pose $dq_0$. \\

If I can show that the by $dq_c$ moved point $dq_c**\ dq_p$ corresponds to the same 6-dimensional point in the system of the new pose $dq_1$ than it holds that the density remains the same under motion by dual quaternions.\\

$$\begin{array}{lcl}
dq_c **\, dq_p &=& [q_{rc},q_{dc}]**\,[q_{rp},q_{dp}] \\
&=& [q_{rc}*q_{rp},q_{rc}*q_{dp}+q_{dc}*q_{rp}] \\
&=& [q_{rc}*q_{rp},1/2 \cdot q_{rc}*q_{tp}*q_{rp}+1/2 \cdot q_{tc}*q_{rc}*q_{rp}] \\
&=& [q_{rc}*q_{rp},1/2 \cdot(q_{rc}*q_{tp}*\overline{q}_{rc}+q_{tc})*q_{rc}*q_{rp}]\\
\end{array}$$
Name $dq_{pNew} := dq_c **\, dq_p$ the moved point, then $q_{rpNew} = q_{rc}*q_{rp}$ and $q_{tpNew} = q_{rc}*q_{tp}*\overline{q}_{rc}+q_{tc}$.\\

First I will show that the back projected point $\mathcal{E}_Q^{-1}(q_{rpNew})$ corresponds to the same point $rp=(u,v,w)^\top$ in the new tangent space with base $B_1$ at the tangent point $q_{r1}$.\\
This is equivalent to showing that $q_{rpNew} = \mathcal{E}_Q(\frac{1}{\sqrt{1+u^2+v^2+w^2}}\cdot B_1\cdot (1,u,v,w)^\top)$.\\
Let be $q_{rc} = [a_c,b_c,c_c,d_c]$ than $q_{r1} = q_{rc}*q_{r0} = [a_0 a_c - b_0 b_c - c_0 c_c - d_0 d_c, a_c b_0 + a_0 b_c + c_c d_0 - c_0 d_c, a_c c_0 + a_0 c_c - b_c d_0 + b_0 d_c, b_c c_0 - b_0 c_c + a_c d_0 + a_0 d_c]$ and thus 
$$B_1 = \begin{pmatrix} 
a_1 & -b_1 & -c_1& -d_1\\
b_1 & a_1 & -d_1& c_1\\
c_1 & d_1 & a_1& -b_1\\
d_1 & -c_1 & b_1& a_1\\
\end{pmatrix}$$ where $a_1 = a_ca_0 - b_cb_0 - c_cc_0 - d_cd_0$, $b_1 = a_cb_0 + b_ca_0 + c_cd_0 - d_c1c_0$, $c_1 = a_cc_0 - b_cd_0 + c_ca_0 + d_cb_0$ and $d_1 = a_cd_0 + b_cc_0 - c_cb_0 + d_ca_0$. \\
Now we can calculate:
$$\mathcal{E}_Q(\frac{1}{\sqrt{1+u^2+v^2+w^2}}\cdot B_1\cdot (1,u,v,w)^\top)$$
$$\begin{array}{lcl}
= \frac{1}{\sqrt{1+u^2+v^2+w^2}}&\cdot&[a_0a_c-b_0b_c-c_0c_c-d_0d_c+(-a_cb_0-a_0b_c-c_cd_0+c_0d_c)u\\
&+&(-a_cc_0-a_0c_c+b_cd_0-b_0d_c)v+(-b_cc_0+b_0c_c-a_cd_0-a_0d_c)w,\\
&&a_cb_0+a_0b_c+c_cd_0-c_0d_c+(a_0a_c-b_0b_c-c_0c_c-d_0d_c)u\\
&+&(-b_cc_0+b_0c_c-a_cd_0-a_0d_c)v+(a_cc_0+a_0c_c-b_cd_0+b_0d_c)w,\\
&&a_cc_0+a_0c_c-b_cd_0+b_0d_c+(b_cc_0-b_0c_c+a_cd_0+a_0d_c)u\\
&+&(a_0a_c-b_0b_c-c_0c_c-d_0d_)v+(-a_cb_0-a_0b_c-c_cd_0+c_0d_c)w\\
&&b_cc_0-b_0c_c+a_cd_0+a_0d_c+(-a_cc_0-a_0c_c+b_cd_0-b_0d_c)u\\
&+&(a_cb_0+a_0b_c+c_cd_0-c_0d_c)v+(a_0a_c-b_0b_c-c_0c_c-d_0d_c)w]\\
\end{array}$$
On the other side:
$$\begin{array}{lcl}
q_{rpNew} &=& q_{rc}*q_{rp} \\
&=& q_{rc}*\mathcal{E}_Q(\frac{1}{\sqrt{1+u^2+v^2+w^2}}\cdot B_0\cdot (1,u,v,w)^\top) \\
&=& \frac{1}{\sqrt{1+u^2+v^2+w^2}}\cdot q_{rc}*\mathcal{E}_Q(B_0\cdot (1,u,v,w)^\top)\\
&=& \frac{1}{\sqrt{1+u^2+v^2+w^2}}\cdot [a_c,b_c,c_c,d_c]*[a_0-b_0u-c_0v-d_0w,b_0+a_0u-d_0v+c_0w,\\
&&c_0+d_0u+a_0v-b_0w,d_0-c_0u+b_0v+a_0w]\\
\end{array}$$
On evaluating this, we get $q_{rpNew} = \mathcal{E}_Q(\frac{1}{\sqrt{1+u^2+v^2+w^2}}\cdot B_1\cdot (1,u,v,w)^\top)$ as we wanted.\\

For the translation part there is no such property as for the rotation because the coordinate system remains unchanged. But we can calculate the vector $v_{tp,t0}$ between the point $tp$ and the point corresponding to the quaternion that represents the position of the system:
$$v_{tp,t0} = tp - \tilde{\mathcal{E}}_Q^{-1}(q_{t0})$$
Now we need to show that $v_{tp,t0} = v_{tpNew,t1}$.
$$\begin{array}{lcl}
v_{tNew,t1} &=& \mathcal{E}_Q^{-1}(q_{rc}*q_{tp}*\overline{q}_{rc}+q_{tc}) - \mathcal{E}_Q^{-1}(q_{rc}*q_{t0}*\overline{q}_{rc}+q_{tc})\\
&=& \tilde{\mathcal{E}}_Q^{-1}(q_{rc}*q_{tp}*\overline{q}_{rc}+q_{tc}-q_{rc}*q_{t0}*\overline{q}_{rc}-q_{tc})\\
&=& \tilde{\mathcal{E}}_Q^{-1}(q_{rc}*q_{tp}*\overline{q}_{rc}- q_{rc}*q_{t0}*\overline{q}_{rc})\\
&=& \tilde{\mathcal{E}}_Q^{-1}(q_{rc}*(q_{tp}-q_{t0})*\overline{q}_{rc})\\
&=& \tilde{\mathcal{E}}_Q^{-1}(q_{tp}-q_{t0})\\
&=& v_{tp,t0}\\
\end{array}$$ 
as we know that $\mathrm{rot}(q) = q_{rc}*q *\overline{q}_{rc}$ is the rotation of $q$ about $q_{rc}$ in quaternionic writing and rotation is a length and orientation preserving operation.
\EndProof

I recapitulate that we model a distribution at some point and then know the parameters of the distribution under some rigid motion. For the rotation part we just rotate the tangent points of the mixture elements, for the translation part we just translate the translation part of the parameter vector. Further remember that the "zero mean" requirement of $\mathrm{PG}_0$ only concerns the rotation part of the parameter vector.\\

\textit{The projected Gaussians fulfill all of the upper named desired requirements of the distribution function:}
\begin{itemize}
\item This density is independent from the coordinate system.
\item The fusion of two probability density informations is supported as well as propagation of uncertain information. I will explain it later on in the more general case of mixtures of projected Gaussians \ref{MoPGs}.
\item The representation just needs the parameters $TS(q_0,B_0),\, \mu$ and $\Sigma$.\\
\end{itemize}

\subsection{Bingham}


The Bingham distribution is an antipodally symmetric probability distribution on a unit hypersphere $S_d$. 
\begin{defn}
The probability density function of a Bingham distribution is defined as:
$$f(x,\Lambda,V)=\frac{1}{F}e^{\sum_{i=1}^d\lambda_i(v_i^\top x)^2}=\frac{1}{F}e^{x^\top C x}$$
where the first expression is the standard form for Bingham distributions.
$F$ is the normalization constant of the distribution, $\Lambda$ is a vector of concentration parameters, the columns of the $(d + 1) \times d$ matrix $V$ are orthogonal unit vectors and $C$ is a $(d + 1) \times (d + 1)$ orthogonal matrix.
\end{defn}
As can be seen from the second form, the Bingham distribution is derived from a zero-mean Gaussian on $\mathbb{R}^{d+1}$. It is conditioned to lie on the surface of the unit hypersphere $S_d$ and thus models rotational probability densities the best. This property is utilized by Alexander and Buxton in their work \cite{ColourData}. Just as the projected Gaussian, the Bingham distribution fits the antipodal symmetry of the quaternions, since the unit quaternions $q$ and $-q$ represent the same rotation in the 3D space. \\
A big disadvantage of the Bingham distribution is the computationally expensive renormalization constant $F$ which does not have a closed form in general. Since the distribution must integrate to 1 over $S_{d}$, this constant can be written as
$$F(\Lambda)=\int_{x\in \mathbb{S}^{d}}e^{\sum_{i=1}^d \lambda_i(v_i^T x)^2} \mathrm{d}x$$  
Glover \cite{Jared} solved this efficiency problem by using a precomputed lookup table to approximate $F$.\\

The main advantage of using a Bingham distribution to model the probabilities of three dimensional orientations on the quaternion hypersphere, is to handle distributions with rotational symmetry in a compact way. In a personal communication Glover told me that they can be used without undue linear approximations like necessary for  distributions such as projected Gaussians, which could cause distortions and require more mixture components. The Bingham distribution is well suited to hyperspherical distributions with high variance. However, it remains to be seen whether the benefits of using the Binghams outweigh their added complexity compared to projected Gaussians. Moreover, in the case of a peaked probability density, the difference in accuracy are supposed to vanish. \\

\textit{In summary which of the desired properties does the Bingham distribution fulfill?}\\
The density function is independent from the coordinate system as we wanted. The procedure to merge two Bingham kernels $f_1(X)$ and $f_2(X)$ with its weights $\lambda$ and $(1-\lambda)$ by a single one, is described by J. Glover in his paper. The idea behind this merge is to find a Bingham distribution $f$ that fits the joint inertia matrix $S_f$ which is easy to calculate:
$$S_f = \lambda \mathrm{E}_1[xx^\top] + (1-\lambda)\mathrm{E}_2[xx^\top]$$ 
where $\mathrm{E}_1[xx^\top]$ is the inertia matrix of $f_1(X)$ and $\mathrm{E}_2[xx^\top]$ is the one of $f_2(X)$. The Inertia matrices can be derived from the exponent of the Bingham density: 
$$ \frac{1}{N} \sum_{i=1}^d\lambda_i(v_i^\top x_i)^2 = v_j^\top S v_j$$
The propagation of uncertain information can't be done straight forward with this density function. The composition of Bingham kernels does not provide another Bingham density. At least by now there is no method known to obtain another Bingham kernel. The true distribution can just be Bingham approximated. Moreover it is difficult to represent correlation between rotation and translation. And finally I want to repeat that Bingham distributions are more complex than projected Gaussians and the little gains in accuracy compared to projected Gaussians don't justify this inefficiency. \\

\subsection{von Mises-Fisher}


The von Mises distribution can be thought of as the spherical analogue of the normal density \cite{SO3}. It is a continuous probability distribution on the $(n - 1)$-dimensional sphere in $\mathbb{R}^n$. For $n=2$ the distribution reduces to the so called circular normal distribution found by von Mises. In the case $n=3$ it is called Fisher distribution. \\
\begin{defn}\label{MF}
In general the density of the von Mises-Fisher distribution for $v$, a $n$-dimensional random unit vector, is given by:
$$p(v, \mu, \kappa)=\frac {\kappa^{n/2-1}} {(2\pi)^{n/2}I_{n/2-1}(\kappa)} \cdot\mathrm{e}^{\kappa \mu^T v}$$
where $I_{n/2-1}$ denotes the modified Bessel function of first kind and of order $\frac{n}{2}-1$, $\|\mu\| = 1$ and $\kappa \geq 0$. Then $\mu$ is called the mean direction and $\kappa$ is the concentration parameter. 
\end{defn}
The parameters $\mu$ and $\kappa$ determine the shape of the density. As $\kappa$ increases, the concentration of the distribution around the mean direction becomes higher and the density approaches a normal density.\\

Back to the case $n=2$, the von Mises density reduces to:
$$ p(v,\mu,\kappa)=\frac{1}{2\pi I_0(\kappa)}e^{\kappa\cos(v-\mu)} $$
where $I_0$ is the modified Bessel function of order 0.\\
We want to define a probability density on the special orthogonal group $SO(3)$ from the von Mises distribution in matrix form. Let $R \in SO(3)$, then $R$ is said to have the von Mises-Fisher matrix density, if: 
$$p(R) = \frac{1}{c_F} \mathrm{e}^{tr\left[F\cdot R^\top\right]}$$
with respect to the uniform distribution $\mathcal{U}(SO(3))$. $F$ is a $3\times 3$ parameter matrix containing the concentration and $1/c_F$ is the normalization constant depending on $F$.\\

\textit{Does the von Mises-Fisher distribution satisfy the desired characteristics?}\\
It is independent from the coordinate system but as in matrix form the density is dependent of the matrix $R$ the necessary number of parameters is nine to represent a 3-dimensional density.\\
In definition \ref{MF} the distribution is defined for any unit sphere, so it could be used for the unit quaternions as well and thus four parameters would be required for the rotation. This distribution would not have the antipodal symmetry, though.\\
Within the limits of this work it was not feasible to check for the applicability of information fusion and propagation of uncertain information.\\

\subsection{Sample Based Description}\label{sample}

Instead of choosing a probability density function to approximate the true distribution of the pose in $SE(3)$, we could also work directly with samples. This has the advantage that sampling is extremely general and flexible. Moreover there is a direct proportionality between the numbers of samples and the accuracy. But to get significant results, a large number (the square of what is needed in $\mathbb{R}^3$) is necessary to accurately describe a probability distribution in a six dimensional space.\\

Thus the desired feature of few parameters is not satisfied and due to the number of necessary samples, the computation gets very slow.\\
On the other hand the value of the sample based representation as universally usable for all distributions, with arbitrary precision should be honored. Further it can at least be used as a vehicle to obtain experimental results at least in off line calculations.\\

\section{Mixture of Projected Gaussians}\label{MoPGs}

A typical application for mixtures of projected Gaussians is the following example:
\begin{quote}
Given the height of each person in a mixed group of people. As the average height of women and men both can be modeled by a Gaussian kernel the mixture of these two kernels models the distribution of heights of the whole group. Now one could calculate the probability of single persons to be male or female just from the information how tall they are.\\ 
\end{quote}

In our framework we want to model an approximation of an unknown probability distribution instead of finding disjoint classes. As the probability distributions that can be represented by a single base element are limited, we want to combine several of them to describe more complex distributions as introduced in \cite{MoG}. This specific use of the concept of mixtures of Gaussian distributions is not as common as the one given in the example above.\\

\begin{defn}\label{MoPG}
Let $\mathrm{PG_i} := \mathcal{N} (TS_i , \mu_i , \Sigma_i )$ be a sequence of projected Gaussians for $i \in \{1,\ldots,n\}$ with $\mu_i$ a d-dimensional mean vector, $\Sigma_i$ a $d\times d$ covariance matrix and $TS_i := TS(q_i, B_i)$ a corresponding tangent space consisting of a tangent point $q_i$ and a basis $B_i$. Furthermore we require $\sum_{i=1}^n \lambda_i = 1$ and $0 \leq \lambda_i \leq 1$ $\forall\, i\in \{1,\ldots,n\}$.\\
Then we define a mixture of projected Gaussians $\mathrm{MoPG}$ as follows:
$$\begin{array}{lcl}
 \mathrm{MoPG} &=& \lambda_1\cdot \mathrm{PG_1} + \lambda_2\cdot \mathrm{PG_2} + \ldots + \lambda_n\cdot \mathrm{PG_n}\\
               &=& \sum_{i = 1}^n \lambda_i\cdot \mathrm{PG_i}\\
               &=& \sum_{i = 1}^n \lambda_i\cdot \mathcal{N} (TS_i , \mu_i , \Sigma_i )\\
\end{array}$$\\
\end{defn}

In four dimensions it is sufficient to specify a tangent point $q_i$ instead of a whole tangent space $TS_i$ as we know the instruction to construct the canonical basis out of the tangent point.\\

Such a mixture of projected Gaussians is similar to the multivariate normal distribution. It behaves like a normal distribution on the tangent space and can be treated as one. Another advantage is the low number of parameters. In the case that the tangent point is situated on the hypersphere $S_3\subset \R^4$ we only have a four dimensional vector in addition to the parameters of the normal distribution $\mu$ and $\Sigma$.\\
A MoPG can describe a wide range of distributions from highly peaked ones to wide spread ones. Furthermore it can easily represent correlation between rotation and translation, what is important to model object features properly. I conjecture that a mixture of projected Gaussians (including mixtures with zero variance) can approximate any antipodally-symmetric density. But to approximate a uniform distribution or step function, which doesn't coincidentally form the shape of a bell, a large number of Gaussian kernels are required, to receive fine results.\\

Of course one has to keep in mind that a MoPG is just an approximation of the true distribution function that would be appropriate for pose estimation on the special Euclidean group. The more base elements the mixture contains, the better the true distribution can be approximated, what is conflicting the wish about efficient computability. \\

\subsection{Probabilistic Inferences on MoPGs}\label{inferences}

From the transformation invariance of the pose density of single base elements it is easy to deduce that a complete mixture of projected Gaussians is independent from the coordinate system as well, because we restrict the mixture to consists of a finite sum of base elements.\\

MoPGs support data fusion. This means the component wise fusion of each element of one mixture with every element of another mixture. Fusing two base elements $PG_i$ and $PG_j$ means calculating a combined base elements $PG_{ij}$ out of the original ones. Therefore a new tangent point $p_{ij}$ on the sphere is determined which lies in the middle between the tangent points $p_i$ and $p_j$ of the base elements to be fused. At $p_{ij}$ the new tangent space is created and the base elements $PG_i$ and $PG_j$ are transformed to this new tangent space. Then the fused covariance matrix $\Sigma_{ij}$ has to be calculated from the covariance matrices $\Sigma_i$ and $\Sigma_j$:
$$\Sigma_{ij} = \Sigma_i\cdot (\Sigma_i + \Sigma_j)^{-1}\cdot \Sigma_j$$
The mean vector $\mu_{ij}$ is obtained from $\mu_i$ and $\mu_j$ by the formula:
$$\mu_{ij} = \Sigma_j\cdot(\Sigma_i + \Sigma_j)^{-1}\cdot \mu_i + \Sigma_i\cdot (\Sigma_i + \Sigma_j)^{-1} \cdot\mu_j$$

A disadvantage of this procedure is the quickly growing number of elements of the mixture. Therefore a reduction algorithm of the number of summands of a MoPGs is introduced, namely the merge of similar base elements to keep the number small. Actually to merge base elements $PG_i$ and $PG_j$, the weights $\lambda_i$ and $\lambda_j$ of these projected Gaussians in the mixture need to be known. That's why I require the input of $\lambda_i$ and $\lambda_j$ to the function. Likewise in data fusion, the new tangent point $p_{ij}$ is calculated from $p_i$ and $p_j$ and $PG_i$ and $PG_j$ are transformed to the new tangent space $TS_{p_{ij}}$ at $p_{ij}$. Now the new mean $\mu_{ij}$ and the new covariance matrix $\Sigma_{ij}$ can be calculated:
$$\mu_{ij} = \frac{\lambda_i}{\lambda_i+\lambda_j}\mu_i + \frac{\lambda_j}{\lambda_i+\lambda_j}\mu_j$$
$$\Sigma_{ij} = \frac{\lambda_i}{\lambda_i+\lambda_j}\Sigma_i+\frac{\lambda_j}{\lambda_i+\lambda_j} \Sigma_j+\frac{\lambda_i\cdot \lambda_j}{\lambda_i+\lambda_j}(\mu_i-\mu_j)(\mu_i-\mu_j)^\top$$
The weight $\lambda_{ij}$ of the new base element $PG_{ij}$ is the sum of the weights of the input base elements $\lambda_i+\lambda_j$.\\
I distinguish clearly between fusing and merging base elements or mixtures as fusion of two MoPGs means examining the information of both of them, in contrast to merging base elements or whole mixtures what stands for joining it to a single one.\\

Further the composition of a certain or uncertain motion mixture and a pose modeling mixture is supported. 
To compose a motion base element with another base element describing a pose, the dual quaternions of the base elements are extracted and executed one to the other. Thereby the new mean $\mu_{ij}$ is obtained automatically. The covariance matrix $\Sigma_{ij}$ is calculated by applying the Jacobian matrix $J$ from both sides to the block covariance matrix:
$$\Sigma_{ij} = J\cdot \begin{pmatrix} \Sigma_i & 0\\ 0 & \Sigma_j\end{pmatrix} \cdot J^\top$$
As the composition usually contains uncertainties which are included in the covariance matrix of the composed base element, we call this instruction random pose transformation. For the case that the motion is secure, the covariance matrix of the rigid motion can easily be set to zero.\\

That MoPGs consist of a reasonably small set of parameters is an advantage that allows efficient computations with this distribution function.\\

\subsection{Comparison of Mixtures of Binghams with MoPG}

Both kinds of mixtures are used in similar applications. Some part of the natural scientists prefers mixtures of Binghams as they are the appropriate distribution to model probabilities on sphere surfaces. The other part is persuaded that the modeling error made by the use of mixtures of projected Gaussians is negligibly small and that the facility of working with this kind of distribution outweighs this error by far. Further as in our context we use the mixture of projected Gaussians to approximate the uncertain pose of an object the accuracy that can be reached is limited. \\
However I want to point out the differences and similarities of the two mixtures.\\

\begin{itemize}
\item For the renormalization constant of both the Bingham distribution and the projected Gaussian no closed form of the integral exists and thus the factor must be approximated. Glover suggests to transform the series expansion of the factor in a way that an approximation of the normalizing constant of the Bingham distribution can be done using standard floating point arithmetic. As this procedure still is very slow he uses a precomputed lookup table in his framework.\\
We approximate the renormalization factor of the projected Gaussian distribution via Monte Carlo integration. For the number of samples $n = 1000$ the error ranges in order of magnitude $10^{-3}$ and the process time is less than half a minute for each base element. For $n = 5000$ the error ranges in order of magnitude $10^{-4}$ and the process time is about one and a half minutes. We don't need more precise results. 
\item In the case that a widely spread distribution similar to an uniform distribution shall be modeled it is undoubted that a mixture of Bingham distributions requires a lower number of elements of the mixture than a MoPG. If a distribution with peaked shape shall be modeled, it stands out to proof whether the computation with mixtures of Binghams or with MoPGs is more efficient as than the number of necessary elements of the mixture approximates.
\item In both distribution functions the low number of parameters is convincing. This could just be depreciated if the number of mixture elements becomes big. 
\item How the merge of two single Gaussian kernels is defined I already introduced in \ref{inferences}. For a mixture of Bingham distributions consisting of two kernels $f(X) = \alpha f_1(X)+(1-\alpha)f_2(X)$ Glover describes an algorithm to approximate $f(X)$ by a single Bingham $g(X)$ in the preprint \cite{JaredOld}. With maximum likelihood parameter estimation the sufficient statistic is the sample inertia matrix $\hat S = 1/N \sum_{i=1}^N x_i\cdot x_i^\top$ for $x_i$ from the sample set $\{x_1,\ldots,x_N\}$. $\hat S$ goes to the true inertia matrix $S = \mathrm{E}[x\cdot x^\top]$ as $N\to\infty$. Thus the Bingham distribution $g(X)$ which fits the inertia matrix $S_f = \alpha \mathrm{E}_1[x\cdot x^\top] + (1-\alpha)\mathrm{E}_2[x\cdot x^\top]$ is the maximum likelihood fit of a single Bingham to the mixture $f(X)$. By the way this Bingham distribution $g(X)$ has minimal KL divergence to $f(X)$.
\item Let $q$ and $r$ be two Bingham distributed random variables. The composition of the Binghams $p(q)$ and $p(r)$ can be done by the use of the method of moments. This yields a Bingham approximation to the true distribution, $p(qr)$, by computing $\mathrm{E}[qr(qr)^\top]$. Glover plans to develop the composition algorithm for mixtures of Binghams in a future paper. By now it is already secure that the result of composing two Bingham distribution doesn't give a Bingham and thus better results can be achieved by the composition of MoPGs or base elements of the mixture like introduces in \ref{inferences}. 
\item By now there is no operation known for data fusion of mixtures of Binghams.
\end{itemize}

\chapter{Approximation}\label{Approx}

The quintessence of this chapter shall be to find suitable approximations of mixtures of projected Gaussians to cut down computational effort with minimal loss of accuracy. At least I want to determine an upper bound for the impreciseness resulting from the approximation.\\
We know that by an infinite mixture of projected Gaussians the pose of a target object could be described user-defined precisely. Of course such a mixture doesn't exist in practice and thus the finite mixtures we use just approximate the true pose. In section \ref{prop} I introduce criteria to reduce the number of base elements of a MoPG without loosing much from the preciseness of the mixture. Further the section handles a convergence criterion that improves the approximation by fusion of information data. Section \ref{AA} is concerned with two approaches how a MoPG can be fitted to another. The expectation maximization algorithm fits a mixture to a set of samples drawn from the other mixture by iteratively increasing the log likelihood function of the sample set. In the other approach the $L^p$ norm between the densities of the mixtures is minimized.\\

Moreover an aim is to study the coherence between particle sets and MoPGs. Therefore let's denote the \textit{composition}, the \textit{fusion} and the \textit{merge} of mixture densities as modification operations.\\
Let $p_1, \, p_2$ and $p_3$ be the density functions of the mixtures $M_1, \, M_2,\, M_3 \in \mathrm{MoPG}$. $p_3$ shall be generated by one of the modification operations out of $p_1$ and $p_2$. If we draw a set of samples $Z_1 = \{z_{1,n}\}_n$ respectively $Z_2 = \{z_{2,n}\}_n$ from each of the densities $p_1$ and $p_2$ and apply the same modification operation to them, we obtain the particle set $Z_3 = \{z_{3,n}\}_n$. The probability density $\tilde p_3$ is obtained by fitting a mixture of projected Gaussians to the sample set $Z_3$.
$$\begin{matrix} 
 & (p_1,p_2) & \xrightarrow{\text{modification operation}} & p_3 & \\
 &   &  & \tilde p_3 & \\
{\text{\small{draw samples}}}&\downarrow \quad \downarrow &  &  \uparrow & {\text{\small{fit mixture}}} \\
 & (Z_1, Z_2) & \xrightarrow{\text{modification operation}} & Z_3
\end{matrix}$$
Now we want to examine the dissimilarity between $p_3$ and $\tilde p_3$. We expect it to disappear for a growing number of samples. I introduce several distance measures and study convergence measures in section \ref{gripper} as it is not clear by now which measure for the dissimilarity of probability density functions is appropriate in out topic. In the context of this work it was not possible to solve the distance problem between $p_3$ and $\tilde p_3$ completely. Hence I just introduce some considerations about convergences of approximations in general in \ref{prop}.\\
We also need to know the similarity respectively the dissimilarity of density functions to evaluate their importance for the accuracy of the mixture and to be able to decide whether an object is graspable for the robot. In the following section these distance measures will be examined further.

\section{Grasp Criterion}\label{gripper}

To decide whether an object is graspable for a robot one has to consider several aspects. For instance the relations of the joints in a robots arm contains weak information what has the effect that the pose of the gripper contains little uncertainties. Further the imperfect sensors of the robot produce weak information data and thus the pose of the target object is uncertain. Hence the probability for failure of any grasping task is composed of several failure probabilities. If for an attempt to grasp a failure threshold of $\varepsilon>0$ is allowed, it has to be split up into the part $\varepsilon'>0$ for the sensor and camera uncertainty, or the pose uncertainty of the gripper $\varepsilon''>0$ and so on.\\
The part $\varepsilon'''$ determines the impreciseness of the MoPG. Thus as many base elements of the mixture $M$ with low weight can be discarded as the remaining approximated mixture $M_{app}$ still has a total preciseness $\geq 1-\varepsilon'''$.\\

Mainly I found two different criteria to measure whether an object is graspable for a robot. \\
The one of them is a kind of \textit{box criterion} because a box $B$ describes the size the gripper can encompass. Thus the robots hand needs to be navigated to a pose where the object fits inside the box which represents the robots hand. In other words the right box $B$ in the set of boxes $\mathbb{B} \subset \R^6$ with defined size has to be selected to enable the robot to grasp the object. The box selection is done by an $\arg \max$ function $f_p$ that chooses the box, which contains the maximum of the probability mass to represent the pose of the object, with respect to the probability density $p$ of the mixture distribution.
$$f_p:=\arg \max_{B\in \mathbb{B}} \int_{B} p(x) d\mu_B$$
where $p(x)$ is the probability density at the random point $x\in B$. Then of course the mean of the pose of the gripper has to be navigated to the center of the box selected by $f_p$.\\
Even though this criterion is intuitively the correct one, the problem arises to find an $\arg \max$ function that is efficient in practical use and determines the correct box containing the maximum of probability mass. This can become difficult in the case of a non-symmetric probability distribution.\\

The other criterion concerns the distance between the probability density that describes the gripper and the one that describes the objects pose. The probability that the robot will succeed on grasping the object has its maximum at the point where the distance of the values of the densities of the random variables is minimal, which describe the pose of the robots hand and the estimated pose of the object. Hence the distance between gripper and object pose needs to be small in terms of rotation and translation.
Further we require the pose density of the gripper to be focused. As a result, also the density describing the object's pose will have to be focused. 
\begin{defn}
Let $p$ be the density of the random variable describing the objects pose and $g$ the reasonably strong focused density of the one that estimates the pose of the gripper. Define a threshold $G$ when the distributions are close enough such that the excepted probability for failure is smaller than $\varepsilon >0$.\\
Then the object is called graspable with error $<\varepsilon$, when: 
$$\mathrm{P}\left(dist(g,p) \leq G\right) >1- \varepsilon$$ 
where $dist(\cdot, \cdot)$ is an appropriate distance measure of probability density functions. 
\end{defn}
The grasp criterion basing on distances of densities of random variables is in most cases the preferable one for our topic.\\

\subsection{Distance Measures}

In this section I want to examine some distance measures and their characteristics. This is not only necessary for the grasp criterion but also to evaluate similarity of Gaussian kernels in a mixture.\\

At first I want to recall some basics about metrics and topologies in general \cite{ProbabilityMetrics}:
\begin{defn}
For any point $x$ in a metric space $M$ we define the open ball of radius $r\, (>0)$ around $x$ as the set: 
$$B(x, r) = \{y \in M : d(x,y) < r\}$$
A subspace of $M$ is a neighborhood of $x$ if it contains an open ball about $x$. 
\end{defn}

Each of these neighborhoods fulfill the axioms of a topology and therefore the union defines a topology on $M$ - the \textit{induced topology}. 
Every metric space is a topological space in a natural manner, i.e. all definitions and theorems about topological spaces also apply to all metric spaces.\\

\begin{defn}
A function $f : M_1 \longrightarrow M_2$, from one topological space $M_1$ to another $M_2$, is \textit{continuous} if and only if the inverse image of every open set is open: \\
$\forall V$ open, $V \subseteq M_2$, the inverse image $f^{-1}(V) = \{x \in M_1 \; | \; f(x) \in V \}$ is open.
\end{defn}

\begin{theo}$(\varepsilon$-$\delta)$-continuity of maps:\\
Let $(M_1,d_1)$ and $(M_2,d_2)$ be metric spaces and $f: M_1 \longrightarrow M_2$ a map. $f$ is continuous if 
$\forall x \in M_1$ and $\forall \varepsilon>0$ $\exists\delta>0$ such that $\forall y \in M_1:$\\
$d_1(x,y)<\delta \Rightarrow d_2(f(x),f(y))<\varepsilon$
\end{theo}

In general a \textit{norm} determines a metric and all metrics induce topologies, but the inverse is not true. A metric defines a norm only if it is \textit{translation invariant}, i.e. $d(x,y) = d(\alpha +x,\alpha+y)$ and \textit{homogeneous}, i.e. $d(\alpha x,\alpha y) = |\alpha| \cdot d(x,y)$.\\

Now I will check the following distance measures for satisfying the desired features:
\begin{itemize}
 \item Does the measure define a metric or at least a pre-metric? \\
 Pre-metric means that it generates a topology on the space of probability distributions.
 \item Is it easy to calculate analytically or is there an efficient numerical calculation?
 \item Is this a measure that is appropriate to measure distances between probability density functions?\\
\end{itemize}

\subsubsection{$L^p$  Norm (especially $L^2$ )}

\begin{defn}
The \textit{Euclidean distance} between two points $x$ and $y$ in the n-dimensio-nal space $\mathbb{R}^n$ is defined as:
$$d_{\mathrm{E}}(x,y) = \sqrt{(x_1-y_1)^2+ \cdots + (x_n-y_n)^2}$$ and $d_{\mathrm{E}}(x,0) = \| x \|_2 = \sqrt{x_1^2+ \cdots + x_n^2} = \sqrt{x^Tx}$ is the \textit{Euclidean norm} (also called 2-norm) of $x$.
\end{defn}
\begin{defn}
If $p$ is a real number, $p \geq 1$, define the $L^p$ norm and $L^p$ distance of $x \in \mathbb{R}^n$ by:
$$\| x\|_p=\left(|x_1|^p+|x_2|^p+\cdots+|x_n|^p\right)^{1/p}$$ $$d_{\mathrm{L}^p}(x,y)=\left(|x_1-y_1|^p+|x_2-x_2|^p+\cdots+|x_n-y_n|^p\right)^{1/p}$$
(while the $L^2$ norm is the familiar \textit{Euclidean norm}, the distance in the $L^1$ norm is known as the \textit{Manhattan distance} or \textit{taxicab norm}). 
\end{defn}

One extends this to $p = \infty$ via $$\| x\|_\infty =\max \left\{|x_1|, |x_2|, \ldots, |x_n|\right\}$$ which is in fact the limit of the $p$ norms for finite $p$. The $L^{\infty}$ norm is also known as the \textit{maximum norm}. \\

It turns out that for all $p \geq 1$ this definition indeed satisfies the following characteristics $\forall x,\,y\in\R^n$ (for $0 < p <1$ the triangle inequality is violated): 
\begin{itemize}
 \item $d_{\mathrm{L}^p}(x,y)=d_{\mathrm{L}^p}(y,x)$ (symmetry)
\item $d_{\mathrm{L}^p}(x,y) \leq d_{\mathrm{L}^p}(x,z) + d_{\mathrm{L}^p}(z,y)$ (triangle inequality)
\item $d_{\mathrm{L}^p}(x,y) \geq 0$ and $d_{\mathrm{L}^p}(x,y) = 0 \Longleftrightarrow x=y$ (non-negativity and identity of indiscernibles)
\item The length of the vector is positive homogeneous with respect to multiplication by a scalar.
\end{itemize}

Furthermore the $L^p$ norm is easy to calculate analytically provided that the integrand is easy to calculate.\\

To define the $L^p$ norm of a function, in our case the density function of a random variable, let $1 \leq p < \infty$ and $(\Omega, \Sigma, \mu)$ be a measure space. Consider the set of all measurable functions from $\Omega$ to $\mathbb{R}$ whose absolute value raised to the $p$-th power has finite integral, i.e. $\|f\|_p := \left({\int |f|^p\;\mathrm{d}\mu}\right)^{1/p}<\infty$. 
Thus we define the $L^p$ norm for the difference of two random variables $X$ and $Y$ with the densities $pdf_X$ and $pdf_Y$ as follows: 
$$d_{\mathrm{L}^p}(pdf_X,pdf_Y)=\left({\int_\Omega |pdf_X(q)-pdf_Y(q)|^p\;\mathrm{d}q}\right)^{1/p}$$

It is well known that the $L^p$ norm is distance preserving under rigid motions.\\


\subsubsection{Mahalanobis Distance}


\begin{defn}
The Mahalanobis distance of a vector $x \in \mathbb{R}^n$  from a set of points with mean $\mu$ and covariance matrix $\Sigma$ is defined as: $$d_{\mathrm{M}}(x) = \sqrt{(x - \mu)^T \Sigma ^{-1} (x-\mu)}$$ 
The Mahalanobis distance with respect to the covariance matrix $\Sigma$ between two $n$-dimensional points $x$ and $y$ in the space $\mathbb{R}^n$ is defined as:
$$d_{\mathrm{M}}(x,y)=\sqrt{(x - y)^T \Sigma ^{-1} (x-y)}$$
\end{defn}
Moreover the Mahalanobis norm of $x$ is $d_{\mathrm{M}}(x,0) = \| x \|_{\mathrm{M}} := \sqrt{x^T \Sigma ^{-1} x}$. \\

Characteristics of this distance are:
\begin{itemize}
 \item $d_{\mathrm{M}}(x,y)=d_{\mathrm{M}}(y,x)$ (symmetry)
 \item $d_{\mathrm{M}}(x,y) \leq d_{\mathrm{M}}(x,z) + d_{\mathrm{M}}(z,y)$ (triangle inequality)
 \item $d_{\mathrm{M}}(x,y) \geq 0$ and $d_{\mathrm{M}}(x,y) = 0 \Longleftrightarrow x=y$ (non-negativity and identity of indiscernibles)
\end{itemize}
This shows that the Mahalanobis distance is a metric and the analytical calculation is easy.\\

\begin{defn}\label{Maha}
The Mahalanobis distance between the densities of two (multivariate) normal distributed random variables $X_1$ and $X_2$ with distributions $\mathcal{N}(\mu_1,\Sigma_1)$ and $\mathcal{N}(\mu_2,\Sigma_2)$ is defined as:
$$d_{\mathrm{M}}(X_1,X_2)=\sqrt{(\mu_1 - \mu_2)^T (\Sigma_1+\Sigma_2) ^{-1} (\mu_1-\mu_2)}$$
\end{defn}

In this case one has to take care that it suffices to have $\mu_1 = \mu_2$ to get $d_{\mathrm{M}}(X_1,X_2) = 0$ even though the distributions might be different with $\Sigma_1 \neq \Sigma_2$.\\

I just defined the Mahalanobis distance for single Gaussian kernels. For this framework one would need to expand the definition to distances of arbitrary random variables or at least to the density function of a mixture of projected Gaussians. The problem that arises hereby is that there is no Mahalanobis distance known by now for such a mixture.\\

\subsubsection{Kullback-Leibler Divergence (or Kullback-Leibler Discrimination)}

I will use the denotation \textit{Kullback-Leibler divergence} though Kullback and Leibler themselves used this term to refer to $d_{KL}(P\|Q) + d_{KL}(Q\|P)$. To me it seems that the denotation divergence is the most common one.\\


An informal motivation for Kullback-Leibler (KL) divergence is given in \cite{Runnalls}:\\
Imagine we can draw independent samples $x_1,\, x_2, \ldots$ which we assume to be either from the probability density function $p(x)$ or from $q(x)$. Now we wish to decide which density is the correct one. An approach might be to continue drawing samples until the likelihood ratio $\prod_i\frac{p(x_i)}{q(x_i)}$ exceeds some predefined threshold $G := 100:1$ in favor on one candidate or the other. Equivalently, we could aim to achieve a sample large enough that the logarithm of the likelihood ratio falls outside the bounds $\pm \log(100)$. We don't know where the data stream actually is coming from, but we suppose it to be from $p(x)$. Then the expected value of the log-likelihood-ratio for a single sample is $\mathrm{E}[\log(\frac{p (x)}{q(x)})]$ what defines the KL divergence. Thus the expected log-likelihood-ratio for the full sample will exceed $\log(100)$ if the sample size becomes larger than $\frac{\log (100)}{\mathrm{E}[\log(\frac{p (x)}{q(x)})]}$.\\ 

The KL divergence is a non-symmetric measure as explained in \cite{KLDivergence} of the difference between two probability distributions $P$ and $Q$. Typically $P$ represents the 'true' distribution of data, $Q$ typically represents a theory, model, description, or approximation of $P$.\\

\begin{defn}
For probability distributions $P$ and $Q$ of a \textit{discrete} random variable the KL divergence is defined to be:
$$d_{\mathrm{KL}}(P\|Q) = \sum_i P(i) \log \frac{P(i)}{Q(i)}$$
($P>0,Q>0 \forall i$)
\end{defn}
\begin{defn}
For distributions $P$ and $Q$ of a \textit{continuous} random variable on $\R$ the KL divergence is defined to be the integral:
$$d_{\mathrm{KL}}(P\|Q) = \int_{-\infty}^\infty p(x) \log \frac{p(x)}{q(x)} \; \mathrm{d}x$$ 
where $p$ and $q$ denote the densities of $P$ and $Q$.
\end{defn}
More generally, if $P$ and $Q$ are probability measures over a set $X$, and $Q$ is absolutely continuous with respect to $P$, then the KL divergence from $P$ to $Q$ is given by:
$$d_{\mathrm{KL}}(P\|Q) = -\int_X \log \frac{d Q}{d P} \; \mathrm{d}P$$
where $\frac{dQ}{dP}$ is the Radon-Nikodym derivative of $Q$ with respect to $P$, and provided the expression on the right-hand side exists. $\mathrm{d}P$ and $\mathrm{d}Q$ are the densities of the measures $P$ and $Q$.\\

A definition of the Radon-Nikodym derivative can be given by the following:
\begin{defn}
Let $\nu$ be a $\sigma$-finite measure on $(X,\Sigma)$ that is absolutely continuous with respect to a $\sigma$-finite measure $\mu$ on $(X,\Sigma)$. Then it holds that $\exists f: X \longrightarrow (0,\infty)$ measurable such that:
$$\nu(A) = \int_A f \mathrm{d}\mu$$
$f$ is usually written as $\frac{\mathrm{d}\nu}{\mathrm{d}\mu}$ and is called the \textbf{Radon-Nikodym derivative}.
\end{defn}

Unfortunately the KL divergence is not a metric. It is non-symmetric, i.e. $d_{\mathrm{KL}}(P\|Q) \neq d_{\mathrm{KL}}(Q\|P)$ and it does not satisfy the triangle inequality. At least it is a pre-metric:\\
If $\{P_i\}_{i = 1}^n$ is a sequence of distributions such that $\lim_{n \rightarrow \infty} d_{\mathrm{KL}}(P_n\|Q) = 0$ then one says $P_n \longrightarrow Q$.\\
In the discrete case the KL divergence further has the property to be non-negative, i.e. $d_{\mathrm{KL}}(P\|Q) \geq 0$ and $d_{\mathrm{KL}}(P\|Q) = 0 \Longleftrightarrow P=Q$.\\

Kullback and Leibler themselves defined a symmetric version of the divergence: 
$$d_{\mathrm{sKL}}(P,Q) := d_{\mathrm{KL}}(P\|Q) + d_{\mathrm{KL}}(Q\|P)$$ 
It is symmetric and nonnegative but still does not satisfy the triangle inequality. Further the symmetrized version of the KL divergence can be calculated without big computational effort.\\


\subsubsection{Convergence in Measure}

Let $E$ be a set of finite measure and $E_n(\varepsilon) = \{x \in X : \mid f_n(x)-f(x)\mid \geq \varepsilon \}$ a set where the values of $f_n$ are at least $\varepsilon$ away from $f$. $f$ and $f_n$ $(n \in \mathbb{N})$ are real valued measurable functions on $E$.
As almost everywhere (a.e.) convergence is the weakened version of point wise convergence, one can say that $\{f_n\}_n$ converges a.e. to $f$ if and only if $$\lim_{n \rightarrow \infty} \mu\left(E \cap \bigcup_{m=n}^\infty E_n(\varepsilon)\right) =0$$ for every $\varepsilon > 0$.\\
Hence convergence in measure over a set of finite measure is equal to a.e. convergence over sets of finite measure. In general this is not true.\\

\begin{defn}
Let $f, f_n\ (n \in \mathbb N): X \to \mathbb R$ be measurable functions on a measure space $(X,\Sigma,\mu)$. The sequence $\{f_n\}$ is said to \textit{converge globally} in measure to $f$ if for every $\varepsilon > 0$:
$$\lim_{n\to\infty} \mu(\{x \in X: |f(x)-f_n(x)|\geq \varepsilon\}) = 0$$
and to \textit{converge locally} in measure to $f$ if for every $\varepsilon > 0$ and every $F \in \Sigma$ with $\mu (F) < \infty$:
$$\lim_{n\to\infty} \mu(\{x \in F: |f(x)-f_n(x)|\geq \varepsilon\}) = 0$$
\end{defn}

There is no metric which includes this sense of convergence, i.e. there are no such properties like triangle inequality for convergence in measure. At least a kind of Cauchy criterion can be defined.\\
A sequence of functions is called \textit{Cauchy in measure} if for every $\varepsilon > 0$: $$\mu \left( \{x\in X :|f_m(x)-f_n(x)| \geq \varepsilon \} \right) \longrightarrow 0$$ for $n,m \longrightarrow \infty, n,m \in \mathbb{N}$.\\


\subsubsection{Mutual Information}

%

\textit{What does mutual information mean intuitively?}\\
It measures the information that the random variable $X$ and $Y$ share. Suppose I know something about one of these variables. How much reduces this the uncertainty about the other?\\
For example, if $X$ and $Y$ are independent, then knowing $X$ does not give any information about $Y$ and vice versa. That means their mutual information is zero. On the other extreme, if $X$ and $Y$ are identical then all information known about $X$ is shared with $Y$ completely. In this case the mutual information is the same as the entropy of any of the random variables.\\

\begin{defn}
The entropy $H$ of a discrete random variable $X$ is defined by:
$$H(X) = \mathrm{E}[I(X)]$$
with $\mathrm{E}$ the expected value, and $I(X)$ the information content or self-information of $X$.
\end{defn}
The information content of an event $x$  with probability $\mathrm{P}(x)$ is given  by $I(x) = -\log(\mathrm{P}(x))$ and hence $I(X)$ is a random variable. If $p$ denotes the probability mass function of $X$ then the entropy can explicitly be written as:
$$H(X) = \sum_{i=1}^n {p(x_i) I(x_i)} = -\sum_{i=1}^n {p(x_i) \log p(x_i)}$$

\begin{defn}
The mutual information, also transinformation, of two \textit{discrete} random variables $X$ and $Y$ can be defined as:
$$I(X,Y) = \sum_{y \in Y} \sum_{x \in X} p(x,y) \log{ \left( \frac{p(x,y)}{p_1(x) p_2(y)} \right) }$$
where $p(x,y)$ is the joint probability distribution function of $X$ and $Y$, and $p_1(x)$ and $p_2(y)$ are the marginal probability distribution functions of $X$ and $Y$ respectively.
\end{defn}
\begin{defn}
In the \textit{continuous} case, the mutual information of $X$ and $Y$ can be defined as:
$$I(X,Y) = \int_Y \int_X p(x,y) \log{ \left( \frac{p(x,y)}{p_1(x) p_2(y)} \right) }\mathrm{d}x \mathrm{d}y$$
where $p(x,y)$ is the joint probability density function of $X$ and $Y$, and $p_1(x)$ and $p_2(y)$ are the marginal probability density functions of $X$ and $Y$ respectively.\\
\end{defn}

As the base of the log function is not specified, these definitions are ambiguous. To change these functions to become unique, the function $I$ could be parameterized as $I(X,Y,b)$ with $b$ the base. An alternative would be to specify the base to be 2, since one bit is the most common unit of measure of mutual information.\\

\textit{Characteristics of the mutual information:}
\begin{itemize}
 \item Mutual information can be expressed as a Kullback-Leibler divergence:
$$I(X,Y) = d_{\mathrm{KL}}(p(x,y)\|p(x)p(y))$$
where $p(x,y)$ is the joint distribution of the random variables $X$ and $Y$.
 \item It is a measure of dependence in the following sense: \\
$I(X, Y) = 0$ if and only if $X$ and $Y$ are independent random variables.
 \item The measure is non-negative, i.e. $I(X,Y) \geq 0$ for all random variables $X$ and $Y$ and symmetric, i.e. $I(X,Y) = I(Y,X)$.
 \item Even though mutual information does not define a metric, $d(X,Y) = H(X,Y) - I(X,Y)$ does, where $H(X,Y)$ is the joint entropy of $X$ and $Y$.
\end{itemize}

\begin{figure}[htp]
  \begin{center}
    \subfigure[$I(X_1,X_2) = k$]{\label{fig:edge-a}\includegraphics[scale=0.36]{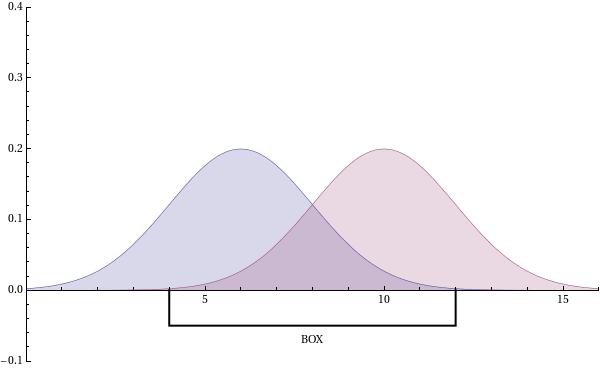}}
    \subfigure[$I(X_1,X_2) = n<k$]{\label{fig:edge-b}\includegraphics[scale=0.36]{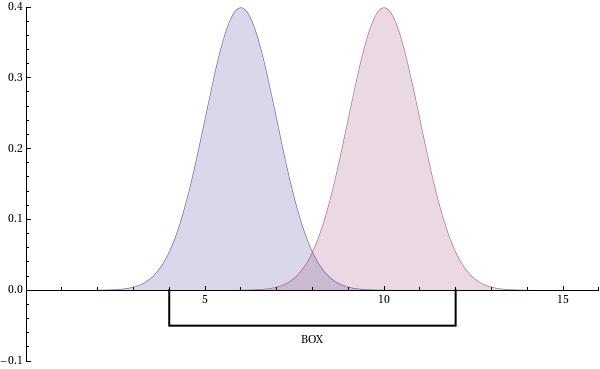}} \\
  \end{center}
  \caption[Mutual information]{}
  \label{fig:edge}
\end{figure}

In this framework we will need to evaluate distances between more than two random variables. Thus one of the various extensions of mutual information has to be chosen. The most established extensions are the conditional mutual information and interaction information.\\
But instead of going more to detail, 
I want to mention, that this measure doesn't fit the problem properly about the ability to grasp.\\
Imagine a robot estimates the pose of a target object at least two times and receives the random variable $X_1$ and $X_2$ with the probability distributions of the base elements $\mathcal{N}(TS_1,\mu_1,\Sigma_1)$ and $\mathcal{N}(TS_2,\mu_2,\Sigma_2)$ with densities $p_1$ and $p_2$. In the case that these Gaussians are far away from each other and thus have a small overlap, the mutual information $I(X_1,X_2) = \int_{X_2} \int_{X_1} p(x_1,x_2) \log{ \left( \frac{p(x_1,x_2)}{p_1(x_1) p_2(x_2)} \right) }\mathrm{d}x_1 \mathrm{d}x_2$ gives the right indication on becoming very small that at least one of the measures is unusable and another measure is required to receive a proper pose estimation.\\
For the case that the Gaussian kernels are close enough for the gripper being able to encompass a high percentage of both of them, we would like the distance measure to show this.  For instance in figure \ref{fig:edge-a} $84\%$ of the probability mass of both of the distributions is inside the box, the robots hand can grasp. Now recall that the more peaked the kernels are, the higher the proportion of the distribution, that is enclosed in the box. This means in case of strongly peaked density functions like an approximation of the Dirac delta function $\delta_{\varepsilon}(x)=\frac{1}{\sqrt{2\pi\varepsilon}}\,\mathrm{e}^{-{x^{2}}/{(2 \varepsilon)}}$ where $\varepsilon>0$, one can be almost sure that the object has its pose somewhere in the box, that contains the probability mass of both of the Gaussians. In the figure \ref{fig:edge-b} below $97,8\%$ of the probability mass is inside the box. \\

Unfortunately the mutual information just shows whether the kernels have big or small overlap, and thus in the case of peaked Gaussian kernels it gives a small value. That's why I arrived at the conclusion that this distance measure is not suitable for our problem.\\


\subsection{Convergence Measures}

This is an attempt to grade the different convergences in order of strength.\\
The properties of sequences of functions (or random variables) can vary a lot for growing indices. Hence one needs quite different kinds of convergences, which usually are with respect to various norms or topologies even though there are sometimes other kinds of convergences like convergence in measure as well. \\

The classical types of convergence are the
\begin{itemize}
\item \textbf{pointwise convergence:}\\
 Let $f_n$ be a sequence of functions on the same domain $D$. One says $f_n$ converges pointwise to the limit $f$ if $f(x) = \lim_{n \to \infty} f_n(x)$
\item and the \textbf{uniform convergence:}\\
A sequence $f_n$ converges uniform to $f$ if maximal differences between $f_n$ and $f$ converge to zero. This is a kind of convergence in terms of the maximum norm.\\
The limit function $f$ has the property that if the sequence is continuous, then the uniform limit also is continuous. Furthermore it holds that the integral of the uniform limit is the limit of the integral of the sequence, i.e. $\lim_{n\to\infty}\int_a^b f_n \mathrm{d}x=\int_a^b f\mathrm{d}x$ and the derivative of the uniform limit is the limit of the derivative of the sequence, i.e. $\lim_{n\to\infty}f_n'=f'$.
\end{itemize}

In measure theory these types of convergence usually are unambiguous and thus one can only define the convergence \textit{almost everywhere}.
\begin{itemize}
\item \textbf{pointwise convergence almost everywhere (a.e.)}\\
The convergence is not true at the most on a set with zero measure.
\item \textbf{convergence in measure}\\
If a sequence converges almost everywhere in a space with finite measure $\mu(\Omega) < \infty$ then it converges in measure. Thus convergence in measure is weaker than convergence a.e.
\item  \textbf{$L^p$ convergence}\\
A sequence converges in $L^p$ if \\
$\lim_{n\to\infty} \| f_n-f \|_p = \lim_{n\to\infty} \left( \int_\Omega \| f_n(x)-f(x) \|^p\, \mathrm d \mu(x) \right)^{1/p} = 0$. Hence from $L^p$ convergence follows convergence in measure.
\item  \textbf{almost uniform convergence}\\
A sequence converges almost uniform if $\forall \, \varepsilon >0\; \exists\, A \in \Sigma: \mu(A)<\varepsilon$ and the sequence converges uniformly on $\Sigma \backslash A$. 
\item \textbf{convergence in probability (weak convergence)}\\
It is related to convergence in measure. There are several equivalent definitions of weak convergence of a sequence of measures (see Portmanteau).
\end{itemize}

There are two hierarchies of convergences $f_n \to f$ in spaces with finite measure $\mu(\Sigma)<\infty$:\\
uniform convergence \\
$\Rightarrow$ pointwise convergence \\
$\Rightarrow$ pointwise convergence almost everywhere $\Leftrightarrow$ almost uniform convergence\\
$\Rightarrow$ convergence in measure \\

uniform convergence \\
$\Rightarrow$ $L^\infty$ convergence\\
$\Rightarrow$ $L^p$ convergence, for all real $0<p<\infty$ \\
$\Rightarrow$ convergence in measure \\
$\Rightarrow$ convergence in probability\\


\section{Behavior and Properties of Approximations of MoPGs}\label{prop}

On approximating mixtures of projected Gaussians one surely wants to know what properties remain. There are various theoretical questions to answer:\\
\begin{itemize}
 \item \textit{What kinds of properties are passed on from the original $\mathrm{MoPG}$s to the new $\mathrm{MoPG}$, one obtains after applying one of modification operations fusing, merging or composing to it?} 
 \item \textit{What do we know about the accuracy of the approximation?\\
 Let's assume to have the $pdf$ of a known $\mathrm{MoPG}$ that fulfills the grasp criterion with a certain level for failure $\varepsilon>0$. On approximating the $pdf$ we want to preserve that the resulting $pdf$ still fulfills the criterion.}
\end{itemize}

About the approximation and simplification step of the mixture: 
\begin{itemize}
 \item \textit{Which kernels can be omitted? Can an upper bound be given for the error that arises from dropping kernels? Which threshold for the weights is appropriate to discard the ones below?}
 \item \textit{Can various kernels be replaced by a single one, if they have similar mean and covariance? And what means 'similar'? Is there a kind of $pdf$ that might replace kernels with equal mean, but different covariances?}
 \item \textit{How many kernels are required and are reasonable to model different kinds of distributions like an identical distribution for instance with a mixture of projected Gaussians?}\\
\end{itemize}

\subsection{Error Estimation on Dropping Base Elements of a Mixture} \label{omit}

Let $M\in\mathrm{MoPG}$ be a mixture of projected Gaussians with the density $p_M(x)=\sum_{i=1}^n \lambda_i\cdot p(\mu_i,\Sigma_i,x)$. Lets denote $p_i = p(\mu_i,\Sigma_i,x)$. Now we want to approximate $M$ by the mixture $M_{app}\in \mathrm{MoPG}$. This approximated mixture $M_{app}$ can easily be achieved by discarding the less relevant base element with smallest weight $\lambda_{i_0}$. For easier notation renumber the weights $\lambda_i$ and densities $p_i$ such that $\lambda_{i_0}$ becomes the last one. As we know that the $\lambda_i$ have to sum to $1 = \sum_{i=1}^n\lambda_i$, we can renormalize the remaining weights by: 
$$\lambda'_i:=\lambda_i+\frac{\lambda_i\lambda_n}{1-\lambda_n}=\frac{\lambda_i}{1-\lambda_n}$$
for all $i\in \{1,\dots,n-1\}$.\\

Now an upper bound for the error of the approximation can be given. I will calculate the total variance which is the maximal error that can occur on trying to grasp an object by using the box criterion. Remember that this uses the $\arg\max$ of an integral over the probability density enclosed in any box $B$ of the set of boxes $\mathbb{B}$. 

\begin{flalign}
\left|\mathrm{P}(B)-\mathrm{P}_{app}(B)\right| &= \left|\int\limits_{B} \sum_{i=1}^n \lambda_i p_i \,\mathrm{d}\mu_B - \int\limits_{B} \sum_{i=1}^{n-1}\lambda'_ip_i \,\mathrm{d}\mu_B \right|\nonumber\\
&= \left|\int\limits_{B} \sum_{i=1}^{n-1} (\lambda_i-\lambda'_i)p_i +\lambda_n p_n  \,\mathrm{d}\mu_B\right|\nonumber\\
&\leq\left|\int\limits_{B} \sum_{i=1}^{n-1} \left(-\frac{\lambda'_i\lambda_n}{1-\lambda_n} \right) p_i \,\mathrm{d}\mu_B\right| + \lambda_n \underbrace{\int\limits_{B}p_n \,\mathrm{d}\mu_B}_{\leq 1}\nonumber\\
&\leq\int\limits_{B} \left(\frac{\lambda_n}{1-\lambda_n} \right) \sum_{i=1}^{n-1} \lambda'_i p_i \,\mathrm{d}\mu_B +\lambda_n\nonumber\\
&=\frac{\lambda_n}{1-\lambda_n}\underbrace{\sum_{i=1}^{n-1}\lambda'_i}_{1-\lambda_n} \underbrace{\int\limits_{B}p_i \,\mathrm{d}\mu_B}_{\leq 1\  \forall \, i} + \lambda_n\nonumber\\
&\leq 2\lambda_n \qquad\quad \forall \, B \in \mathbb{B} \nonumber
\end{flalign}

Of course the approximation $\left|\int\limits_{B}p_i \,\mathrm{d}\mu_B\right|\leq 1$ is very rough but it suffices to show that the difference between the approximated and the original probability at most is $2\lambda_{i_0}$.\\

\subsection{Fusing MoPGs}\label{fuse}

Robots commonly have a stereo system of cameras and make several localization attempts of the target object from different points of view. Thus in general a couple of mixtures are obtained that seem to be reliable and all describe the same object. If we believe in these observations, we need to think of a solution to join the information we get from the single mixtures. \\
Our approach to utilize all the information, is to \textit{fuse the mixtures} in order to receive the best possible probability distribution for the pose.\\

Let $M_1,\, M_2 \in \mathrm{MoPG}$ be two mixtures of projected Gaussians with densities $p_{M_1}$ and $p_{M_2}$. To obtain the base elements of the fused mixture $p_{M_3} = fuse(p_{M_1},p_{M_2})$ each of the base elements of $M_1$ has to be fused with all of the base elements of $M_2$. How this fusion works is briefly introduced in \cite{6D} and I explained it in section \ref{inferences}.\\
It doesn't make sense to fuse widely separated base elements as then a systematic overestimation of the concentration of the covariance matrix results. Thus we require all covariance matrices of the mixtures to be sufficiently well peaked. There are further things that shall be payed attention to:
\begin{itemize}
\item The tangent points $q_i$ and $q_j$ of the the base elements $\mathrm{PG}_i \in M_1$ and $\mathrm{PG}_j \in M_2$ to be fused need to be sufficiently close. In practice it turned out to make no sense to allow a bigger angle than $15^\circ$ between the point $q_i$ and $q_j$ on the hypersphere $S_3$. To assure that the base elements are compatible a weighting factor $$\alpha_{ij} = \mathrm{e}^{-5\cdot \arccos ((q_i\cdot q_j)^2)}$$ is introduced. The angle $\theta$ between $q_i$ and $q_j$ can be calculated with $\theta = \arccos(q_i\cdot q_j)$. By taking the square of the scalar product of the tangent points $(q_i\cdot q_j)^2$ it is secured that the exponent of the function and thus also the function is antipodal symmetric on the sphere. The factor $-5$ was obtained by heuristics and has the effect that the whole function goes to 0 reasonably quick.
\item If both base elements $\mathrm{PG}_i$ and $\mathrm{PG}_j$ shall be applied at the same moment, the dissimilarity of the distribution functions has to be small as well. Note that I require the base elements to be projected to the same tangent space already. We use the Mahalanobis distance to weight the fused base element $\mathrm{PG}_{i,j}$:
$$\delta_{ij} = \mathrm{e}^{-1/2\cdot (\mu_i-\mu_j)(\Sigma_i+\Sigma_j)^{-1}(\mu_i-\mu_j)^\top}$$
This expresses that even if the base elements share the same tangent space, they could be incompatible because the distributions might be too different. A disadvantage of this weighting function is that for base elements that by accident have the same mean $\mu_i = \mu_j$ the maximal weight is returned. 
\end{itemize}
All together each of the summands of the fused mixture has the form:
$$C\cdot \lambda_i \lambda_j \alpha_{ij} \delta_{ij}\cdot \mathrm{PG}_{ij}$$
where $C = 1/(\sum_{i = 1}^{n_1}\sum_{j=1}^{n_2} \lambda_i \lambda_j \alpha_{ij} \delta_{ij})$ is the normalizing constant for $p_{M_3}$ to be a probability density function, $\lambda_i$ and $\lambda_j$ are the weights of the base elements in their original mixtures and $\mathrm{PG}_{ij}$ is the fused base element with density $p_{M_3}$.\\


The problem that arises with this approach to fuse mixtures is that the resulting mixture $M_3$ consists of $n_1\cdot n_2$ summands for $M_1$ consisting of $n_1$ and $M_2$ consisting of $n_2$ elements. To reduce this rapidly increasing number of base elements, kernels with low weight can be omitted as I mentioned already in \ref{omit}. Another strategy is to merge similar kernels what I will explain in the following.\\

\subsection{Merging MoPGs}\label{merge}

The moment-preserving merge \cite{Robert} is a common procedure to substitute two elements of a MoPG by a new one, matching the zeroth, first and second-order moments of the original mixture. I described this merge already in section \ref{inferences}.\\
The more interesting point of merging elements of a mixture is the best choice of the projected Gaussians. The Mahalanobis distance of two Gaussian kernels, I introduced in definition \ref{Maha} might match the problem to check the compatibility of single kernels, but for this topic I have to chose another dissimilarity measure that fits whole mixtures of projected Gaussians. Actually I don't need to know \textit{which kernels of the mixture are the most similar}, I want to find the kernels such that after merging them, \textit{the whole approximated mixture is the least dissimilar from the mixture before the merge.}\\


Williams searches in his master's thesis \cite{GaussianMixtureReduction} for a scalar cost function which measures the difference between the density of the original mixture and the approximated mixture in order to evaluate whether one merge is 'better' than another. He proposes to use the square of the $L^2$ norm
$$d_{ISD}(f_1,f_2)=\int (f_1(x)-f_2(x))^2\;\mathrm{d}x$$
which he refers to as integral square difference measure (ISD).\\
Further he introduces the Kolmogorov variational distance $d_K(f_1,f_2) = \int |f_1(x)-f_2(x)|\;\mathrm{d}x$ which has an intuitively appealing probability mass interpretation and the Maximum Likelihood measure $d_{ML}(f_1,f_2) = \int f_1(x)log(f_2(x))\;\mathrm{d}x$ which would fit the requirements of a cost function the best. But only the ISD measure provides the advantage to be computable in closed form. Anyway the Williams criterion is disadvantageous for the reason that the optimization often finds local minima and Runnalls constructed an example that showed the scale-dependency of the ISD cost measure which also can lead to anomalies.\\
Salmond with his criterion reduces the number of components by repeatedly choosing the two most similar components and merging them.
The similarity is derived from a statistical analysis of the variance. 
For any two mixture elements $G_i = \lambda_i\cdot\mathcal{N}(\mu_i,\Sigma_i)$ and $G_j = \lambda_j\cdot\mathcal{N}(\mu_j,\Sigma_j)$ the dissimilarity measure proposed by Salmond is defined as:
$$d_S(G_i,G_j) = \mathrm{tr}(\Sigma^{-1} \frac{\lambda_i\lambda_j}{\lambda_i+\lambda_j}(\mu_i-\mu_j)(\mu_i-\mu_j)^\top)$$
where $\Sigma$ is the 'overall variance' of the mixture, $\Sigma = \sum_{i=1}^n\lambda_i\Sigma_i + \sum_{i=1}^n \lambda_i(\mu_i-\mu)(\mu_i-\mu)^\top$ and $\mu = \sum_{i=1}^n\lambda_i\mu_i$ is the 'overall mean' of the mixture.\\
Major drawbacks are that the measure just depends on the means of the components, not on their individual covariances and that adding a new component might alter the merge order of existing components. Thus in various cases unfavored behavior of the merging algorithm arises.\\

A more promising criterion is the dissimilarity measure based on Kullback-Leibler (KL) divergence. I developed a variant based on the symmetrized version of the Kl divergence which I will refer to as sKL divergence.\\

From Runnalls \cite{Runnalls} paper we know that the following holds:
\begin{theo}\label{thr}
 Let $p_1(x)$ be the density of a d-dimensional Gaussian distribution $\mathcal{N}(\mu_1,\Sigma_1)$ and $p_2(x)$ be the density of a d-dimensional distribution $\mathcal{N}(\mu_2,\Sigma_2)$.\\
 Then:
 $$2d_{KL}(p_1,p_2) = \mathrm{tr}\left(\Sigma_2^{-1}(\Sigma_1-\Sigma_2+(\mu_1-\mu_2)(\mu_1-\mu_2)^\top)\right)+\log\frac{\det(\Sigma_2)}{\det(\Sigma_1)}$$
\end{theo}

This implies:
$$\begin{array}{lcl}
d_{sKL}(p_1,p_2) &=& \frac{1}{2}\mathrm{tr}\left(\Sigma_2^{-1}(\Sigma_1-\Sigma_2+(\mu_1-\mu_2)(\mu_1-\mu_2)^\top)\right)\\ 
&+& \frac{1}{2}\mathrm{tr}\left(\Sigma_1^{-1}(\Sigma_2-\Sigma_1+(\mu_2-\mu_1)(\mu_2-\mu_1)^\top)\right)\\
&=& \frac{1}{2}\mathrm{tr}\left(\Sigma_2^{-1}\Sigma_1+\Sigma_1^{-1}\Sigma_2+(\Sigma_1^{-1}+\Sigma_2^{-1})(\mu_1-\mu_2)(\mu_1-\mu_2)^\top\right)-d
\end{array}$$
which can be calculated much faster as the logarithm cancels out.\\

Regrettably there is no closed form expression neither for the KL divergence of two mixtures of projected Gaussians, nor for the sKL divergence of two mixtures of projected Gaussians, as the mixture density $p_M = \sum_{i=1}^n \lambda_i\, p_i$ consists of a sum, where the $p_i$s are the densities of the single projected Gaussian kernels.\\
For this reason Runnalls thought of an upper bound of the KL divergence between the mixture before the merge and the mixture after the merge of two similar Gaussian kernels. He denominated this upper bound $B(i,j)$. Analogously I will refer to my upper bound of the \textit{symmetrized KL divergence} as $B_s(i,j)$, which I will derive now.\\

\begin{theo}\label{th1}
 If $f_1(x)$, $f_2(x)$ and $h(x)$ are any pdfs over d dimensions, $0\leq \omega\leq1$ and writing $\oomega$ for $1-\omega$, then:
 $$d_{sKL}(\omega f_1+\oomega h, \omega f_2+\oomega h)\leq \omega\, d_{sKL}(f_1,f_2)$$\\
\end{theo}

\textit{Proof}:
\begin{flalign}
\omega d_{sKL}(f_1,f_2)&- d_{sKL}(\omega f_1+\oomega h, \omega f_2+\oomega h)= \nonumber\\
                &= \omega \int_{\R^d}f_1 \log\frac{f_1}{f_2}\mathrm{d}x+ \omega \int_{\R^d}f_2 \log\frac{f_2}{f_1}\mathrm{d}x\nonumber\\
		&- \int_{\R^d}(\omega f_1+\oomega h) \log\frac{\omega f_1+ \oomega h}{\omega f_2+ \oomega h}\mathrm{d}x
		-\int_{\R^d}(\omega f_2+\oomega h) \log\frac{\omega f_2+ \oomega h}{\omega f_1+ \oomega h}\mathrm{d}x\nonumber\\
		&=\omega\int_{\R^d}f_1 \log\frac{f_1(\omega f_2+ \oomega h)}{f_2(\omega f_1+ \oomega h)}\mathrm{d}x
		+\omega\int_{\R^d}f_2 \log\frac{f_2(\omega f_1+ \oomega h)}{f_1(\omega f_2+ \oomega h)}\mathrm{d}x -0\nonumber\\
		&\stackrel{\ast}\geq \omega\int_{\R^d}f_1 \left(1-\frac{f_2(\omega f_1+ \oomega h)}{f_1(\omega f_2+ \oomega h)}\right)\mathrm{d}x
		+ \omega\int_{\R^d}f_2\left(1-\frac{f_1(\omega f_2+ \oomega h)}{f_2(\omega f_1+ \oomega h)}\right)\mathrm{d}x\nonumber\\
		&=\int_{\R^d}\left(\frac{(\omega f_1-\omega f_2)\oomega h)}{\omega f_2+ \oomega h}+\frac{(\omega f_2-\omega f_1)\oomega h)}{\omega f_1+ \oomega h}\right)\mathrm{d}x\nonumber\\
		&=\int_{\R^d}\frac{(\omega f_1-\omega f_2)^2\oomega h}{(\omega f_1+ \oomega h)(\omega f_2+ \oomega h)}\mathrm{d}x\nonumber\\
		&\geq 0 \nonumber
\end{flalign}
\EndProof

That $\ast$ holds can be seen from the following:
\begin{lem}\label{lemma}
For all $a,\,b\in \Z$ it holds that: $$\log\frac{a}{b}\geq 1-\frac{b}{a}$$\\
\end{lem}

\textit{Proof}:
$$\log\frac{a}{b}\geq 1-\frac{b}{a} \quad \Longleftrightarrow\quad \frac{a}{b} \log\frac{a}{b}\geq\frac{a}{b}-1$$
Substitute $x:= \frac{a}{b}$
$$x\log x \geq x-1 \quad \Longleftrightarrow\quad 1-x + x\log x \geq0$$
Define $f:\Q \to \Q$:\\
$$\begin{array}{lcl}
 f(x) &=& 1-x+x\log x\\
   f'(x) &=& -1+\log x+1 = \log x
\end{array}$$
Now it can be seen that $f(x)$ has its global minimum at $x=1$ and $f(1) = 0$. Thus $1-x + x\log x \geq0$ is true.
\EndProof

Let $f_1$ be the density of the normalized mixture $M_{i+j}$ consisting of the two base elements $\mathrm{PG}_i$ and $\mathrm{PG}_j$, $i\neq j$, that shall be merged. If the base elements do not share the same tangent space anyway I project them to a common one $TS_{ij}$ at the tangent point $p_{ij} = \frac{p_i + p_j}{\|p_i + p_j\|}$ by the double projection: Central projection $\prod_{p_i}$ respectively $\prod_{p_j}$ to the sphere followed by the inverse of the central projection $\prod_{p_{ij}}^{-1}$ to the common tangent space. For easier notation I name these new base elements with $\mathrm{PG}_i$ and $\mathrm{PG}_j$ as well, as the postulation that they have the same tangent space, doesn't involve further changes. Hence from now on it can be assumed for the whole remainder of this section that the rotation part of $\mathrm{PG}_i$ and $\mathrm{PG}_j$ live in the same tangent space.\\
Let $f_2$ be the density of the mixture $M_{ij}$ consisting of the single kernel $\mathrm{PG}_{ij}$ that is obtained by merging the elements $\mathrm{PG}_i$ and $\mathrm{PG}_j$ with the moment-preserving merge. Note that $\mathrm{PG}_{ij}$ also has the same tangent space as $\mathrm{PG}_i$ and $\mathrm{PG}_j$. Further let $h$ be the remaining mixture except the two particular kernels of $f_1$.\\
Then we get from theorem \ref{th1} that the divergence of the whole mixture after merging the components $\mathrm{PG}_i$ and $\mathrm{PG}_j$ from the mixture before the merge will not exceed $\omega \cdot d_{sKL}(f_1,f_2)$, where $\omega = \lambda_i+\lambda_j$ and $d_{sKL}(f_1,f_2)$ is the divergence of the normalized mixture $M_{i+j}$ and the mixture of the merged single Gaussian $M_{ij}$.\\

\begin{theo}\label{th2}
 If $f(x)$, $h_1(x)$ and $h_2(x)$ are any pdfs over d dimensions, $0\leq \omega\leq1$ and writing $\oomega$ for $1-\omega$, then:
 $$d_{sKL}(\omega h_1 + \oomega h_2,f) \leq \omega\, d_{sKL}(h_1,f) + \oomega\, d_{sKL}(h_2,f)$$\\
\end{theo}

\textit{Proof}:
\begin{flalign}
\omega d_{sKL}(\omega h_1 + \oomega h_2,f) &= \int_{\R^d}(\omega h_1 + \oomega h_2) \log\frac{\omega h_1 + \oomega h_2}{f}\mathrm{d}x+ \int_{\R^d}f \log\frac{f}{\omega h_1 + \oomega h_2}\mathrm{d}x\nonumber\\
	&= \omega\int_{\R^d}h_1 \log\frac{\omega h_1 + \oomega h_2}{f}\mathrm{d}x + \oomega\int_{\R^d}h_2 \log\frac{\omega h_1 + \oomega h_2}{f}\mathrm{d}x\nonumber\\
	&+\int_{\R^d}f \log\frac{f}{\omega h_1 + \oomega h_2}\mathrm{d}x\nonumber
\end{flalign}

\begin{flalign}
\omega\, d_{sKL}(h_1,f) + \oomega\, d_{sKL}(h_2,f) &= \omega\int_{\R^d}h_1 \log\frac{h_1}{f}\mathrm{d}x+ \omega\int_{\R^d}f \log\frac{f}{h_1}\mathrm{d}x\nonumber\\
	&+ \oomega\int_{\R^d}h_2 \log\frac{h_2}{f}\mathrm{d}x+ \oomega\int_{\R^d}f \log\frac{f}{h_2}\mathrm{d}x\nonumber
\end{flalign}

If we can show that:
$$\omega\int_{\R^d}h_1 \log\frac{\omega h_1 + \oomega h_2}{f}\mathrm{d}x\leq \omega\int_{\R^d}h_1 \log\frac{h_1}{f}\mathrm{d}x$$ 
it follows directly: 
$$\oomega\int_{\R^d}h_2 \log\frac{\omega h_1 + \oomega h_2}{f}\mathrm{d}x \leq \oomega\int_{\R^d}h_2 \log\frac{h_2}{f}\mathrm{d}x$$
and just remains to check whether:
$$\int_{\R^d}f \log\frac{f}{\omega h_1 + \oomega h_2}\mathrm{d}x \leq \omega\int_{\R^d}f \log\frac{f}{h_1}\mathrm{d}x+ \oomega\int_{\R^d}f \log\frac{f}{h_2}\mathrm{d}x$$

Now calculate:
\begin{flalign}
\omega\int_{\R^d}h_1 \log\frac{h_1}{f}\mathrm{d}x&-\omega\int_{\R^d}h_1 \log\frac{\omega h_1 + \oomega h_2}{f}\mathrm{d}x = \omega\int_{\R^d}h_1 \left(\log\frac{h_1}{f}-\log\frac{\omega h_1 + \oomega h_2}{f}\right)\mathrm{d}x \nonumber\\
      &=\omega\int_{\R^d}h_1 \left(\log\frac{h_1}{\omega h_1 + \oomega h_2}\right)\mathrm{d}x \nonumber\\
      &\stackrel{\ref{lemma}}\geq \omega\int_{\R^d}h_1 \left(1-\frac{\omega h_1 + \oomega h_2}{h_1}\right)\mathrm{d}x \nonumber\\
      &= \omega\int_{\R^d}h_1 -(\omega h_1 + \oomega h_2)\,\mathrm{d}x \nonumber\\
      &= \omega\underbrace{\int_{\R^d}h_1\,\mathrm{d}x}_{=1} - \omega\underbrace{\int_{\R^d}\omega h_1 + \oomega h_2\,\mathrm{d}x}_{=1} \nonumber\\
      &=0\nonumber
\end{flalign}

\begin{flalign}
\int_{\R^d}f &\log\frac{f}{\omega h_1 + \oomega h_2}\mathrm{d}x - \omega\int_{\R^d}f \log\frac{f}{h_1}\mathrm{d}x- \oomega\int_{\R^d}f \log\frac{f}{h_2}\mathrm{d}x \nonumber\\
      &\leq \int_{\R^d}f \left(1-\frac{\omega h_1 + \oomega h_2}{f}\right)\mathrm{d}x - \omega\int_{\R^d}f \left(1-\frac{h_1}{f} \right)\mathrm{d}x- \oomega\int_{\R^d}f \left(1-\frac{h_2}{f}\right)\mathrm{d}x \nonumber\\
      &= \int_{\R^d}\left(f-(\omega h_1+\oomega h_2)-\omega f + \omega h_1-\oomega f+ \oomega h_2\right) \mathrm{d}x\nonumber\\
      &= \int_{\R^d}\left(f-(\omega +\oomega) f\right) \mathrm{d}x\nonumber\\
      &=0\nonumber
\end{flalign}
\EndProof

Now let $h_1$ be the density of $\mathrm{PG}_i$ and $h_2$ the one of $\mathrm{PG}_j$. Thus we have the normalized mixture:
$$M_{i+j} = \frac{\lambda_i}{\lambda_i+\lambda_j}\mathrm{PG}_i+ \frac{\lambda_j}{\lambda_i+\lambda_j}\mathrm{PG}_j$$
and $M_{ij} := 1\cdot \mathrm{PG}_{ij}$.\\
Then we get from theorem \ref{th2} that $d_{sKL}(f_1,f_2)$ will never raise above: 
$$\frac{1}{\lambda_i+\lambda_j}\left(\lambda_i\, d_{sKL}(h_1,f)+ \lambda_j\, d_{sKL}(h_2,f)\right)$$
What can equivalently be written in terms of mixtures and base elements:
$$d_{sKL}(M_{i+j},M_{ij})\leq \frac{1}{\lambda_i+\lambda_j}\left(\lambda_i\, d_{sKL}(\mathrm{PG}_i,\mathrm{PG}_{ij})+ \lambda_j\, d_{sKL}(\mathrm{PG}_j,\mathrm{PG}_{ij})\right)$$

Putting this together with the upper result, from theorem \ref{th1}, it follows:\\
\begin{theo}
The sKL divergence of the whole mixture of projected Gaussians $\mathrm{MoPG}$ before the merge from the mixture $\mathrm{MoPG}_{app}$ after the merge of the components $\mathrm{PG}_i$ and $\mathrm{PG}_j$ of the mixture $\mathrm{MoPG}$, $i\neq j$, will not exceed:
$$B_s(i,j) := \lambda_i\, d_{sKL}(\mathrm{PG}_i,\mathrm{PG}_{ij})+ \lambda_j\, d_{sKL}(\mathrm{PG}_j,\mathrm{PG}_{ij})$$\\
\end{theo}

This formula can still be simplified using the result from theorem \ref{thr}. Further substitute $\lambda_{ij} := \lambda_i+\lambda_j$ and $\mu_{ij} := \frac{\lambda_i}{\lambda_{ij}}\mu_i + \frac{\lambda_j}{\lambda_{ij}}\mu_j$. \\
Then the upper bound for the sKL divergence can be written as: 
\begin{flalign}
 B_s(i,j) &= \lambda_i\, d_{sKL}(\mathrm{PG}_i,\mathrm{PG}_{ij})+ \lambda_j\, d_{sKL}(\mathrm{PG}_j,\mathrm{PG}_{ij})\nonumber\\
          &= \frac{1}{2}\lambda_i\, \mathrm{tr}\left(\Sigma_{ij}^{-1}\Sigma_i+\Sigma_i^{-1}\Sigma_{ij}+(\Sigma_i^{-1}+\Sigma_{ij}^{-1})(\mu_i-\mu_{ij})(\mu_i-\mu_{ij})^\top\right)-d\,\lambda_i\nonumber\\
	  &+ \frac{1}{2}\lambda_j\,\mathrm{tr}\left(\Sigma_{ij}^{-1}\Sigma_j+\Sigma_j^{-1}\Sigma_{ij}+(\Sigma_j^{-1}+\Sigma_{ij}^{-1})(\mu_j-\mu_{ij})(\mu_j-\mu_{ij})^\top\right)-d\,\lambda_j\nonumber\\
	  &= \frac{1}{2} 
          \lambda_i\,\mathrm{tr}
\left(\Sigma_i^{-1}\Sigma_{ij}+
\left(\frac{\lambda_j}{\lambda_{ij}}\right)^2
(\Sigma_i^{-1}+\Sigma_{ij}^{-1})(\mu_i-\mu_j)(\mu_i-\mu_j)^\top+
\left(\Sigma_i^{-1}\Sigma_{ij}\right)^{-1}\right)\nonumber\\
  &+ \frac{1}{2}\lambda_j\,\mathrm{tr}
\left(\Sigma_j^{-1}\Sigma_{ij}+
\left(\frac{\lambda_i}{\lambda_{ij}}\right)^2 
(\Sigma_j^{-1}+\Sigma_{ij}^{-1})(\mu_i-\mu_j)(\mu_i-\mu_j)^\top+
\left(\Sigma_j^{-1}\Sigma_{ij}\right)^{-1}\right) \nonumber\\
&- d\,\lambda_{ij}\nonumber
\end{flalign}

where $\Sigma_{ij} := \frac{1}{\lambda_{ij}}\cdot \left(\lambda_i\Sigma_i+\lambda_j\Sigma_j+\lambda_i\, \lambda_j(\mu_i-\mu_j)(\mu_i-\mu_j)^\top\right)$ and $d$ is the dimension of the base elements.\\
This means $B_s(i,j)$ can be calculated directly from the densities of the base elements $\mathrm{PG}_i$ and $\mathrm{PG}_j$.\\

\subsection{Formulation of Conjectures}

For the whole section let $g$ be the density of the mixture of projected Gaussians that describes the pose of the robots gripper in the $SE(3)$. Further let $p$, $p_\infty$ respectively $p_{app}$ be the densities of the mixtures of projected Gaussians $M\in \mathrm{MoPG}$, $M_\infty \in \mathrm{MoPG}$  respectively $M_{app} \in \mathrm{MoPG}$. $M$ is any mixture that estimates the pose of the target object. $M_\infty$ is the infinitely long mixture that just exists in theory and which would determine the pose of the target object precisely. Finally $M_{app}$ is a mixture approximating $M$ that consists of less summands than $M$ and can be achieved by merging or dropping base elements of $M$.

\subsubsection{Coherence of the Introduced Grasp Criteria}

For an appropriate distance measure $dist$ there is always a threshold $G\geq0$ for which it holds that if $dist(g-p) \leq G$ the box the gripper can encompass at its pose close to the estimated pose of the object is the one which contains the maximum of the probability mass.\\

Of course one aims to find a strictly positive thresholds $G\gneqq 0$ to have a bigger range of tolerance for the action of grasping. Further I suppose that in this case the $L^p$ norm should be chosen as distance measure instead of the KL divergence, as it determines the absolute difference of the densities. 

\subsubsection{Convergence of Approximation}

\begin{sen}
Let $p$ and $p_{app}$ be mixture densities like defined above. Define a small threshold $G\geq0$. 
If we know that through the uncertainty of the approximation we just get a small error $\delta \geq0$ i.e. $\mathrm{P}(\|p-p_{app}\|>G)\leq \delta$, then it holds: \\
$$\mathrm{P}(\| g-p \| \leq G) >1- \varepsilon \Longrightarrow \mathrm{P}(\| g-p_{app} \| \leq 2G) >1- \tilde\varepsilon$$
where $\tilde\varepsilon := \varepsilon + \delta$.\\
\end{sen}

\textit{Proof:}
\begin{flalign}
\mathrm{P}\left(\| g - p_{app} \| > 2G\right) &= \mathrm{P}\left(\| g - p + p -  p_{app} \| > 2G\right)  \nonumber\\
		 &\leq  \mathrm{P}\left( \| g - p \|+\| p -  p_{app} \| > 2G\right)  \nonumber\\
		 &\leq  \underbrace{\mathrm{P}\left(\| g - p \| > G\right)}_{\leq \varepsilon} 
		 +  \underbrace{ \mathrm{P}\left(\| p -  p_{app} \| > G\right)}_{\leq \delta}  \nonumber\\
		 &\leq \tilde \varepsilon \nonumber
\end{flalign}
This is equivalent to: $\mathrm{P}\left(\| g - p_{app} \| \leq 2G\right)> 1-\tilde\varepsilon$
\EndProof

\begin{theo} Cauchy convergence\\
Let $\{p_n\}_n$ be a family of densities of mixtures of projected Gaussians consisting of $n$ summands. The first mixture density $p_1$ just consists of the density of a single base element. The other mixture densities are achieved by recursively concatenating another projected Gaussian density to the existing mixture density $p_n$ in each step $n\to n+1$. The weights need to be renormalized each time.\\
On repeating the recursion infinitely long concatenating base elements that estimate the pose of a target object, one would determine the pose of the target by $p_\infty$.\\
We assume that $\forall \varepsilon>0$ $\exists N_1 \in \mathbb{N}$ such that $\forall n> N_1$: 
$$P(\| g-p_n \| \leq G) >1- \varepsilon$$
Further we know that we have a Cauchy sequence $\{p_n\}_n$, i.e.: $\forall \delta>0$ $\exists N_2 \in \mathbb{N}$ $\forall m_0, m_1> N_2:$ $\|p_{m_0},p_{m_1}\|< \delta$. Allow $\delta$ to be big enough for that $N_2+1<N_1$. \\
Then choose an arbitrary $m$ with $N_2<m<N_1$ and it holds:
$$P(\| g-p_m \| \leq G + \delta) >1- \varepsilon$$\\
\end{theo}

\textit{Proof:}
Let be $n>N_1>N_2+1$ and $N_2<m<N_1$ like above.
\begin{flalign}
\mathrm{P}(\| g - p_m \| > G+\delta) &= \mathrm{P}(\| g - p_n + p_n -  p_m \| > G+\delta )  \nonumber\\
		 &\leq  \mathrm{P}( \| g - p_n \|+\underbrace{\| p_n -  p_m \|}_{<\delta} > G+\delta)  \nonumber\\
		 &\leq  \mathrm{P}(\| g - p_n \| > G+\delta -\delta )  \nonumber\\
		 &= \mathrm{P}(\| p -  p_n \| > G)  \nonumber\\
		 &\leq \varepsilon \nonumber
\end{flalign}
\EndProof

\section{Algorithms for Approximation } \label{AA}

Let $M_0,\, M_{app} \in \mathrm{MoPG}$ be two mixtures of projected Gaussians. $M_0$ is a mixture that describes the probability distribution of the pose of a target object, but might be complicated to calculate or might contain parameters which are unknown to us. The second mixture $M_{app}$ using a reduced number of base elements should be fitted to the first mixture. This can be done in a number of different ways. One is the fit of a set of samples drawn from the first distribution by use of the expectation maximization algorithm. Another possibility is the numerical minimization of the Euclidean norm of the difference of the two probability density functions $pdf$s, as a function of the parameters $\lambda_i$, $\mu_i$ and $\Sigma_i$. In this case the $2$-norm is chosen as it is easy to deal with and the absolute difference between the mixtures shall be minimized. In a next step these approaches to reduce the number of elements of a mixture would have to be compared. This stands out to be done in future work.\\

\subsection{Expectation Maximization}\label{EM}


In the following I will explain how a mixture of projected Gaussians can be fitted to a set of samples drawn from another mixture. This algorithm is well known for mixtures of Gaussians and can be applied to the projected Gaussians in the same manner because we can easily switch between tangent spaces by the double projection, central projection to the sphere and back to another tangent space which is reasonably close to the original one.\\

A mixture $M_0 \in \mathrm{MoPG}$ is defined as $\sum_{i = 1}^n \lambda_i\cdot \mathcal{N} (TS_i , \mu_i , \Sigma_i )$ as we know from definition \ref{MoPG} and has the density 
$$p(x) = \sum_{i = 1}^n \lambda_i\cdot \varphi (x| TS_i , \mu_i , \Sigma_i )$$
where $\varphi(x| TS_i ,\mu_i , \Sigma_i ) = \varphi_{ TS_i ,\mu_i , \Sigma_i }(x)$ is the density of the $i$-th base element.\\

Let us introduce now the latent variable $z$. In this context latent means to be hidden. $z$ is a $n$-dimensional binary random variable consisting of a 1-of-$n$ representation what means a certain element $z_i = 1$ and all the other $n-1$ elements equal 0. Together the values of $z_i$ thus satisfy $\sum_{i=1}^n z_i = 1$ and there are $n$ possible states which element of the vector $z$ is nonzero. We define the marginal distribution $\mathrm{P}(z)$ over $z$ in terms of the weighting coefficients $\lambda_i$ corresponding to the weights of the mixture $M$: 
$$\mathrm{P}(z_i = 1) := \lambda_i \quad \mathrm{for}\ i = 1,\ldots,n$$
As $z$ has a 1-of-$n$ representation we can write the probability distribution of $z$ in the form 
$$\mathrm{P}(z) = \prod_{i=1}^n \lambda_i^{z_i}$$
In the same way the conditional probability of $x$ given a particular value for $z$ is the distribution function of the corresponding base element
$$\mathrm{P}(x|z_i = 1) = \mathcal{N}(x|TS_i , \mu_i , \Sigma_i )$$
which can also be written in the form 
$$\mathrm{P}(x|z) = \prod_{i=1}^n  \mathcal{N}(x|TS_i , \mu_i , \Sigma_i )^{z_i}$$
Then the joint distribution $\mathrm{P}(x,z)$ is given by $\mathrm{P}(z)\cdot \mathrm{P}(x|z)$ and thus the distribution of the whole mixture of projected Gaussians is obtained by summing over all possible states of $z$:
$$\mathrm{P}(x) = \sum_{i = 1}^n \mathrm{P}(z)\cdot \mathrm{P}(x|z) = \sum_{i = 1}^n \lambda_i\cdot \mathcal{N} (x|TS_i ,\mu_i , \Sigma_i )$$

This means MoPGs can be interpreted in terms of \textit{discrete latent variables}. And a general technique for finding maximum likelihood estimators in latent variable models is the \textbf{expectation maximization (EM) algorithm} which is an algorithm that has brought applicability \cite{Combining}. At first I will give a more informal motivation of the EM algorithm and describe explicitly the steps of this algorithm afterwards.\\

The log likelihood function for a data set $X = \{x_1, \ldots x_N\}$ of independently drawn samples from a distribution $\mathrm{P}(X)$ is given by: 
$$\ln \mathrm{P}(X|TS, \mu, \Sigma, \lambda) = \sum_{j=1}^{N}{\ln \left(\sum_{i=1}^n{\lambda_i \mathcal{N}(x_j|TS_i, \mu_i,\Sigma_i)}\right)}$$
It expresses how probable the observed data set is for different settings of the parameters $TS$, $\mu$, $\Sigma$ and $\lambda$. Note that the likelihood function is not a probability distribution. \\

\begin{sen}Bayes' theorem\\
For two events $A$ and $B$ with positive probability $\mathrm{P}(B)>0$ it holds:
$$\mathrm{P}(A|B) =\frac{\mathrm{P}(B|A) \mathrm{P}(A)}{\mathrm{P}(B)}$$
For a segmentation of the sample space $\Omega$ into a finite number of disjunct events $A_i$, $i=1,\ldots, N$ and an event $B$ with $\mathrm{P}(B)>0$ it holds:
$$\mathrm{P}(A_i|B) = \frac{\mathrm{P}(B|A_i) \mathrm{P}(A_i)}{\sum_i \mathrm{P}(B|A_i)\mathrm{P}(A_i)} = \frac{\mathrm{P}(B|A_i) \mathrm{P}(A_i)}{\mathrm{P}(B)}$$\\
\end{sen}
Set the \textit{posterior probability} which is also called \textit{responsibility}: 
$$\gamma(z_{i}) \equiv \mathrm{P}(z_i = 1|x) = \frac{\lambda_i\mathcal{N}(x|TS_i, \mu_i,\Sigma_i)}{\sum_{k=1}^{n} \lambda_k\mathcal{N}(x|TS_k, \mu_k,\Sigma_k)}$$
where the values can be found using the Bayes' theorem.\\
Thus we obtain:
$$\gamma(z_{j,i}) := \frac{\lambda_i\mathcal{N}(x_j|TS_i, \mu_i,\Sigma_i)}{\sum_{k=1}^{n}{\lambda_k\mathcal{N}(x_j|TS_k, \mu_k,\Sigma_k)}}$$\\
Note that the samples $x_i\in S_3\times \R^3$ $\forall i\in \{1,\ldots,N\}$ are drawn from the special Euclidean group such that the rotation part lies on the 3-sphere. To assign the appropriate responsibilities to these samples they have to be reprojected by $\prod_{p_k}^{-1}(x_i)$ to the tangent space of the Gaussian kernel with tangent point $p_k$ for $k = 1,\ldots, n$.\\

Maximizing the log likelihood function for a projected Gaussian mixture model turns out to be a more complex problem than for the case of a single projected Gaussian. Fortunately the EM algorithm is an elegant and powerful method for finding maximum likelihood solutions for models with latent variables. It consists of the two following steps between which we alternate until the algorithm converged as stated in \cite{Bishop}. Initially there are arbitrary values chosen for the means, covariances and weighting coefficients.
\begin{itemize}
\item \textbf{Expectation step (E step):}\\
The current values are used for the parameters to evaluate the posterior probabilities or responsibilities.
\item \textbf{Maximization step (M step):}\\
The probabilities obtained in the E step are used to reestimate the means, covariances and mixing coefficients. Then the tangent spaces are changed by double projection so that we obtain Gaussian kernels with zero mean for the rotation.
\end{itemize}

It can be shown that each update to the parameters resulting from an E step followed by an M step is guaranteed to increase the log likelihood function $\ln \mathrm{P}(X|TS, \mu, \Sigma, \lambda)$. In practice, one expects the algorithm to have converged when the change in the log likelihood function, or alternatively in the parameters, falls below some fixed threshold. It is well known that the EM algorithm needs comparatively many iteration steps and that each cycle is computationally expensive. Thus it is common to run other algorithms like the $n$-means algorithm first to achieve better initial values than randomly chosen ones. Further I want to mention that another disadvantage of the EM algorithm arises from the fact that it might get stuck in some local maxima of the log likelihood function instead of finding the global maximum. This is a second indication for the need to choose the initial values carefully.\\

\textbf{Summary of the EM algorithm}
\begin{enumerate}
\item Set the initial value for the means $\mu_i$, covariance matrices $\Sigma_i$ and weighting coefficients $\lambda_i$ and evaluate the log likelihood with these values.
\item E step:\\
Evaluate the responsibilities $\gamma(x_{n,i})$ using the current parameter values
$$\gamma(z_{j,i}) := \frac{\lambda_i\mathcal{N}(x_j|TS_i, \mu_i,\Sigma_i)}{\sum_{k}{}{\lambda_k\mathcal{N}(x_j|TS_k, \mu_k,\Sigma_k)}}$$
\item M step: \\
 Reestimate the parameters using the current responsibilities 
 $$\mu_i^{new} = \frac{1}{N_i} \sum_{j=1}^{N}{\gamma(z_{j,i})\cdot x_j}$$
 $$\Sigma_i^{new} = \frac{1}{N_i} \sum_{j=1}^{N}{\gamma(z_{j,i})(x_j-\mu_i^{new})(x_j-\mu_i^{new})^\top}$$ $$\lambda_i^{new} = \frac{N_i}{N}$$
 where $N_i = \sum_{j=1}^{N}{\gamma(z_{j,i})}$
\item Evaluate the log likelihood: 
$$\ln \mathrm{P}(X|TS, \mu, \Sigma, \lambda) = \sum_{j=1}^{N}{\ln \left(\sum_{i=1}^n{\lambda_i \mathcal{N}(x_j|TS_i, \mu_i,\Sigma_i)}\right)}$$
and check for convergence of either the parameters or the log likelihood. If the convergence criterion is not satisfied return to the E step.\\
\end{enumerate}

\subsection{Monte Carlo}\label{MC}

Let $p_0$ be the density of the mixture of projected Gaussians $M_0\in \mathrm{MoPG}$ and $p_{app}$ be the density of $M_{app}\in  \mathrm{MoPG}$. The square of the $L^2$ norm of these probability density functions $p_0$ and $p_{app}$ is defined as:
$${\|p_0-p_{app}\|_2}^2:=\int_{S_3\times \mathbb{R}^3} (p_0(q)-p_{app}(q))^2\;\mathrm{d}q$$
The minimization of this integral is equivalent to the minimization of the Euclidean norm. The calculation of it can be done using the Monte Carlo algorithm which relies on summation instead of the costly integration. \\

\textit{What is Monte Carlo (MC) integration in general?}\\
MC integration is a numerical integration that randomly chooses the points at which the integrand is evaluated.
First I will specify the region $A$ to integrate over. To estimate the area of interest $D$, pick a simple area $A$ which is easy to calculate and which contains $D$. Then pick a sequence of random points that fall within $A$. Some fraction of these points will also fall within $D$. The area of $D$ is then estimated as this fraction multiplied by the area of $A$. \\
As we handle a mixture of projected Gaussians, we know for each kernel $PG_i$ the center of mass $D_i$ and thus easily can deduce an approximation of the center of mass of the mixture. Of course we require the area $A\subset S_3\times\R^4$ to contain $D_i\, \forall i$.\\
Now the algorithm has to be defined:
\begin{itemize}
\item $\{a_n\}_n, \ n\in \N$ is a random sequence of identically distributed points in the integration area $A$.
\item $g$ is the integrand. For a function of one variable the average value of $g(x)$ can be estimated by:
$$\tilde {g_N} \approx \frac{1}{N} \sum_{n=1}^N{g(a_n)}, \ N\geq 1$$
\item $M_A$ is the mass of the whole integration area.
\item Iteration:
\begin{enumerate}
\item $V_1 = g(a_1)$ is the value of the first point
\item $V_{n+1} = \frac{n}{n+1} \cdot V_n + \frac{1}{n+1} \cdot g(a_{n+1})$ defines the steps from $n\,(\geq 1)$ to $n+1$
\end{enumerate}
\item $V := \lim_{n \to \infty} V_n$
\item The value of the integral $I$ is then:
$$ \int_{A} g(x) \,\mathrm{d}M_A = V \cdot M_A$$ and an approximation of the integral can be given by $I  \approx M_A \cdot \tilde g_N$
\end{itemize}

An estimate for the error is given by:
$$err = M_A\cdot \sqrt{\frac{{\tilde {g^2}}_N - {\tilde{ g_N}}^2}{N}}$$
where ${\tilde{ g^2}}_N := \frac{1}{N} \sum_{n=1}^N{ g^2(a_n)}$.\\

On integrating with this algorithm the values converge with order $o(\frac{1}{\sqrt{N}})$ regardless of the smoothness of the integrand. MC integration is not competitive in one or two dimensions, but in higher dimensions. Further keep in mind that each time the MC algorithm is implemented using the same sample size $N$, it will come up with a slightly different value as the integration points are picked randomly. Obviously larger values of $N$ produce more accurate approximations.\\
I suppose the iteration can be stopped when the difference $|V_{n+1} - V_n|$ remained sufficiently long under a majorant with sufficiently small sum. Then we assume the algorithm to converge significantly to the true value of the integral $I$.\\
The most important advantage of this approximation of the integral is that the algorithm is easy and fast.\\

The traditional Monte Carlo algorithm distributes the evaluation points uniformly over the integration region like mentioned above. But there are also adaptive algorithms such as VEGAS and MISER:
\begin{itemize}
\item \textbf{MISER Monte Carlo}\\
This algorithm of Press and Farrar \cite{MiserMC} is based on recursive stratified sampling. This technique aims to reduce the overall integration error by concentrating integration points in the regions of highest variance.
\item \textbf{VEGAS Monte Carlo}\\
The algorithm of G. P. Lepage \cite{Vegas} is based on importance sampling. It samples points from the probability distribution described by the absolute value of the function $|g|$, so that the points are concentrated in the regions that make the largest contribution to the integral.\\
\end{itemize}

We would like to approximate $ {\|p_0-p_{app}\|_2}^2$ by: 
$$\frac{1}{N} \sum_{i=1}^N {(p_0(a_i)-p_{app}(a_i))^2\cdot M_A}$$
where $a_i\in A$ are the elements of the sample set $\{a_1, \ldots, a_N\}$ and $A\subset S_3\times\R^4$ is the region to integrate. It turned out that the specification of the region $A$ lacks an easy solution but we found a possibility to elegantly eludes this specification which I will introduce in the following.\\

\subsubsection{Importance Sampling}


We know that the Monte Carlo estimator of $\mathrm{E}[g(X)]$ is $\tilde{g_n}(X) = \frac{1}{n} \sum_{i=1}^{n}{g(x_i)}$ for $X$ being a uniformly distributed continuous random variable. Furthermore this estimator is unbiased, what means $\mathrm{E}[\tilde {g_n}(X)] = \mathrm{E}[g(X)]$.\\

An important thing to note is that there is \textit{no restriction} that says that the random variables must be uniformly distributed.
It is obvious that the choice of distribution from which to draw the random variables will affect the quality of their Monte Carlo estimator. This implies \textit{importance sampling} \cite{ImportantSampling} is choosing a good distribution from which to simulate the random variables.\\ 

Now consider $X$ to be a continuous random variable with any probability density function $f_X(x) >0$ $\forall x\in \R$. Then the expected value of a function $g$ of $X$ is: 
$$\mathrm{E}_{f_X}[g(X)] = \int_{x \in \R} g(x)f_X (x) \, \mathrm{d}x$$
This is deduced from the following:\\
If $X$ is a continuous random variable defined on a probability space $(\Omega, \Sigma, P)$, then the expected value of $X$ is defined as: 
$$\mathrm{E}_{f_X}[X] = \int_\Omega X\, \mathrm{d}P$$
When this integral converges absolutely, it is called the expectation of $X$. If the probability distribution of $X$ admits a probability density function $f_X(x)$ on $\Omega$, then the expected value can be computed as: 
$$\mathrm{E}_{f_X}[X] = \int_{-\infty}^\infty x f_X(x)\, \mathrm{d}x$$
And the expected value of an arbitrary function of $X$, $g(X)$, with respect to the probability density function $f_X(x)$ is given by the inner product of $f_X$ and $g$.\\

Then we can estimate the value of $\mathrm{E}_{f_X}[\frac{g(x)}{f_X (x)}]$ by generating a number of random samples according to $f_X$, computing $\frac{g}{f_X}$ for each sample, and finding the average of these values. As more and more samples are taken, this average is guaranteed to converge to the expected value, which is also the value of the integral $I$.  \\

\begin{dat}\label{mc}
Let $f_X(x)$ be a density for a continuous random variable $X$ which this time only takes values in $A$ so that $\int_{x \in A}f_X(x)\,\mathrm{d}x = 1$ and $\mathrm{E}_{f_X}$ denotes the expectation with respect to the density $f_X$: 
$$\int_{x \in A}g(x)\,\mathrm{d}x = \int_{x \in A}g(x)\cdot\frac{f_X(x)}{f_X(x)}\,\mathrm{d}x = \int_{x \in A}\frac{g(x)}{f_X(x)}\cdot f_X(x)\,\mathrm{d}x =  \mathrm{E}_{f_X}\left[\frac{g(x)}{f_X(x)}\right]$$
so long as $f_X(x) \neq 0$ for any $x \in A$ for which $g(x) \neq 0$.\\
From this follows that the Monte Carlo estimator is: 
$$\tilde g_{n,f_X} (X) := \frac{1}{n} \sum_{i=1}^{n}{\frac{g(x_i)}{f_X(x_i)}}$$
where $x_i \sim f_X(x)$.\\
\end{dat}

\textit{Proof:}
\begin{flalign}
\mathrm{E}_{f_X}\left[ \tilde g_{n,f_X} (X) \right] &= \frac{1}{n} \sum_{i=1}^{n}\mathrm{E}_{f_X}\left[ \frac{g(x_i)}{f_X(x_i)} \right]\nonumber\\
	&= \frac{1}{n} \sum_{i=1}^{n}\int_A \frac{g(x)}{f_X(x)}f_X(x) \mathrm{d}x\nonumber\\
	&= \frac{n}{n} \int_A g(x)\frac{f_X(x)}{f_X(x)} \mathrm{d}x\nonumber\\
	&= \int_A g(x)\mathrm{d}x\nonumber\\
	&= I\nonumber
\end{flalign}
given that $\frac{g(x)}{f_X(x)}$ is finite $\forall \,x$.
\EndProof

Finding a MC estimator that provides good estimates in a reasonable amount of computing time is not a trivial task.\\
An assessment for MC estimators can be given by the variance \cite{MCMethods} which is defined by:
$$\mathrm{Var}[\tilde g_{n,f_X}(X)] = \frac{1}{n} \int_{x\in \R} \left(g(x) - \mathrm{E}[g(X)]\right)^2 f_X(x) \mathrm{d}x$$
The smaller the variance for the same amount of computational effort the better the estimator in comparison to its competitors. Thus we are looking for an importance sampling function $f_X(x)$ that has the following properties:
\begin{itemize}
\item $f_X(x) > 0$ whenever $g(x) = 0$
\item $f_X(x)$ should be close to being proportional to $|g(x)|$
\item It should be easy to simulate values from $f_X(x)$.
\item It should be easy to compute the density $f_X(x)$ for any value $x$ one might realize.
\end{itemize}

I want to point out that serious difficulties arise if $f_X(x)$ gets small much faster than $g(x)$ out in the tails. Though drawing a sample from the tail of the distribution is unlikely the MC estimator will give a big error if it occurs. $\frac{g(x_i)}{f_X(x_i)}$ for such an unlikely $x_i$ may be orders of magnitude larger than the typical values of $\frac{g(x_i)}{f_X(x_i)}$.\\

To estimate the absolute error of the MC integration with importance sampling the central limit theorem can be used. It states that $\tilde g_{n,f_X}(X)$ converges to the normal distribution as $n\to\infty$. Let's denote $Y_i := \frac{g(x_i)}{f_X(x_i)}$ and $Y:= Y_1$. In particular the central limit theorem gives for $t\in \R$ and $\sigma(Y)$ the standard deviation of $Y$:
$$\lim_{n \to \infty} \mathrm{P}\left[\frac{1}{n} \sum_{i=1}^{n} Y_i- \mathrm{E}[Y] \leq t\cdot \frac{\sigma(Y)}{\sqrt{n}}\right] = \frac{1}{\sqrt{2\pi}} \int_{-\infty}^{t} \mathrm{e}^{- x^2/2} \, \mathrm{d}x$$
Hence we receive the following equation for the error:
$$\mathrm{P}\left[|\tilde g_{n,f_X} - I| \geq t\cdot \sigma(\tilde g_{n,f_X})\right] = \sqrt{\frac{2}{\pi}} \int_{t}^{\infty} \mathrm{e}^{- x^2/2} \, \mathrm{d}x$$
where the standard deviation of $\tilde g_{n,f_X}$ is $\sigma(\tilde g_{n,f_X})=\frac{1}{\sqrt{n}}\sigma(Y)$. \\

By the way I want to bring up that the treatment of higher order error estimation is not just an academic point. Lazopoulos deals in his paper \cite{QuasiMC} with first-order errors of MC integration that is the error directly on the integral estimate and second-order errors that is the error on the error estimate of the integration. A mis-estimate of the integration error can lead to a serious under and over estimate of the confidence level and thus it's estimation should be done carefully. \\

Finally I want to explain why for Monte Carlo integration with importance sampling over mixtures of Gaussians as distribution function no limits of integration are needed. Let's calculate the MC estimator $\tilde g_{n,f_X} (X)$ for $n$ samples of a mixture of projected Gaussians with length $d$ with a set of normally distributed samples $\{x_1,\ldots,x_n\}$. We define the fraction
\begin{flalign}
\frac{g(x)}{f_X(x)}:&= \sum_{i=1}^{d}\frac{g_i(x)}{f_{X,i}(x)}\nonumber\\
	&= \sum_{i=1}^{d}\frac{\lambda_i\cdot 1/C(x)\cdot 1/\sqrt{\det(2\pi\Sigma_i)}\cdot \exp(-\frac{1}{2}(x-\mu_i)^\top \Sigma_i^{-1}(x-\mu_i))}{1/\sqrt{\det(2\pi\Sigma_i)}\cdot \exp(-\frac{1}{2}(x-\mu_i)^\top \Sigma_i^{-1}(x-\mu_i))}\nonumber\\
	&= \sum_{i=1}^{d}\lambda_i\cdot 1/C(x)\nonumber
\end{flalign}
where $1/C(x)$ is the correction weight for the parameterization in the integration. This means for $x=(x_1,x_2,x_3,x_4,x_5,x_6)^\top \in \R^6$ it is defined as $1/C(x) = 1/(1+ x_1^2+x_2^2+x_3^2)$ as I already showed at the end of section \ref{PG}. For all $i = 1,\ldots, d$ we know that $f_{X,i}(x)$ will never be smaller than $g_i(x)$ as $\lambda_i \leq 1$ and $1/C(x)\leq1$ $\forall x\in\R^6$. Thus there is no risk for abnormal behavior in the tails of the probability density. As we know from theorem \ref{mc} it holds that $I = \mathrm{E}_{f_X}\left[ \tilde g_{n,f_X} (X) \right]$. Hence we can calculate an approximation of the integral by the following formula:
$$I \approx \sum_{i=1}^{d}\lambda_i \frac{1}{n}\left(\sum_{j=1}^{n}\frac{1}{C(x_{i_j})}\right)$$
where $x_{i_j}$ is the $j$th element of the $\varphi(\mu_i,\Sigma_i)$ distributed sample set with $\varphi(\mu_i,\Sigma_i):= 1/\sqrt{\det(2\pi\Sigma_i)}\cdot \exp(-\frac{1}{2}(x-\mu_i)^\top \Sigma_i^{-1}(x-\mu_i))$.\\

Formally we would have to pick the samples $x_{i_j}$ out of a box shaped integration area $A$ including the area of interest $D$ and than let the side length of the box go to infinity always normalizing with the density of the underlying sample distribution. But it can be seen directly that each of the summands of the mixture and thus the whole mixture itself does not contain significant mass in the tails as we just work with finite mixtures of projected Gaussians. \\

What stands out to be done in future work is the minimization of the square of the $L^2$ norm of the densities of the mixtures $M_0$ and $M_{app}$:
$$h(\lambda_i,TS_i, \mu_i,\Sigma_i) := \min \| p_0-p_{app}\|_2^2$$
By now the Monte Carlo integration can at least be used to validate the fit of a mixture achieved with the expectation maximization algorithm.\\

\chapter{Implementation and Experimental Verification} \label{Sensor}


Recall that a robot makes several localization attempts to estimate the pose of a target object. In the following I will explain how the robot draws conclusions from the separated 3D SIFT features it detects to the object pose. This procedure is called sensor model.

\begin{itemize}
\item Every object of the robots data base of 3D models is given a Cartesian coordinate system $\mathrm{CS}_O$. To systematize the arbitrary choice of origin and axis, we postulate the origin of the coordinate system to be the midpoint of the bottom of the object. Then we define the $x$- and $y$-axes to be the main axes in the basement of the object the way that together with the $z$-axis, which is straight up, they produce a right-handed coordinate system. Now any point feature on the objects surface can be described by a 3-dimensional position and an orientation in 2 dimensions similar to the description in \cite{3DPoseTracking}.
\item Each camera or 3D sensor of the robot has a normalized coordinate system $\mathrm{CS}_C$ which we define the way that the viewing direction equals the $z$-axis. As we want the coordinate system to be a right-handed Cartesian one, the other axes are determined on claiming the $x$-axis to point to the right from viewing direction and thus the $y$-axis points down.
\item If the robot detects a feature with its camera, a new Cartesian coordinate system $\mathrm{CS}_F$ for the feature has to be defined. $\mathrm{CS}_F$ has its origin at the mean of the estimation for the features pose. To take into account that the most likely hypothesis is frontal perspective to the feature we define the $z$-axis to point to the origin of $\mathrm{CS}_C$. As features have an orientation on the locally planar objects surface, the $x$- and $y$-axes are predefined through the 3D object model and the orthogonality to the $z$-axis.
\end{itemize}

\begin{wrapfigure}{l}{7cm}
\centering
  \includegraphics[width=0.45\textwidth]{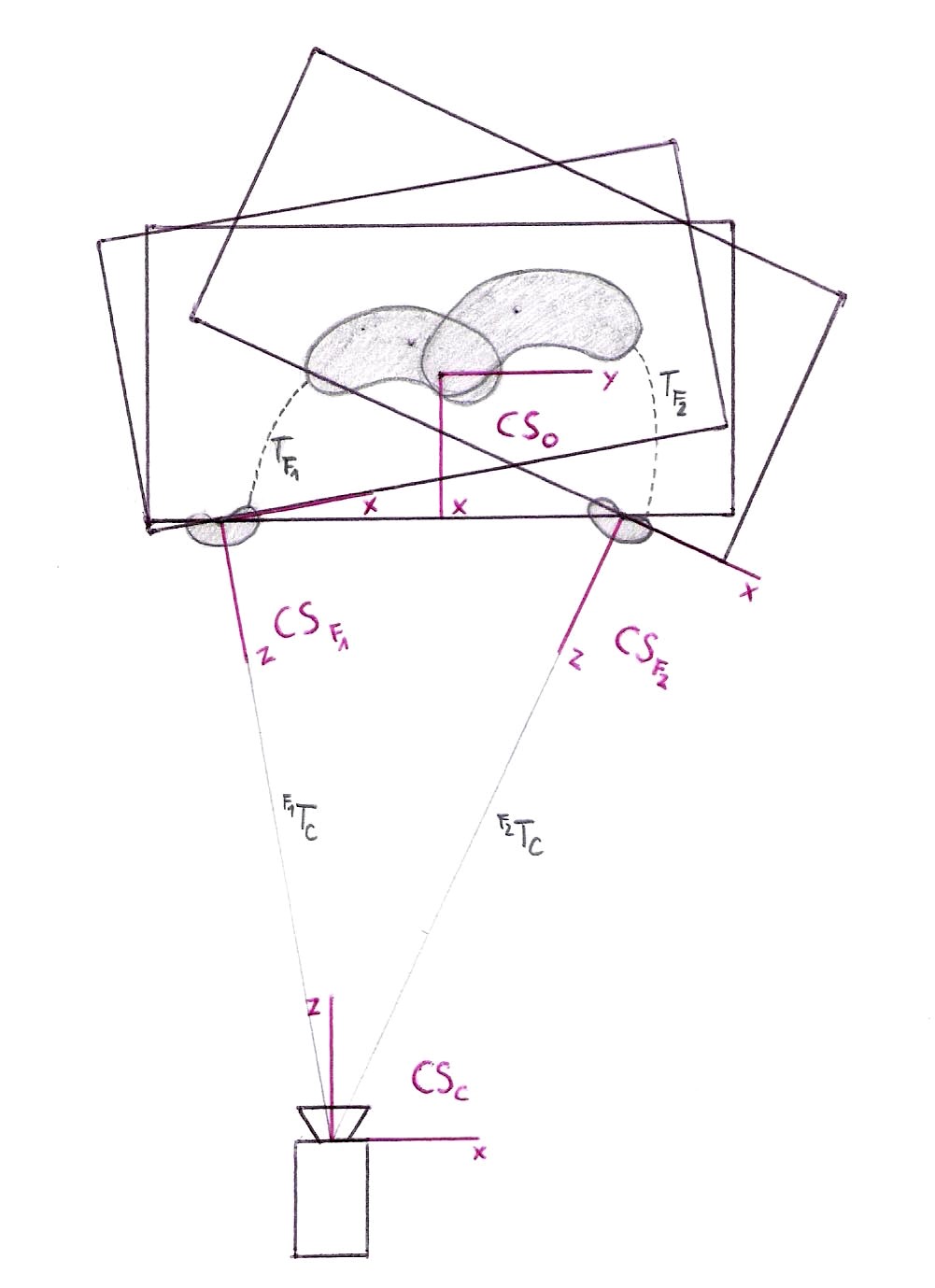}
  \caption[Sensor model]{ }
\label{plotSelf}
\end{wrapfigure}

Let ${}^FT_C: \mathrm{CS}_C \to \mathrm{CS}_F$ be the transformation from the camera coordinate system to the feature coordinate system. As I create the mixture that models the distribution of the feature in camera coordinates, I have to change the coordinate system to represent the feature at its true pose. As mentioned already we have the 3D model of the object and can describe any feature on the object by the function $T_O$. But to grasp the object, we need to model the object pose by knowing the feature. The inverse of the function $T_O$ is the transformation $T_F = (T_O)^{-1}$ that lets us draw conclusions from the feature pose on the object's surface about the object pose.\\
As we represent any rigid motion by a rotation matrix and a translation vector or equivalently by a dual quaternions, the inverse of the dual quaternion represents exactly the inverse rigid motion.\\

Features have a predefined orientation and thus the possible rotation around the $z$-axis is very small, in contrary around the $x$- and $y$-axis I allowed as uncertainty rotations in the interval $[-15^\circ, 15^\circ]$. We know about the translation that there can be little shifting in the $x$-$y$-plane, but as the scale of the feature also is uncertain, the dislocation in direction of the $z$-axis can be fairly big. Hence we need a mixture of 6D projected Gaussians $\mathrm{MoPG}_0$ to model the pose of the feature the robots camera detected. This model now has to be shifted to the pose, where the feature was detected by the base element $\mathrm{PG}_1 = {}^FT_C$.\\
As the 3D models of the objects are imperfect, one has to calculate with little uncertainties in position and orientation of the feature on the objects surface. Thus another 6D projected Gaussian $\mathrm{PG}_2 = T_F$ is required.\\

Now the distribution of the object's pose can be estimated in camera coordinates by:
$$\mathrm{PG}_1\circ \mathrm{MoPG}_0\circ \mathrm{PG}_2$$
Of course a single feature is not sufficient to determine the pose of an object. In reality about 50 features on one taget object are needed to estimate the pose precisely enough.\\

\section{Python Code}\label{code}

Now I want to give a brief documentation about the code I wrote to create the framework.
As Python is a object oriented programming language I structured the system in mainly five object classes. These are:
\begin{itemize}
 \item MoPG\_tangentSpace
 \item MoPG\_baseElement
 \item MoPG\_mixture
 \item Quaternion
 \item DualQuaternion
\end{itemize}

A tangent space consists of the tangent point $p$ on the (hyper-)sphere surface and a basis $B$ of the tangent space in world coordinates.
That means if the point $p\in\R^d$, the basis is a $d \times (d-1)$ matrix, that is completed to a basis of the $d$-dimensional space by concatenating the vector $p$ as first column.\\

The class \textbf{MoPG\_tangentSpace} contains several functions. \textit{\_init\_} always is the first method of a class and creates a representative. Furthermore \textit{equal} tests whether two tangent spaces are equal and \textit{display} changes the tangent space to a printable formate on the display.\\
The method \textit{tangentSpaceToSphereCentralProjection} projects a 3-dimensional vector in the tangent space by central projection to the sphere whereas \textit{sphereToTangentSpaceCentralProjection} projects any point on the sphere surface to any given tangent space except for the case that the tangent point and the point on the sphere to be projected have an angle of $\pi/2$ between themselves. With \textit{transformFromSelfToTS} the tangent space can be changed. Therefore a vector $v$ in the first tangent space with tangent point $p_1$ is projected to the sphere and then is back projected to the second tangent space with tangent point $p_2$. In these methods the translational part remains unchanged and can be inputted to the function also.\\
Finally this class has the method \textit{poseTransformationTS} which transforms a 6-dimensio-nal vector consisting of a rotation and a translation part with a given tangent space by a vector with its appropriate tangent space to another 6D vector in a third tangent space.\\

A base element is a projected Gaussian consisting of a tangent space, like defined above, the mean value of the gauss kernel and the appropriate covariance matrix. The mean value is a vector with dimension $d-1$ for the rotation part if the tangent point on the sphere is $d$-dimensional. \\

The class \textbf{MoPG\_baseElement} also contains the methods \textit{\_init\_}, \textit{equal} and \textit{display}. Moreover I wrote a function \textit{dimensions} to determine the dimensions of the rotation and the translation part. \textit{computeMassMonteCarlo6D} computes the mass of the base element by using Monte Carlo integration with importance sampling. It is needed for renormalization.\\
Any base element contains information about orientation and position in the special Euclidean group which can also be represented by a dual quaternion. This pose is calculated by \textit{extractDualQuaternion}. \textit{changeTS} changes the tangent space of a base element by falling back on the tangent space method \textit{transformFromSelfToTS}. The method \textit{mahalanobisDistance} determines the Mahalanobis distance between two base elements. Furthermore I implemented the functions \textit{fuse}, \textit{merge} and \textit{randomPoseTransformation} which fuse, merge and compose base elements respectively.\\
If it is desired to obtain the mean vector $\mu_3 = 0$ for the fused respectively merged base element, the 'modeFlag' has to be set to 1.\\
\textit{density} is a method that returns the density of a point which has a 4D rotation part already projected to the sphere and a 3D translation part. If the point is close to the equator (for the tangent point being a pole) the back projected point to the tangent space gets to infinity at least in one dimension. Then we set the density to be 0.\\
Finally the methods \textit{draw1Sample}, \textit{draw1SampleMat}, \textit{drawNSamples} and \textit{paintCS} all sample from the distribution of the base element and are needed for visualization.\\

\textbf{MoPG\_mixture} denotes the class of mixtures of projected Gaussians. Each mixture consists of a list of base elements, each with its weight i.e. it is an array with two columns: 
$$\mathrm{Mixture} = [[\mathrm{PG_1},\lambda_1], [\mathrm{PG_2}, \lambda_2],\ldots ,[\mathrm{PG_n},\lambda_n]]$$
The methods of this class are \textit{\_init\_} and \textit{equal} like in the other classes, but instead of \textit{display}, this class has the function \textit{toList} what changes the mixture into a list which is printable and has all list features implemented in Python, but does not give a reasonable output on the display. \\
The method \textit{density} calculates the mixture density of a point by using the base element method of same name. The other methods \textit{computeMassMonteCarlo6D}, \textit{fuseMoPG} and \textit{randomMoPGTransformation} also just fall back to the corresponding methods for base elements and apply them on each entry of the mixture. Furthermore I want to mention that the renormalization constant $C$ of the function \textit{randomMoPGTransformation} just consists of the weights $\lambda_i$ and $\lambda_j$. That means $C= 1/(\sum_{i,j}\lambda_i*\lambda_j)$ as there is no question of whether two base elements can be applied at the same time, because the probability distributions are assumed to be independent.\\
\textit{drawNSamples}, \textit{drawNSamplesV} and \textit{paint} are used for visualization of the mixture and sample from each of its base elements.\\

The class of \textbf{quaternions} is well known in algebra, but the implemented module in Python is not structured well. Therefore I wrote the code for my own class of quaternions with the following methods:
\begin{itemize}
\item \textit{\_init\_()} makes a quaternion out of a list with four entries [a,b,c,d].
 \item \textit{norm()} calculates the norm of a quaternion whereas \textit{norm2()} gives the square of the norm
 \item \textit{normalize()} and \textit{normalizeD()} normalize the quaternion. The first method has a copy as output, the second one is destructive on the input quaternion
\item \textit{conj()} and \textit{conjD()} conjugate the input quaternion and give it back in copy or destructive.
 \item \textit{toList()} prints the quaternion to a list to receive a readable output.
 \item \textit{inv()} and \textit{invD()} calculate the inverse of a quaternion (copy and destructive version)
 \item \textit{equal()} tests whether two quaternions are equal.
 \item \textit{copy()} creates a copy of the input quaternion.
 \item \textit{plus()} adds two quaternions: $Quat1+Quat2 = Quat1.plus(Quat2)$
 \item \textit{scalar()} multiplies a scalar $\lambda \in \R$ to a quaternion: $\lambda*Quat = Quat.scalar(\lambda)$
\item \textit{times()} multiplies one quaternion Quat1 with another quaternion Quat2: $Quat1*Quat2 = Quat1.times(Quat2)$
 \item \textit{toMatrix()} returns the $3\times3$ rotation matrix of a quaternion.
 \item \textit{randomUnit()} creates a unit quaternion randomly from an equal distribution on $S_3$ whereas \textit{randomImaginary()} creates an imaginary quaternion randomly.
 \item \textit{radAxis()} returns a unit quaternion corresponding to a rotation by radian measure around a given axis, \textit{degreeAxis()} does the same with degree.\\
\end{itemize}

The \textbf{dualQuaternions} are the last class I created to complete the framework. They have the corresponding methods:\\
\textit{\_init\_()}, \textit{plus()}, \textit{equal()}, \textit{copy()} and \textit{toList()}.\\
\textit{times()} multiplies two dual quaternions by the multiplication defined in \ref{DQ} whereas \textit{conjQuat()} returns the quaternion conjugate, \textit{conjDual()} returns the dual conjugation and \textit{conjTotal()} the total conjugation. These conjugations are also defined in \ref{DQ}.
Because every dual quaternion describes a rigid motion it contains a rotation and a translation which can also be written as rotation matrix and translation vector. That does the method \textit{transformationMatrix()}. Finally \textit{inv()} calculates the inverse of the input dual quaternion.\\

Furthermore there exist the following functions which are not related to any class:
\begin{itemize}
\item \textit{computeJacobian} calculates the Jacobian from the tangent space method \textit{transformFromSelfToTS} at the point given as input to the function. 
\item \textit{rottransVector6D} returns a 6D vector that contains the information about rotation and translation, not in quaternion style, but in a quite normal 6D vector.
\item \textit{makeSamples} creates a given number of samples which fulfill special requirements like the distribution where they are drawn from.
\item \textit{makeInitMixture} makes a mixture out of a list of samples.
\item \textit{rotToQuat} is a function that returns the quaternion which represents the same rotation like a given rotation matrix.
\item \textit{transformationToDQ} changes the input of a tuple of quaternions $q_r$ and $q_t$ to a dual quaternion with dual part $q_d = 1/2\cdot q_t*q_r$.
\item \textit{DQToTransformation} makes the inverse transformation from a dual quaternion to a tuple of quaternions.
\item The function \textit{transformPose} returns a dual quaternion which is obtained after applying a transformation dual quaternion to a pose dual quaternion.\\
\end{itemize}

\section{Visualization}\label{example}

We have a range of options introduced to improve the pose estimation like weighting the measurement results, fusing, merging, dropping base elements and making another localization attempt. An algorithm shall evaluate what to apply to the distribution function describing the object pose to improve it and when the distribution is peaked enough for that a grasp criterion is fulfilled. For this framework it stands out to define such an evaluation algorithm. A possible approach is the further development of the MC-SOPE algorithm Glover introduced in \cite{Jared}. This algorithm to solve the single object pose estimation uses importance sampling to generate a weighted set of pose samples of the target distribution and then returns the top $n$ samples ranked by weight. We would need to sample from the target distribution that is achieved after applying different options like the modification operations and then compare the distributions fitted to the resulting sample sets.\\

\begin{wrapfigure}{l}{7.5cm}
\centering
  \includegraphics[width=0.4\textwidth]{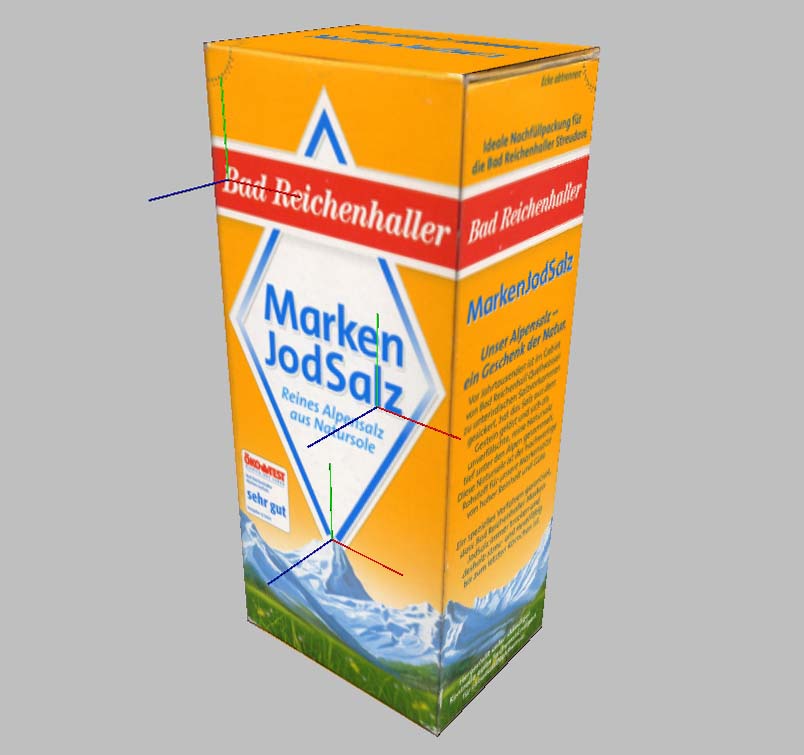}
  \caption[Salt box]{ }
\label{plot2}
\end{wrapfigure}

In the following I describe a simulated experiment how such an algorithm might proceed. The programming language Python offers a Coin binding to enable visualizations like the ones I made. \cite{Inventor} explains how to write applications using the Open Inventor toolkit in this Coin binding.\\
Imagine a robot that has the task to grasp the salt box of figure \ref{plot2}. Let's assume that the robot detects the features 'B' of the word \textit{Bad} and 'l' of \textit{Salz}. Later on I will also work with the mountain top as a third feature.\\

\begin{wrapfigure}{l}{7.5cm}
\centering
  \includegraphics[width=0.4\textwidth]{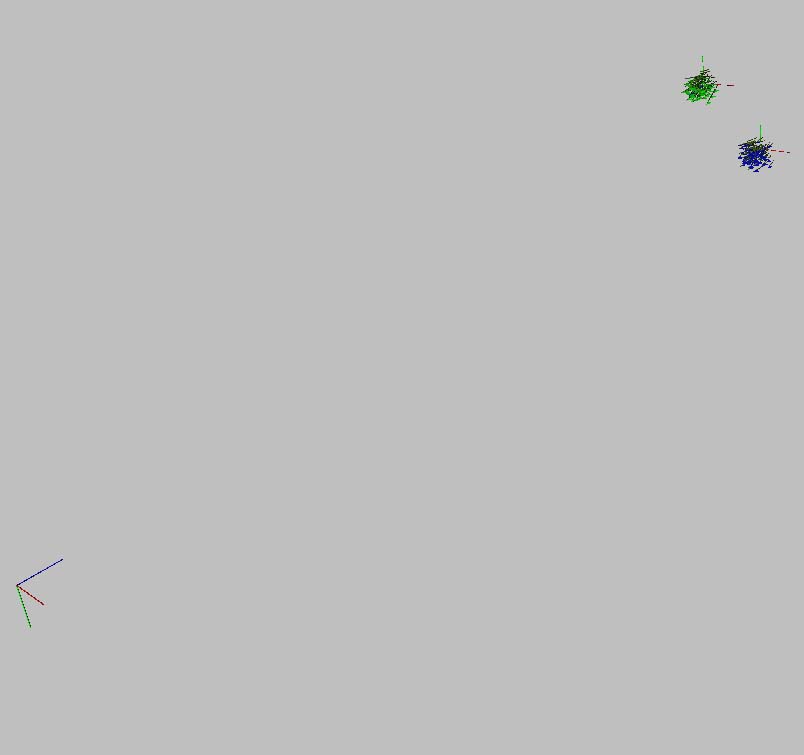}
  \caption[Feature poses - 2 features]{}
  \label{plot3}
  \vspace{10mm}
   \includegraphics[width=0.4\textwidth]{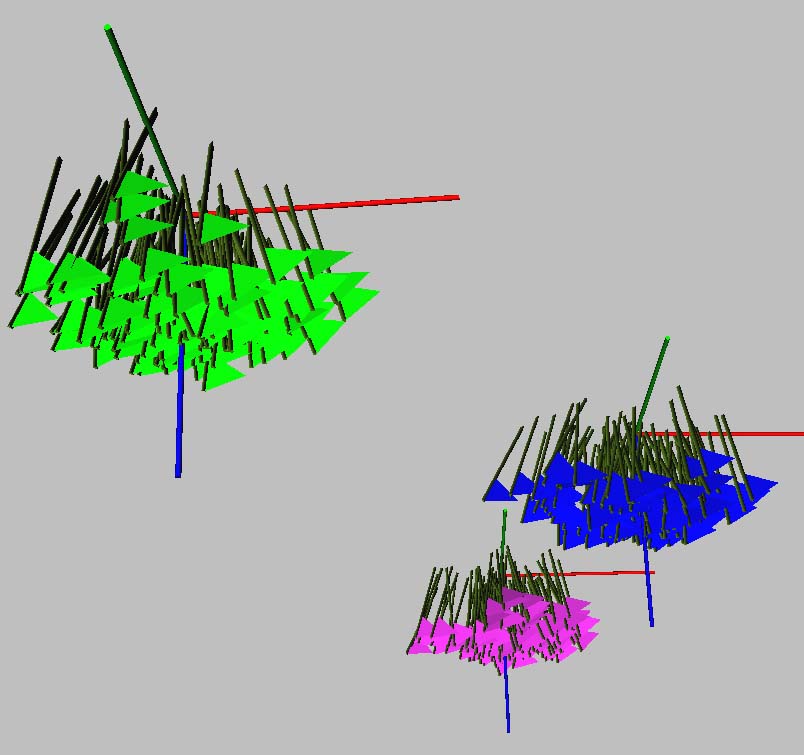}
  \caption[Feature poses - 3 features]{}
\label{plot4}
  \vspace{10mm}
  \includegraphics[width=0.4\textwidth]{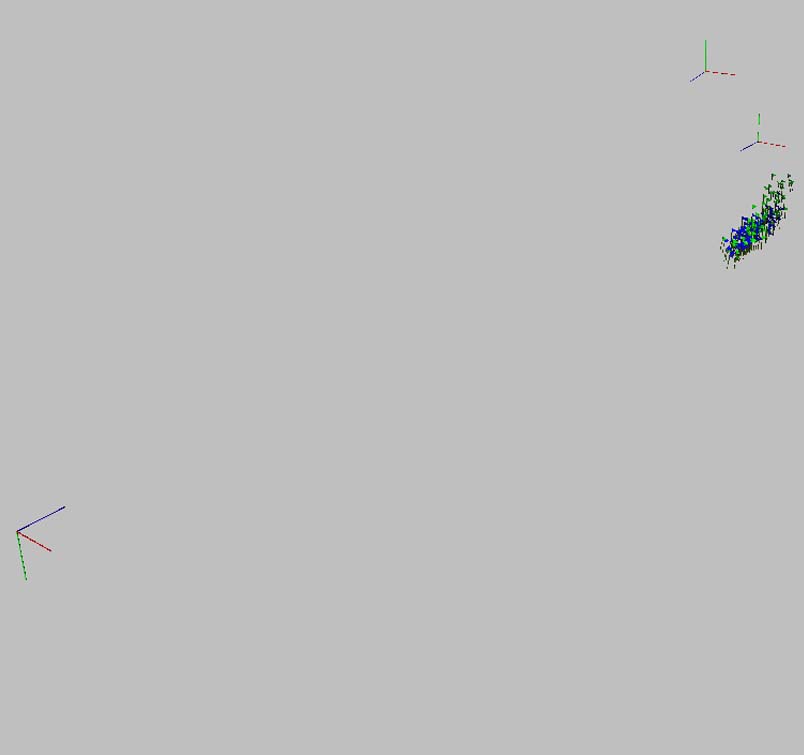}
  \caption[Object pose - 2 features]{}
\label{plot5}
\end{wrapfigure}

The flags represent samples out of the mixture describing the pose of the features in $S_3\times\R^3$. The most likely hypothesis is that the robot detects features which are frontal to the camera and thus we model the probability distribution as represented in figure \ref{plot3}. The green flags represent the 'B' and the blue flags the 'l'. Further the coordinate system on the left is the camera coordinate system $CS_C$.\\

As mentioned already I assume that the the robot will detect the feature mountain top on the salt box as well. This feature will be modeled by the pink samples drawn from the mixture of projected Gaussians describing the probability distribution of the pose of the mountain top.\\
The figure \ref{plot4} shows the sample sets in a view from above the salt box.\\

In figure \ref{plot5} the green and blue samples are drawn from the mixture distribution describing the pose of the object. The result is what we get on drawing conclusions from the feature pose to the pose of the object. \\
For easier orientation again the coordinate system of the camera can be seen in the the lower left corner of the figure as well.\\

\begin{wrapfigure}{l}{7.5cm}
\centering
  \includegraphics[width=0.4\textwidth]{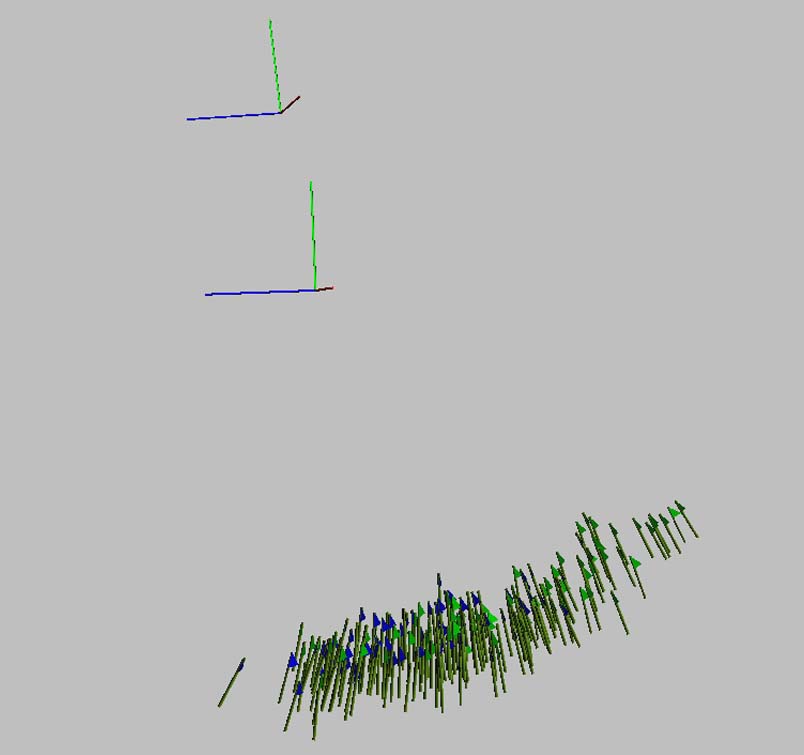}
  \caption[Object pose - 2 features]{}
\label{plot6}
  \vspace{10mm}
\centering
  \includegraphics[width=0.4\textwidth]{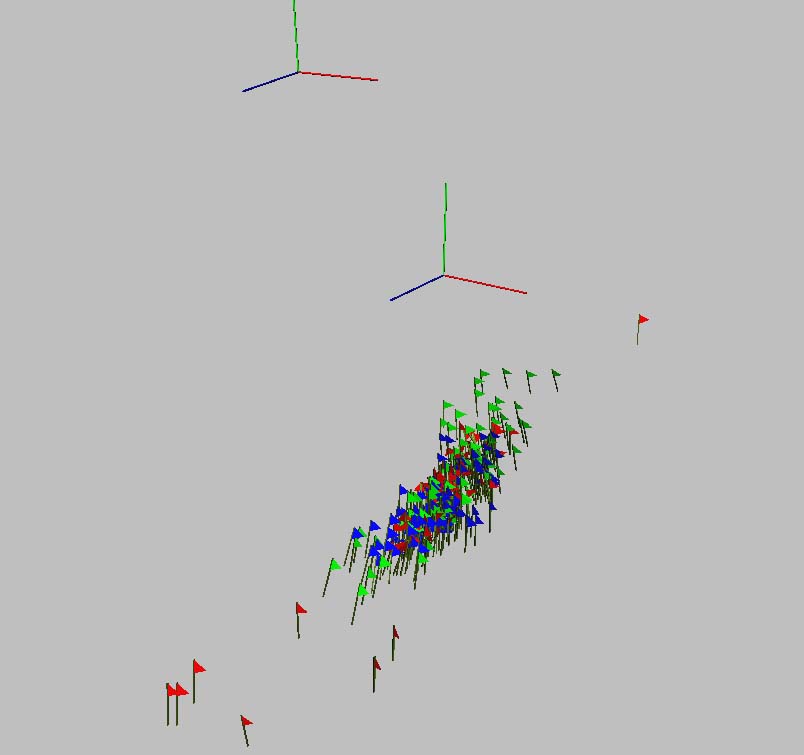}
  \caption[Object pose after applying fusion]{}
\label{plot7}
  \vspace{10mm}
\centering
  \includegraphics[width=0.4\textwidth]{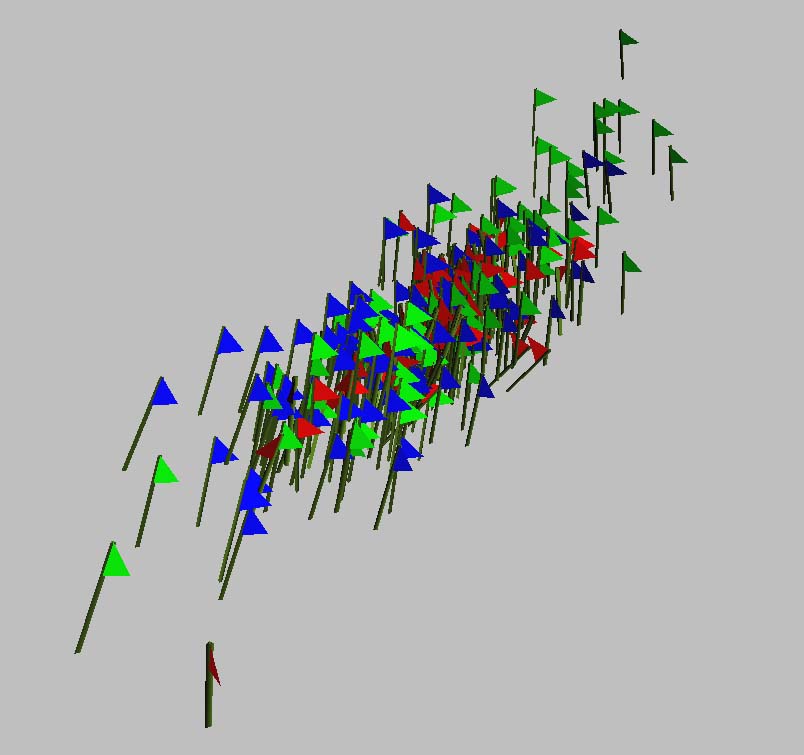}
  \caption[Object pose after dropping kernels]{}
\label{plot8}
\end{wrapfigure}

The coordinate systems which are located at the position of the mean vectors of the distribution describing the poses of the features 'B' and 'l' are rotated the way that the $z$-axes point to the camera. Further figure \ref{plot6} shows the big variance in direction of the $z$-axis of the camera which equals the viewing direction of the camera.\\

Now we apply the modification operation fusion to the green and the blue mixture. As both distribution functions provide reasonable sample sets this is an ordinary step of the evaluation algorithm. Figure \ref{plot7} shows the new red sample set which is obtained after fusing the other mixtures. The adjustment of the weights which are used in the operation of fusing might be improved to avoid scattering of the red samples.\\

Through this operation the number of elements of the mixture grew to 49 as the blue and the green mixture consisted of 7 kernels each. Many summands of the mixture have low weight but scatter and thus disturb the evaluation algorithm. This is the reason why I decided to drop 10 kernels with low weight. Figure \ref{plot8} shows that as result the red samples become more concentrated.\\

\begin{wrapfigure}{l}{7.5cm}
\centering
  \includegraphics[width=0.4\textwidth]{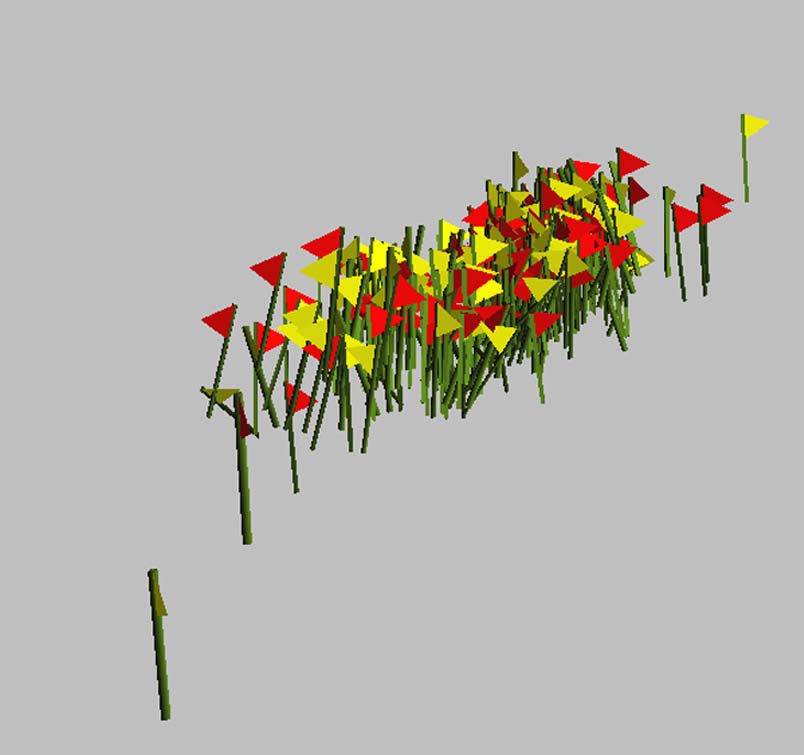}
  \caption[Comparison before and after merge]{}
\label{plot9}
  \vspace{10mm}
\centering
  \includegraphics[width=0.4\textwidth]{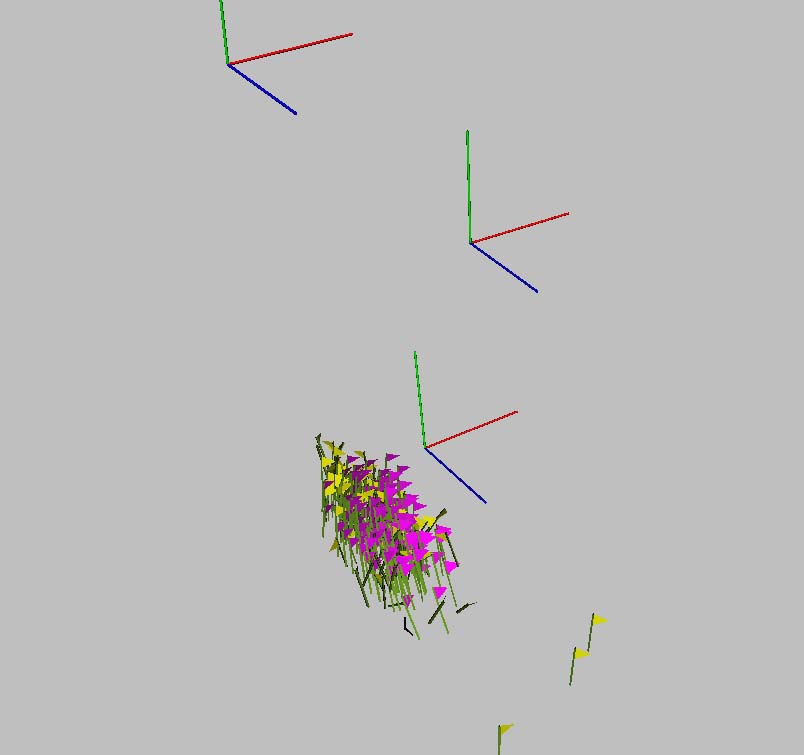}
  \caption[New localization attempt]{}
\label{plot10}
  \vspace{10mm}
\centering
  \includegraphics[width=0.4\textwidth]{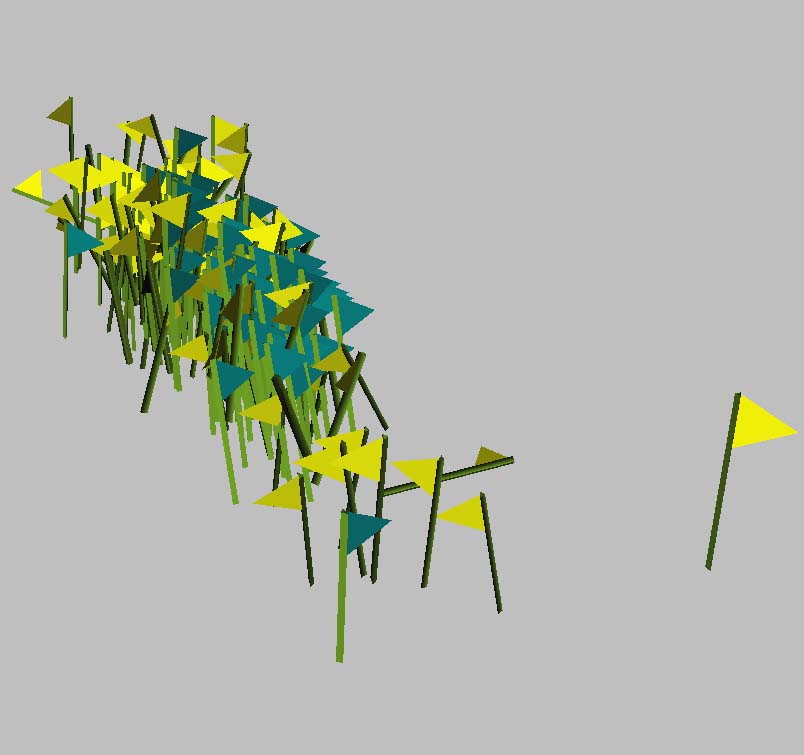}
  \caption[Object pose after applying merge]{}
\label{plot11}
\end{wrapfigure}

In figure \ref{plot9} the red samples still represent the fused and reduced mixture which is obtained already in the figure before. The yellow sample set are drawn from the mixture after merging the mixture down until the number of base elements that remain is 10. This smaller mixture is computationally efficient and doesn't seem to lack any information. That the flags are slightly different comes from the fact that the sample points are chosen randomly. \\

As the results we got by now seem reliable we will go on with the yellow mixture we obtained after the modification operation of mering. An appropriate next step for the evaluating algorithm would be to make a new localization attempt to get more information data. Figure \ref{plot10} shows the pink samples which are drawn from the mixture one obtains from drawing conclusion from the feature pose of the mountain top to the object pose.\\

Figure \ref{plot11} shows the turquoise sample set of the mixture which is obtained after fusing the pink and the yellow mixtures and further dropping the irrelevant base elements with low weights as well. This mixture still consists of 36 base elements.\\

\begin{wrapfigure}{l}{7.5cm}
\centering
  \includegraphics[width=0.4\textwidth]{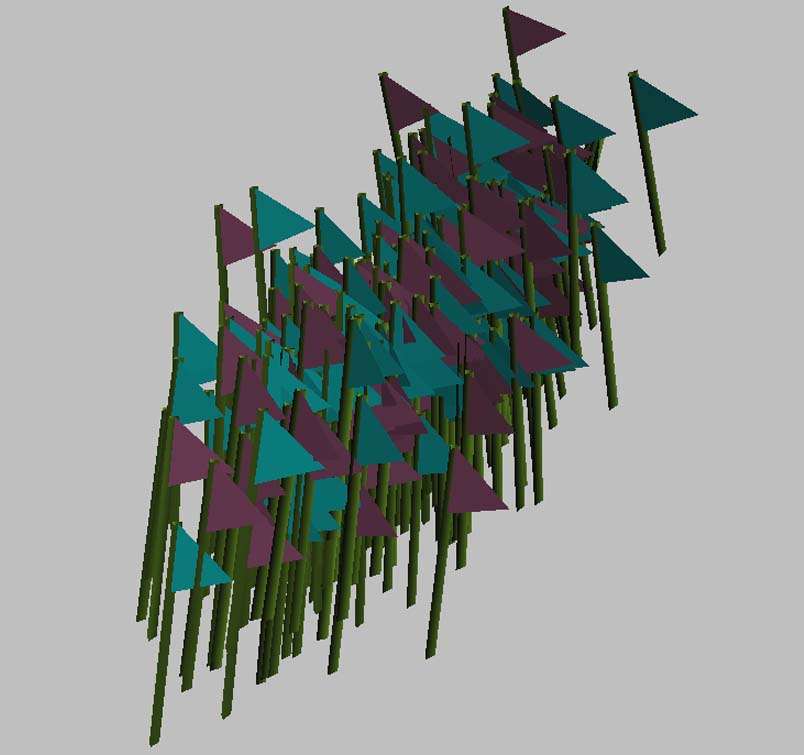}
  \caption[Comparison before and after merge]{}
\label{plot12}
  \vspace{10mm}
\centering
  \includegraphics[width=0.4\textwidth]{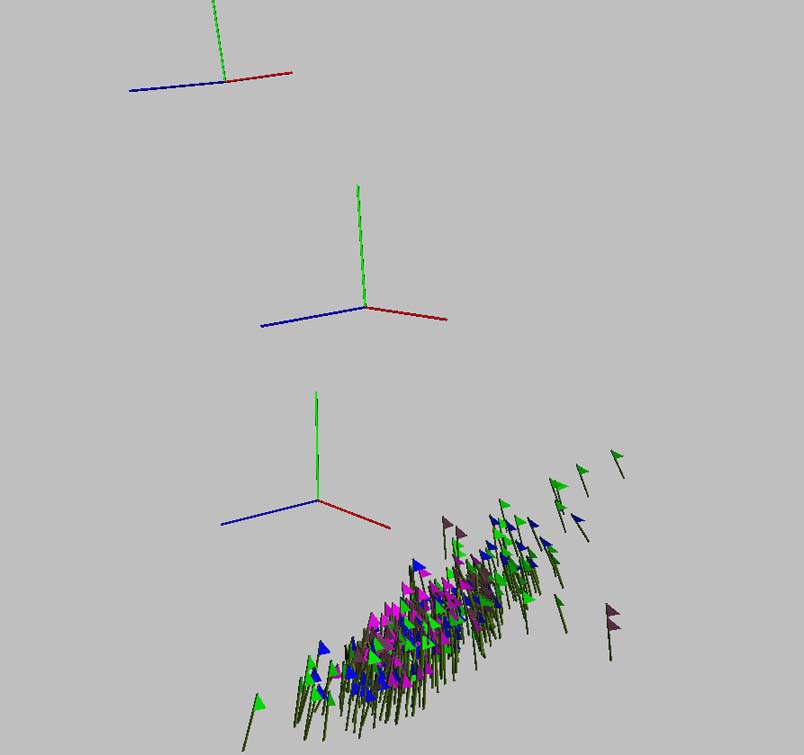}
  \caption[New localization attempt]{}
\label{plot13}
  \vspace{10mm}
\centering
  \includegraphics[width=0.4\textwidth]{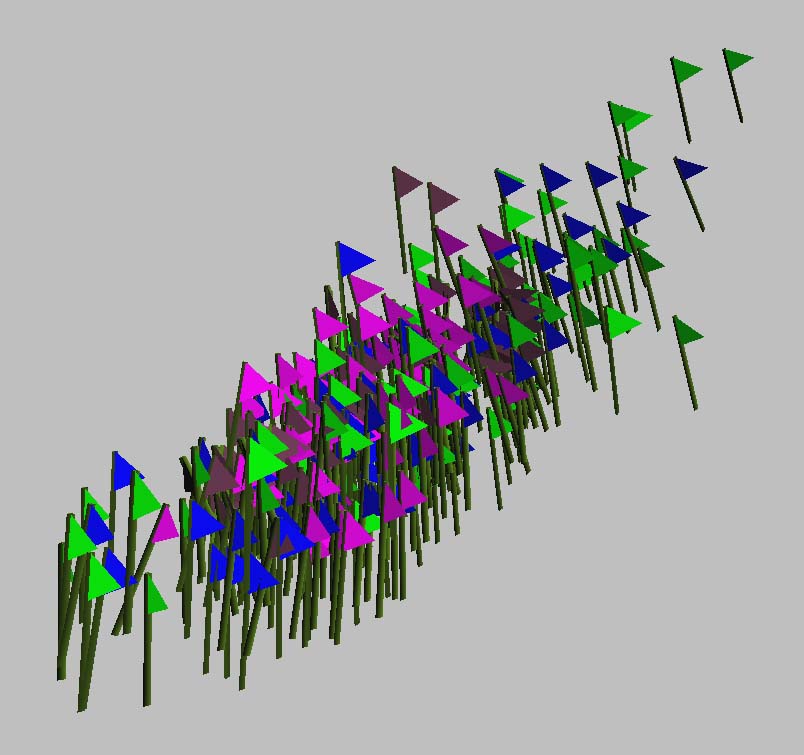}
  \caption[Object pose after applying merge]{}
\label{plot14}
\end{wrapfigure}

A remarkable effect of making the new localization attempt is that the fused sample set is way better ordered in orientation than before. It is seen in figure \ref{plot12} that from merging the turquoise mixture down to the new purple mixture consisting of 10 summands no loss of information arises. The mixture is strongly peaked already and estimates the true pose of the salt box in $S_3\times\R^3$.\\

Finally the last two figures \ref{plot13} and \ref{plot14} show the distribution of the object pose that is estimated directly from the three features represented by the green, blue and pink sample set and the distribution after simulating an evaluation algorithm. The purple sample set containing all information data fits the center of mass of the other mixtures and furthermore provides a peaked distribution function for the pose estimation. \\
From this example one gets a sense of how the evaluation algorithm should work. The steps I introduced need to be repeated until a stop criterion is reached, like i.e. that a grasp criterion is fulfilled. If no information gain can be registered any more possible further steps might be to involve other kinds of features like edge extraction or to try a new localization attempt from another side.\\

\chapter{Outlook}\label{outlook}

\section{Result}

\section{Improvements}

\begin{itemize}
\item In \ref{PG} Projected Gaussian I explain how a basis $B_0$ of the $\R^4$ can be calculated in a canonical way. I guess that there is also a canonical way to construct a basis over any odd dimension. This stands out to be proven.
\item It would be nice to implement a variation of our framework using the Bingham distribution to compare the performance and the time needed for the computations. Further the accuracy of the modeling by Binghams and projected Gaussians stands out to be compared. The composition of Bingham distributions doesn't give another Bingham but can be approximated by one. The result of this approximation should be compared with a composed projected Gaussian distribution, as well. 
\item By now I can't constitute which dissimilarity measure is appropriate for specific applications. Experiments are necessary to get informations about what grasp criterion and which distance measure should be applied.
\item It remains to check whether the KL divergence or the sKL divergence is faster to calculate. As in 
$$B(i,j) = \frac{1}{2} \left( (\lambda_i+\lambda_j)\log \det(\Sigma_{ij})-\log \det(\Sigma_{i})-\log \det(\Sigma_{j})\right)$$
one has to calculate three logarithms the computing time is very slow. For that reason I assume that the much longer expression $B_s(i,j)$ might be slightly faster. To verify this the number of necessary operations need to be determined for each formula.
\item The studies of the coherence between particle sets and MoPGs should be finished. This means the comparison of the results one achieves by applying a modification operation to a mixture and to a particle set.
\item The approaches to reduce the number of elements of a mixture introduced in \ref{AA} need to be developed and a check for the accuracy of the results has to be implemented.
\item In future work experimental results would be desirable. Therefore the evaluation algorithm has to be developed. Then further simulated experiments are necessary as well as real world experiments to validate the truth of the algorithm.
\item The implemented framework was planned to be open to any dimensions but in the course of the work I realized that this requires too many case-by-case analysis. Thus I decided to implement a part of the functions just for the 4-dimensional case. This could be universalized. 
\end{itemize}


\newpage
\begin{appendix}

\chapter{Appendix}

In \ref{A1} the complete Python code is appended usually including the input and output format for each function. Except in the obvious case. 

\section{}\label{A1}

For the code the following conventions were made:
\begin{itemize}
 \item $(\cdot,\ldots, \cdot)$ denotes vectors
 \item $[\cdot, \cdot,\cdot,\cdot,]$ stands for quaternions and lists
 \item $[\cdot,\cdot] = [[\cdot, \cdot,\cdot,\cdot,],[\cdot, \cdot,\cdot,\cdot,]]$ denotes a dual quaternion and
 \item $[[\cdot, \cdot],\ldots, [\cdot, \cdot]]$ is a mixture of base elements or equivalently a double list
\end{itemize}

Further the following modules need to be installed and imported:
\begin{itemize}
 \item numpy
 \item numpy.linalg
 \item numpy.random
 \item scipy
 \item math
 \item pyviblib.calc.common
 \item scipy.integrate.quadpack
\end{itemize}

The following code consists of auxiliary functions that don't belong to any specific class. These are basic rules that are needed for the methods of the classes I introduce already in \ref{code}. 


\makeatletter
\def\PY@reset{\let\PY@it=\relax \let\PY@bf=\relax%
    \let\PY@ul=\relax \let\PY@tc=\relax%
    \let\PY@bc=\relax \let\PY@ff=\relax}
\def\PY@tok#1{\csname PY@tok@#1\endcsname}
\def\PY@toks#1+{\ifx\relax#1\empty\else%
    \PY@tok{#1}\expandafter\PY@toks\fi}
\def\PY@do#1{\PY@bc{\PY@tc{\PY@ul{%
    \PY@it{\PY@bf{\PY@ff{#1}}}}}}}
\def\PY#1#2{\PY@reset\PY@toks#1+\relax+\PY@do{#2}}

\def\PY@tok@gd{\def\PY@tc##1{\textcolor[rgb]{0.63,0.00,0.00}{##1}}}
\def\PY@tok@gu{\let\PY@bf=\textbf\def\PY@tc##1{\textcolor[rgb]{0.50,0.00,0.50}{##1}}}
\def\PY@tok@gt{\def\PY@tc##1{\textcolor[rgb]{0.00,0.25,0.82}{##1}}}
\def\PY@tok@gs{\let\PY@bf=\textbf}
\def\PY@tok@gr{\def\PY@tc##1{\textcolor[rgb]{1.00,0.00,0.00}{##1}}}
\def\PY@tok@cm{\let\PY@it=\textit\def\PY@tc##1{\textcolor[rgb]{0.25,0.50,0.50}{##1}}}
\def\PY@tok@vg{\def\PY@tc##1{\textcolor[rgb]{0.10,0.09,0.49}{##1}}}
\def\PY@tok@m{\def\PY@tc##1{\textcolor[rgb]{0.40,0.40,0.40}{##1}}}
\def\PY@tok@mh{\def\PY@tc##1{\textcolor[rgb]{0.40,0.40,0.40}{##1}}}
\def\PY@tok@go{\def\PY@tc##1{\textcolor[rgb]{0.50,0.50,0.50}{##1}}}
\def\PY@tok@ge{\let\PY@it=\textit}
\def\PY@tok@vc{\def\PY@tc##1{\textcolor[rgb]{0.10,0.09,0.49}{##1}}}
\def\PY@tok@il{\def\PY@tc##1{\textcolor[rgb]{0.40,0.40,0.40}{##1}}}
\def\PY@tok@cs{\let\PY@it=\textit\def\PY@tc##1{\textcolor[rgb]{0.25,0.50,0.50}{##1}}}
\def\PY@tok@cp{\def\PY@tc##1{\textcolor[rgb]{0.74,0.48,0.00}{##1}}}
\def\PY@tok@gi{\def\PY@tc##1{\textcolor[rgb]{0.00,0.63,0.00}{##1}}}
\def\PY@tok@gh{\let\PY@bf=\textbf\def\PY@tc##1{\textcolor[rgb]{0.00,0.00,0.50}{##1}}}
\def\PY@tok@ni{\let\PY@bf=\textbf\def\PY@tc##1{\textcolor[rgb]{0.60,0.60,0.60}{##1}}}
\def\PY@tok@nl{\def\PY@tc##1{\textcolor[rgb]{0.63,0.63,0.00}{##1}}}
\def\PY@tok@nn{\let\PY@bf=\textbf\def\PY@tc##1{\textcolor[rgb]{0.00,0.00,1.00}{##1}}}
\def\PY@tok@no{\def\PY@tc##1{\textcolor[rgb]{0.53,0.00,0.00}{##1}}}
\def\PY@tok@na{\def\PY@tc##1{\textcolor[rgb]{0.49,0.56,0.16}{##1}}}
\def\PY@tok@nb{\def\PY@tc##1{\textcolor[rgb]{0.00,0.50,0.00}{##1}}}
\def\PY@tok@nc{\let\PY@bf=\textbf\def\PY@tc##1{\textcolor[rgb]{0.00,0.00,1.00}{##1}}}
\def\PY@tok@nd{\def\PY@tc##1{\textcolor[rgb]{0.67,0.13,1.00}{##1}}}
\def\PY@tok@ne{\let\PY@bf=\textbf\def\PY@tc##1{\textcolor[rgb]{0.82,0.25,0.23}{##1}}}
\def\PY@tok@nf{\def\PY@tc##1{\textcolor[rgb]{0.00,0.00,1.00}{##1}}}
\def\PY@tok@si{\let\PY@bf=\textbf\def\PY@tc##1{\textcolor[rgb]{0.73,0.40,0.53}{##1}}}
\def\PY@tok@s2{\def\PY@tc##1{\textcolor[rgb]{0.73,0.13,0.13}{##1}}}
\def\PY@tok@vi{\def\PY@tc##1{\textcolor[rgb]{0.10,0.09,0.49}{##1}}}
\def\PY@tok@nt{\let\PY@bf=\textbf\def\PY@tc##1{\textcolor[rgb]{0.00,0.50,0.00}{##1}}}
\def\PY@tok@nv{\def\PY@tc##1{\textcolor[rgb]{0.10,0.09,0.49}{##1}}}
\def\PY@tok@s1{\def\PY@tc##1{\textcolor[rgb]{0.73,0.13,0.13}{##1}}}
\def\PY@tok@sh{\def\PY@tc##1{\textcolor[rgb]{0.73,0.13,0.13}{##1}}}
\def\PY@tok@sc{\def\PY@tc##1{\textcolor[rgb]{0.73,0.13,0.13}{##1}}}
\def\PY@tok@sx{\def\PY@tc##1{\textcolor[rgb]{0.00,0.50,0.00}{##1}}}
\def\PY@tok@bp{\def\PY@tc##1{\textcolor[rgb]{0.00,0.50,0.00}{##1}}}
\def\PY@tok@c1{\let\PY@it=\textit\def\PY@tc##1{\textcolor[rgb]{0.25,0.50,0.50}{##1}}}
\def\PY@tok@kc{\let\PY@bf=\textbf\def\PY@tc##1{\textcolor[rgb]{0.00,0.50,0.00}{##1}}}
\def\PY@tok@c{\let\PY@it=\textit\def\PY@tc##1{\textcolor[rgb]{0.25,0.50,0.50}{##1}}}
\def\PY@tok@mf{\def\PY@tc##1{\textcolor[rgb]{0.40,0.40,0.40}{##1}}}
\def\PY@tok@err{\def\PY@bc##1{\fcolorbox[rgb]{1.00,0.00,0.00}{1,1,1}{##1}}}
\def\PY@tok@kd{\let\PY@bf=\textbf\def\PY@tc##1{\textcolor[rgb]{0.00,0.50,0.00}{##1}}}
\def\PY@tok@ss{\def\PY@tc##1{\textcolor[rgb]{0.10,0.09,0.49}{##1}}}
\def\PY@tok@sr{\def\PY@tc##1{\textcolor[rgb]{0.73,0.40,0.53}{##1}}}
\def\PY@tok@mo{\def\PY@tc##1{\textcolor[rgb]{0.40,0.40,0.40}{##1}}}
\def\PY@tok@kn{\let\PY@bf=\textbf\def\PY@tc##1{\textcolor[rgb]{0.00,0.50,0.00}{##1}}}
\def\PY@tok@mi{\def\PY@tc##1{\textcolor[rgb]{0.40,0.40,0.40}{##1}}}
\def\PY@tok@gp{\let\PY@bf=\textbf\def\PY@tc##1{\textcolor[rgb]{0.00,0.00,0.50}{##1}}}
\def\PY@tok@o{\def\PY@tc##1{\textcolor[rgb]{0.40,0.40,0.40}{##1}}}
\def\PY@tok@kr{\let\PY@bf=\textbf\def\PY@tc##1{\textcolor[rgb]{0.00,0.50,0.00}{##1}}}
\def\PY@tok@s{\def\PY@tc##1{\textcolor[rgb]{0.73,0.13,0.13}{##1}}}
\def\PY@tok@kp{\def\PY@tc##1{\textcolor[rgb]{0.00,0.50,0.00}{##1}}}
\def\PY@tok@w{\def\PY@tc##1{\textcolor[rgb]{0.73,0.73,0.73}{##1}}}
\def\PY@tok@kt{\def\PY@tc##1{\textcolor[rgb]{0.69,0.00,0.25}{##1}}}
\def\PY@tok@ow{\let\PY@bf=\textbf\def\PY@tc##1{\textcolor[rgb]{0.67,0.13,1.00}{##1}}}
\def\PY@tok@sb{\def\PY@tc##1{\textcolor[rgb]{0.73,0.13,0.13}{##1}}}
\def\PY@tok@k{\let\PY@bf=\textbf\def\PY@tc##1{\textcolor[rgb]{0.00,0.50,0.00}{##1}}}
\def\PY@tok@se{\let\PY@bf=\textbf\def\PY@tc##1{\textcolor[rgb]{0.73,0.40,0.13}{##1}}}
\def\PY@tok@sd{\let\PY@it=\textit\def\PY@tc##1{\textcolor[rgb]{0.73,0.13,0.13}{##1}}}

\def\PYZbs{\char`\\}
\def\PYZus{\char`\_}
\def\PYZob{\char`\{}
\def\PYZcb{\char`\}}
\def\PYZca{\char`\^}
\def\PYZat{@}
\def\PYZlb{[}
\def\PYZrb{]}
\makeatother



The class defining the tangent space contains the following methods.

\makeatletter
\def\PY@reset{\let\PY@it=\relax \let\PY@bf=\relax%
    \let\PY@ul=\relax \let\PY@tc=\relax%
    \let\PY@bc=\relax \let\PY@ff=\relax}
\def\PY@tok#1{\csname PY@tok@#1\endcsname}
\def\PY@toks#1+{\ifx\relax#1\empty\else%
    \PY@tok{#1}\expandafter\PY@toks\fi}
\def\PY@do#1{\PY@bc{\PY@tc{\PY@ul{%
    \PY@it{\PY@bf{\PY@ff{#1}}}}}}}
\def\PY#1#2{\PY@reset\PY@toks#1+\relax+\PY@do{#2}}

\def\PY@tok@gd{\def\PY@tc##1{\textcolor[rgb]{0.63,0.00,0.00}{##1}}}
\def\PY@tok@gu{\let\PY@bf=\textbf\def\PY@tc##1{\textcolor[rgb]{0.50,0.00,0.50}{##1}}}
\def\PY@tok@gt{\def\PY@tc##1{\textcolor[rgb]{0.00,0.25,0.82}{##1}}}
\def\PY@tok@gs{\let\PY@bf=\textbf}
\def\PY@tok@gr{\def\PY@tc##1{\textcolor[rgb]{1.00,0.00,0.00}{##1}}}
\def\PY@tok@cm{\let\PY@it=\textit\def\PY@tc##1{\textcolor[rgb]{0.25,0.50,0.50}{##1}}}
\def\PY@tok@vg{\def\PY@tc##1{\textcolor[rgb]{0.10,0.09,0.49}{##1}}}
\def\PY@tok@m{\def\PY@tc##1{\textcolor[rgb]{0.40,0.40,0.40}{##1}}}
\def\PY@tok@mh{\def\PY@tc##1{\textcolor[rgb]{0.40,0.40,0.40}{##1}}}
\def\PY@tok@go{\def\PY@tc##1{\textcolor[rgb]{0.50,0.50,0.50}{##1}}}
\def\PY@tok@ge{\let\PY@it=\textit}
\def\PY@tok@vc{\def\PY@tc##1{\textcolor[rgb]{0.10,0.09,0.49}{##1}}}
\def\PY@tok@il{\def\PY@tc##1{\textcolor[rgb]{0.40,0.40,0.40}{##1}}}
\def\PY@tok@cs{\let\PY@it=\textit\def\PY@tc##1{\textcolor[rgb]{0.25,0.50,0.50}{##1}}}
\def\PY@tok@cp{\def\PY@tc##1{\textcolor[rgb]{0.74,0.48,0.00}{##1}}}
\def\PY@tok@gi{\def\PY@tc##1{\textcolor[rgb]{0.00,0.63,0.00}{##1}}}
\def\PY@tok@gh{\let\PY@bf=\textbf\def\PY@tc##1{\textcolor[rgb]{0.00,0.00,0.50}{##1}}}
\def\PY@tok@ni{\let\PY@bf=\textbf\def\PY@tc##1{\textcolor[rgb]{0.60,0.60,0.60}{##1}}}
\def\PY@tok@nl{\def\PY@tc##1{\textcolor[rgb]{0.63,0.63,0.00}{##1}}}
\def\PY@tok@nn{\let\PY@bf=\textbf\def\PY@tc##1{\textcolor[rgb]{0.00,0.00,1.00}{##1}}}
\def\PY@tok@no{\def\PY@tc##1{\textcolor[rgb]{0.53,0.00,0.00}{##1}}}
\def\PY@tok@na{\def\PY@tc##1{\textcolor[rgb]{0.49,0.56,0.16}{##1}}}
\def\PY@tok@nb{\def\PY@tc##1{\textcolor[rgb]{0.00,0.50,0.00}{##1}}}
\def\PY@tok@nc{\let\PY@bf=\textbf\def\PY@tc##1{\textcolor[rgb]{0.00,0.00,1.00}{##1}}}
\def\PY@tok@nd{\def\PY@tc##1{\textcolor[rgb]{0.67,0.13,1.00}{##1}}}
\def\PY@tok@ne{\let\PY@bf=\textbf\def\PY@tc##1{\textcolor[rgb]{0.82,0.25,0.23}{##1}}}
\def\PY@tok@nf{\def\PY@tc##1{\textcolor[rgb]{0.00,0.00,1.00}{##1}}}
\def\PY@tok@si{\let\PY@bf=\textbf\def\PY@tc##1{\textcolor[rgb]{0.73,0.40,0.53}{##1}}}
\def\PY@tok@s2{\def\PY@tc##1{\textcolor[rgb]{0.73,0.13,0.13}{##1}}}
\def\PY@tok@vi{\def\PY@tc##1{\textcolor[rgb]{0.10,0.09,0.49}{##1}}}
\def\PY@tok@nt{\let\PY@bf=\textbf\def\PY@tc##1{\textcolor[rgb]{0.00,0.50,0.00}{##1}}}
\def\PY@tok@nv{\def\PY@tc##1{\textcolor[rgb]{0.10,0.09,0.49}{##1}}}
\def\PY@tok@s1{\def\PY@tc##1{\textcolor[rgb]{0.73,0.13,0.13}{##1}}}
\def\PY@tok@sh{\def\PY@tc##1{\textcolor[rgb]{0.73,0.13,0.13}{##1}}}
\def\PY@tok@sc{\def\PY@tc##1{\textcolor[rgb]{0.73,0.13,0.13}{##1}}}
\def\PY@tok@sx{\def\PY@tc##1{\textcolor[rgb]{0.00,0.50,0.00}{##1}}}
\def\PY@tok@bp{\def\PY@tc##1{\textcolor[rgb]{0.00,0.50,0.00}{##1}}}
\def\PY@tok@c1{\let\PY@it=\textit\def\PY@tc##1{\textcolor[rgb]{0.25,0.50,0.50}{##1}}}
\def\PY@tok@kc{\let\PY@bf=\textbf\def\PY@tc##1{\textcolor[rgb]{0.00,0.50,0.00}{##1}}}
\def\PY@tok@c{\let\PY@it=\textit\def\PY@tc##1{\textcolor[rgb]{0.25,0.50,0.50}{##1}}}
\def\PY@tok@mf{\def\PY@tc##1{\textcolor[rgb]{0.40,0.40,0.40}{##1}}}
\def\PY@tok@err{\def\PY@bc##1{\fcolorbox[rgb]{1.00,0.00,0.00}{1,1,1}{##1}}}
\def\PY@tok@kd{\let\PY@bf=\textbf\def\PY@tc##1{\textcolor[rgb]{0.00,0.50,0.00}{##1}}}
\def\PY@tok@ss{\def\PY@tc##1{\textcolor[rgb]{0.10,0.09,0.49}{##1}}}
\def\PY@tok@sr{\def\PY@tc##1{\textcolor[rgb]{0.73,0.40,0.53}{##1}}}
\def\PY@tok@mo{\def\PY@tc##1{\textcolor[rgb]{0.40,0.40,0.40}{##1}}}
\def\PY@tok@kn{\let\PY@bf=\textbf\def\PY@tc##1{\textcolor[rgb]{0.00,0.50,0.00}{##1}}}
\def\PY@tok@mi{\def\PY@tc##1{\textcolor[rgb]{0.40,0.40,0.40}{##1}}}
\def\PY@tok@gp{\let\PY@bf=\textbf\def\PY@tc##1{\textcolor[rgb]{0.00,0.00,0.50}{##1}}}
\def\PY@tok@o{\def\PY@tc##1{\textcolor[rgb]{0.40,0.40,0.40}{##1}}}
\def\PY@tok@kr{\let\PY@bf=\textbf\def\PY@tc##1{\textcolor[rgb]{0.00,0.50,0.00}{##1}}}
\def\PY@tok@s{\def\PY@tc##1{\textcolor[rgb]{0.73,0.13,0.13}{##1}}}
\def\PY@tok@kp{\def\PY@tc##1{\textcolor[rgb]{0.00,0.50,0.00}{##1}}}
\def\PY@tok@w{\def\PY@tc##1{\textcolor[rgb]{0.73,0.73,0.73}{##1}}}
\def\PY@tok@kt{\def\PY@tc##1{\textcolor[rgb]{0.69,0.00,0.25}{##1}}}
\def\PY@tok@ow{\let\PY@bf=\textbf\def\PY@tc##1{\textcolor[rgb]{0.67,0.13,1.00}{##1}}}
\def\PY@tok@sb{\def\PY@tc##1{\textcolor[rgb]{0.73,0.13,0.13}{##1}}}
\def\PY@tok@k{\let\PY@bf=\textbf\def\PY@tc##1{\textcolor[rgb]{0.00,0.50,0.00}{##1}}}
\def\PY@tok@se{\let\PY@bf=\textbf\def\PY@tc##1{\textcolor[rgb]{0.73,0.40,0.13}{##1}}}
\def\PY@tok@sd{\let\PY@it=\textit\def\PY@tc##1{\textcolor[rgb]{0.73,0.13,0.13}{##1}}}

\def\PYZbs{\char`\\}
\def\PYZus{\char`\_}
\def\PYZob{\char`\{}
\def\PYZcb{\char`\}}
\def\PYZca{\char`\^}
\def\PYZat{@}
\def\PYZlb{[}
\def\PYZrb{]}
\makeatother

\begin{Verbatim}[commandchars=\\\{\}, fontsize = \footnotesize]
\PY{k}{class} \PY{n+nc}{tangentSpace}\PY{p}{:}
    \PY{l+s+sd}{"""tangentSpace is of the from [point, basis]. point is a vector and}
\PY{l+s+sd}{    has the dimension d and basis is a d x (d-1)-matrix. Basis consists}
\PY{l+s+sd}{    of column vectors but as the input in Python in line wise one has to}
\PY{l+s+sd}{    make a vector in the common sense out of the input list."""}
    \PY{k}{def} \PY{n+nf}{\PYZus{}\PYZus{}init\PYZus{}\PYZus{}}\PY{p}{(}\PY{n+nb+bp}{self}\PY{p}{,} \PY{n}{tangentPoint}\PY{p}{)}\PY{p}{:}
        \PY{k}{assert} \PY{n+nb}{isinstance}\PY{p}{(}\PY{n}{tangentPoint}\PY{p}{,} \PY{n}{matrix}\PY{p}{)}
        \PY{n}{pl} \PY{o}{=} \PY{n}{tangentPoint}
        \PY{n}{one} \PY{o}{=} \PY{n}{sqrt}\PY{p}{(}\PY{n}{pl}\PY{p}{[}\PY{l+m+mi}{0}\PY{p}{]}\PY{o}{*}\PY{o}{*}\PY{l+m+mi}{2} \PY{o}{+} \PY{n}{pl}\PY{p}{[}\PY{l+m+mi}{1}\PY{p}{]}\PY{o}{*}\PY{o}{*}\PY{l+m+mi}{2} \PY{o}{+} \PY{n}{pl}\PY{p}{[}\PY{l+m+mi}{2}\PY{p}{]}\PY{o}{*}\PY{o}{*}\PY{l+m+mi}{2} \PY{o}{+} \PY{n}{pl}\PY{p}{[}\PY{l+m+mi}{3}\PY{p}{]}\PY{o}{*}\PY{o}{*}\PY{l+m+mi}{2}\PY{p}{)}
        \PY{k}{assert} \PY{n+nb}{abs}\PY{p}{(}\PY{l+m+mi}{1}\PY{o}{-}\PY{n}{one}\PY{p}{)} \PY{o}{<} \PY{l+m+mi}{10}\PY{o}{*}\PY{o}{*}\PY{p}{(}\PY{o}{-}\PY{l+m+mi}{4}\PY{p}{)}
        \PY{n+nb+bp}{self}\PY{o}{.}\PY{n}{p} \PY{o}{=} \PY{n}{pl}
        \PY{n}{basis} \PY{o}{=} \PY{n}{makeBase}\PY{p}{(}\PY{n}{tangentPoint}\PY{p}{)}
        \PY{n+nb+bp}{self}\PY{o}{.}\PY{n}{b} \PY{o}{=} \PY{n}{basis}

    \PY{k}{def} \PY{n+nf}{equal}\PY{p}{(}\PY{n+nb+bp}{self}\PY{p}{,} \PY{n}{newTS}\PY{p}{)}\PY{p}{:}
        \PY{n}{sp} \PY{o}{=} \PY{n+nb+bp}{self}\PY{o}{.}\PY{n}{p}
        \PY{n}{sb} \PY{o}{=} \PY{n+nb+bp}{self}\PY{o}{.}\PY{n}{b}
        \PY{n}{n} \PY{o}{=} \PY{n+nb}{len}\PY{p}{(}\PY{n}{sp}\PY{p}{)}
        \PY{n}{k} \PY{o}{=} \PY{l+m+mi}{0}
        \PY{k}{for} \PY{n}{i} \PY{o+ow}{in} \PY{n+nb}{range}\PY{p}{(}\PY{n}{n}\PY{p}{)}\PY{p}{:}
            \PY{k}{if} \PY{n}{sp}\PY{p}{[}\PY{n}{i}\PY{p}{]}\PY{p}{[}\PY{l+m+mi}{0}\PY{p}{]} \PY{o}{==} \PY{p}{(}\PY{n}{newTS}\PY{o}{.}\PY{n}{p}\PY{p}{)}\PY{p}{[}\PY{n}{i}\PY{p}{]}\PY{p}{[}\PY{l+m+mi}{0}\PY{p}{]}\PY{p}{:}
                \PY{k}{for} \PY{n}{j} \PY{o+ow}{in} \PY{n+nb}{range}\PY{p}{(}\PY{n}{n}\PY{o}{-}\PY{l+m+mi}{1}\PY{p}{)}\PY{p}{:}
                    \PY{n}{sl} \PY{o}{=} \PY{n}{sb}\PY{o}{.}\PY{n}{tolist}\PY{p}{(}\PY{p}{)}
                    \PY{n}{nb} \PY{o}{=} \PY{n}{newTS}\PY{o}{.}\PY{n}{b}
                    \PY{n}{nl} \PY{o}{=} \PY{n}{nb}\PY{o}{.}\PY{n}{tolist}\PY{p}{(}\PY{p}{)}
                    \PY{k}{if} \PY{n}{sl}\PY{p}{[}\PY{n}{i}\PY{p}{]}\PY{p}{[}\PY{n}{j}\PY{p}{]} \PY{o}{==} \PY{n}{nl}\PY{p}{[}\PY{n}{i}\PY{p}{]}\PY{p}{[}\PY{n}{j}\PY{p}{]}\PY{p}{:}
                        \PY{n}{k} \PY{o}{=} \PY{n}{k}\PY{o}{+}\PY{l+m+mi}{1}
        \PY{k}{if} \PY{n}{k} \PY{o}{==} \PY{n}{n}\PY{o}{*}\PY{p}{(}\PY{n}{n}\PY{o}{-}\PY{l+m+mi}{1}\PY{p}{)}\PY{p}{:}
            \PY{k}{print} \PY{n+nb+bp}{True}
            \PY{k}{return} \PY{n+nb+bp}{True}
        \PY{k}{else}\PY{p}{:}
            \PY{k}{print} \PY{n+nb+bp}{False}
            \PY{k}{return} \PY{n+nb+bp}{False}

    \PY{k}{def} \PY{n+nf}{display}\PY{p}{(}\PY{n+nb+bp}{self}\PY{p}{)}\PY{p}{:}
        \PY{l+s+sd}{"""Attention: THE OUTPUT HERE IS LINE WISE!}
\PY{l+s+sd}{        Thus the first row of the matrix is the first column vector}
\PY{l+s+sd}{        of the basis."""}
        \PY{n}{sp} \PY{o}{=} \PY{n+nb+bp}{self}\PY{o}{.}\PY{n}{p}
        \PY{n}{listp} \PY{o}{=} \PY{n}{sp}\PY{o}{.}\PY{n}{tolist}\PY{p}{(}\PY{p}{)}
        \PY{n}{sb} \PY{o}{=} \PY{n+nb+bp}{self}\PY{o}{.}\PY{n}{b}
        \PY{n}{listb} \PY{o}{=} \PY{n}{sb}\PY{o}{.}\PY{n}{tolist}\PY{p}{(}\PY{p}{)}
        \PY{c}{#print [list(flatten(listp)),listb]}
        \PY{k}{return} \PY{p}{[}\PY{n+nb}{list}\PY{p}{(}\PY{n}{flatten}\PY{p}{(}\PY{n}{listp}\PY{p}{)}\PY{p}{)}\PY{p}{,} \PY{n}{listb}\PY{p}{]}

    \PY{c}{#Achtung: Es wird eine ORTHONORMALBASIS benoetigt!}
    \PY{k}{def} \PY{n+nf}{tangentSpaceToSphereCentralProjection}\PY{p}{(}\PY{n+nb+bp}{self}\PY{p}{,} \PY{n}{tangentSpacePoint}\PY{p}{)}\PY{p}{:}
        \PY{l+s+sd}{"""tangentSpacePoint = [x1,x2,...,x(n-1)] is the point in the}
\PY{l+s+sd}{        tangential (hyper-) plane whose value shall be projected on the}
\PY{l+s+sd}{        (p-1)sphere. The first r entries of tangentSpacePoint are the}
\PY{l+s+sd}{        ones to describe the rotation, the last ones t = n-1-r describe}
\PY{l+s+sd}{        the translation and stay unchanged. tangentSpace = [p,B] is the}
\PY{l+s+sd}{        tangential (hyper-) plane with p the tangent point and B the}
\PY{l+s+sd}{        basis of the tangential (hyper-) plane. p has n dimensions.}
\PY{l+s+sd}{        The input of tangentSpacePoint has to be a list and tangentSpace}
\PY{l+s+sd}{        really has to be a tangentSpace!}
\PY{l+s+sd}{        The output is the wanted vector."""}
        \PY{n}{p} \PY{o}{=} \PY{n+nb+bp}{self}\PY{o}{.}\PY{n}{p}
        \PY{n}{B} \PY{o}{=} \PY{n+nb+bp}{self}\PY{o}{.}\PY{n}{b}
        \PY{n}{k} \PY{o}{=} \PY{n+nb}{len}\PY{p}{(}\PY{n}{p}\PY{p}{)}\PY{o}{-}\PY{l+m+mi}{1}
        \PY{n}{rvec} \PY{o}{=} \PY{n}{tangentSpacePoint}\PY{p}{[}\PY{p}{:}\PY{n}{k}\PY{p}{]}
        \PY{n}{rtvec} \PY{o}{=} \PY{n}{matrix}\PY{p}{(}\PY{n}{rvec}\PY{p}{)}\PY{o}{.}\PY{n}{T}
        \PY{n}{rtWorldCoord} \PY{o}{=} \PY{n}{B}\PY{o}{*}\PY{n}{rtvec} \PY{o}{+} \PY{n}{p}
        \PY{n}{normalizationFactor} \PY{o}{=} \PY{n}{linalg}\PY{o}{.}\PY{n}{norm}\PY{p}{(}\PY{n}{rtWorldCoord}\PY{p}{)}
        \PY{n}{svec} \PY{o}{=} \PY{n}{rtWorldCoord}\PY{o}{/}\PY{n}{normalizationFactor}
        \PY{n}{lsvec} \PY{o}{=} \PY{n}{svec}\PY{o}{.}\PY{n}{tolist}\PY{p}{(}\PY{p}{)}
        \PY{n}{tvec} \PY{o}{=} \PY{n}{tangentSpacePoint}\PY{p}{[}\PY{n}{k}\PY{p}{:}\PY{p}{]}
        \PY{n}{value} \PY{o}{=} \PY{p}{[}\PY{n}{lsvec}\PY{p}{,} \PY{n}{tvec}\PY{p}{]}
        \PY{n}{value} \PY{o}{=} \PY{n+nb}{list}\PY{p}{(}\PY{n}{flatten}\PY{p}{(}\PY{n}{value}\PY{p}{)}\PY{p}{)}
        \PY{c}{#print value}
        \PY{n}{value} \PY{o}{=} \PY{n}{makeVector}\PY{p}{(}\PY{n}{value}\PY{p}{)}
        \PY{k}{return} \PY{n}{value}


    \PY{k}{def} \PY{n+nf}{sphereToTangentSpaceCentralProjection}\PY{p}{(}\PY{n+nb+bp}{self}\PY{p}{,} \PY{n}{pointOnSphere}\PY{p}{)}\PY{p}{:}
        \PY{l+s+sd}{"""pointOnSphere = [q1,q2,...,qn] is the point to be projected}
\PY{l+s+sd}{        consisting of rotation and translation part.}
\PY{l+s+sd}{        tangentSpace = [p,B] is the tangential (hyper-) plane with p the}
\PY{l+s+sd}{        tangent point and B the basis of the tangential (hyper-) plane.}
\PY{l+s+sd}{        The input of pointOnSphere has to be a list and tangentSpace}
\PY{l+s+sd}{        really has to be a tangentSpace! The output is a vector in the}
\PY{l+s+sd}{        tangent space (including translation)."""}
        \PY{n}{p} \PY{o}{=} \PY{n+nb+bp}{self}\PY{o}{.}\PY{n}{p}
        \PY{n}{B} \PY{o}{=} \PY{n+nb+bp}{self}\PY{o}{.}\PY{n}{b}
        \PY{n}{Bt} \PY{o}{=} \PY{n}{B}\PY{o}{.}\PY{n}{T}
        \PY{n}{n} \PY{o}{=} \PY{n+nb}{len}\PY{p}{(}\PY{n}{p}\PY{p}{)}
        \PY{n}{qr} \PY{o}{=} \PY{n}{pointOnSphere}\PY{p}{[}\PY{p}{:}\PY{n}{n}\PY{p}{]}
        \PY{n}{qrT} \PY{o}{=} \PY{n}{matrix}\PY{p}{(}\PY{p}{[}\PY{n}{qr}\PY{p}{]}\PY{p}{)}\PY{o}{.}\PY{n}{T}
        \PY{n}{scalar1} \PY{o}{=} \PY{n}{qr}\PY{o}{*}\PY{n}{p}
        \PY{n}{scalar2} \PY{o}{=} \PY{n}{scalar1}\PY{o}{.}\PY{n}{tolist}\PY{p}{(}\PY{p}{)}
        \PY{n}{scalar3} \PY{o}{=} \PY{n}{scalar2}\PY{p}{[}\PY{l+m+mi}{0}\PY{p}{]}\PY{p}{[}\PY{l+m+mi}{0}\PY{p}{]}
        \PY{c}{#um aus der 1x1 matrix wirklich ein skalar zu machen....}
        \PY{n}{qrInTangentSpace} \PY{o}{=} \PY{n}{Bt}\PY{o}{*}\PY{p}{(}\PY{l+m+mf}{1.0}\PY{o}{/}\PY{p}{(}\PY{n}{scalar3}\PY{p}{)}\PY{o}{*}\PY{n}{qrT}\PY{o}{-}\PY{n}{p}\PY{p}{)}
        \PY{n}{qt} \PY{o}{=} \PY{n}{pointOnSphere}\PY{p}{[}\PY{n}{n}\PY{p}{:}\PY{p}{]}
        \PY{n}{lqr} \PY{o}{=} \PY{n}{qrInTangentSpace}\PY{o}{.}\PY{n}{tolist}\PY{p}{(}\PY{p}{)}
        \PY{n}{value} \PY{o}{=} \PY{p}{[}\PY{n}{lqr}\PY{p}{,} \PY{n}{qt}\PY{p}{]}
        \PY{n}{value} \PY{o}{=} \PY{n+nb}{list}\PY{p}{(}\PY{n}{flatten}\PY{p}{(}\PY{n}{value}\PY{p}{)}\PY{p}{)}
        \PY{c}{#print value}
        \PY{n}{value} \PY{o}{=} \PY{n}{makeVector}\PY{p}{(}\PY{n}{value}\PY{p}{)}
        \PY{k}{return} \PY{n}{value}


    \PY{k}{def} \PY{n+nf}{transformFromSelfToTS}\PY{p}{(}\PY{n+nb+bp}{self}\PY{p}{,} \PY{n}{point}\PY{p}{,} \PY{n}{tangentSpace\PYZus{}new}\PY{p}{)}\PY{p}{:}
        \PY{l+s+sd}{"""point = [x1,x2,...,xn] is the point in the old tangentSpace1}
\PY{l+s+sd}{        that shall be projected to a new tangentSpace.}
\PY{l+s+sd}{        self = [p1,B1] is the old tangent space.}
\PY{l+s+sd}{        tangentSpace\PYZus{}new = [p2,B2] is the new tangent space.}
\PY{l+s+sd}{        In the input 'point' has to be of the type list but the output}
\PY{l+s+sd}{        it is a vector."""}
        \PY{k}{assert} \PY{n+nb}{isinstance}\PY{p}{(}\PY{n}{tangentSpace\PYZus{}new}\PY{p}{,} \PY{n}{tangentSpace}\PY{p}{)}
        \PY{n}{Tn} \PY{o}{=} \PY{n}{tangentSpace\PYZus{}new}
        \PY{n}{pointOnSphere} \PY{o}{=} \PY{n+nb+bp}{self}\PY{o}{.}\PY{n}{tangentSpaceToSphereCentralProjection}\PY{p}{(}\PY{n}{point}\PY{p}{)}
        \PY{n}{listpoint} \PY{o}{=} \PY{n}{pointOnSphere}\PY{o}{.}\PY{n}{tolist}\PY{p}{(}\PY{p}{)}
        \PY{n}{lp} \PY{o}{=} \PY{n+nb}{list}\PY{p}{(}\PY{n}{flatten}\PY{p}{(}\PY{n}{listpoint}\PY{p}{)}\PY{p}{)}
        \PY{n}{value} \PY{o}{=}\PY{n}{tangentSpace\PYZus{}new}\PY{o}{.}\PY{n}{sphereToTangentSpaceCentralProjection}\PY{p}{(}\PY{n}{lp}\PY{p}{)}
        \PY{c}{#print value}
        \PY{k}{return} \PY{n}{value}


    \PY{k}{def} \PY{n+nf}{poseTransformationTS}\PY{p}{(}\PY{n+nb+bp}{self}\PY{p}{,} \PY{n}{pointold}\PY{p}{,} \PY{n}{transpoint}\PY{p}{,}
                             \PY{n}{tangentSpace\PYZus{}transform}\PY{p}{,} \PY{n}{tangentSpace\PYZus{}new}\PY{p}{)}\PY{p}{:}
        \PY{l+s+sd}{"""Input shall be vectors and tangent spaces.}
\PY{l+s+sd}{        I want the output to consist of vectors."""}
        \PY{k}{assert} \PY{n+nb}{isinstance}\PY{p}{(}\PY{n}{tangentSpace\PYZus{}transform}\PY{p}{,} \PY{n}{tangentSpace}\PY{p}{)}
        \PY{k}{assert} \PY{n+nb}{isinstance}\PY{p}{(}\PY{n}{tangentSpace\PYZus{}new}\PY{p}{,} \PY{n}{tangentSpace}\PY{p}{)}
        \PY{k}{assert} \PY{n+nb}{isinstance}\PY{p}{(}\PY{n}{pointold}\PY{p}{,} \PY{n}{matrix}\PY{p}{)}
        \PY{k}{assert} \PY{n+nb}{isinstance}\PY{p}{(}\PY{n}{transpoint}\PY{p}{,} \PY{n}{matrix}\PY{p}{)}
        \PY{n}{T\PYZus{}trans} \PY{o}{=} \PY{n}{tangentSpace\PYZus{}transform}
        \PY{n}{T\PYZus{}new} \PY{o}{=} \PY{n}{tangentSpace\PYZus{}new}
        \PY{n}{pt\PYZus{}list} \PY{o}{=} \PY{n+nb}{list}\PY{p}{(}\PY{n}{flatten}\PY{p}{(}\PY{n}{pointold}\PY{o}{.}\PY{n}{tolist}\PY{p}{(}\PY{p}{)}\PY{p}{)}\PY{p}{)}
        \PY{n}{tp\PYZus{}list} \PY{o}{=} \PY{n+nb}{list}\PY{p}{(}\PY{n}{flatten}\PY{p}{(}\PY{n}{transpoint}\PY{o}{.}\PY{n}{tolist}\PY{p}{(}\PY{p}{)}\PY{p}{)}\PY{p}{)}
        \PY{n}{pointsphere} \PY{o}{=} \PY{n+nb+bp}{self}\PY{o}{.}\PY{n}{tangentSpaceToSphereCentralProjection}\PY{p}{(}\PY{n}{pt\PYZus{}list}\PY{p}{)}
        \PY{n}{tpsphere} \PY{o}{=} \PY{n}{T\PYZus{}trans}\PY{o}{.}\PY{n}{tangentSpaceToSphereCentralProjection}\PY{p}{(}\PY{n}{tp\PYZus{}list}\PY{p}{)}
        \PY{n}{l} \PY{o}{=} \PY{n+nb}{list}\PY{p}{(}\PY{n}{flatten}\PY{p}{(}\PY{n}{pointsphere}\PY{o}{.}\PY{n}{tolist}\PY{p}{(}\PY{p}{)}\PY{p}{)}\PY{p}{)}
        \PY{n}{rot} \PY{o}{=} \PY{n}{quaternion}\PY{p}{(}\PY{n}{l}\PY{p}{[}\PY{l+m+mi}{0}\PY{p}{]}\PY{p}{,} \PY{n}{l}\PY{p}{[}\PY{l+m+mi}{1}\PY{p}{]}\PY{p}{,} \PY{n}{l}\PY{p}{[}\PY{l+m+mi}{2}\PY{p}{]}\PY{p}{,} \PY{n}{l}\PY{p}{[}\PY{l+m+mi}{3}\PY{p}{]}\PY{p}{)}
        \PY{n}{trans} \PY{o}{=} \PY{n}{quaternion}\PY{p}{(}\PY{l+m+mi}{0}\PY{p}{,} \PY{n}{l}\PY{p}{[}\PY{l+m+mi}{4}\PY{p}{]}\PY{p}{,} \PY{n}{l}\PY{p}{[}\PY{l+m+mi}{5}\PY{p}{]}\PY{p}{,} \PY{n}{l}\PY{p}{[}\PY{l+m+mi}{6}\PY{p}{]}\PY{p}{)}
        \PY{n}{rotTrans} \PY{o}{=} \PY{p}{[}\PY{n}{rot}\PY{p}{,} \PY{n}{trans}\PY{p}{]} \PY{c}{#entsteht aus pointold}
        \PY{n}{pointquat} \PY{o}{=} \PY{n}{transformationToDQ}\PY{p}{(}\PY{n}{rotTrans}\PY{p}{)}
        \PY{n}{li} \PY{o}{=} \PY{n+nb}{list}\PY{p}{(}\PY{n}{flatten}\PY{p}{(}\PY{n}{tpsphere}\PY{o}{.}\PY{n}{tolist}\PY{p}{(}\PY{p}{)}\PY{p}{)}\PY{p}{)}
        \PY{n}{trot} \PY{o}{=} \PY{n}{quaternion}\PY{p}{(}\PY{n}{li}\PY{p}{[}\PY{l+m+mi}{0}\PY{p}{]}\PY{p}{,} \PY{n}{li}\PY{p}{[}\PY{l+m+mi}{1}\PY{p}{]}\PY{p}{,} \PY{n}{li}\PY{p}{[}\PY{l+m+mi}{2}\PY{p}{]}\PY{p}{,} \PY{n}{li}\PY{p}{[}\PY{l+m+mi}{3}\PY{p}{]}\PY{p}{)}
        \PY{n}{ttrans} \PY{o}{=} \PY{n}{quaternion}\PY{p}{(}\PY{l+m+mi}{0}\PY{p}{,} \PY{n}{li}\PY{p}{[}\PY{l+m+mi}{4}\PY{p}{]}\PY{p}{,} \PY{n}{li}\PY{p}{[}\PY{l+m+mi}{5}\PY{p}{]}\PY{p}{,} \PY{n}{li}\PY{p}{[}\PY{l+m+mi}{6}\PY{p}{]}\PY{p}{)}
        \PY{n}{trotTrans} \PY{o}{=} \PY{p}{[}\PY{n}{trot}\PY{p}{,} \PY{n}{ttrans}\PY{p}{]} \PY{c}{#entsteht aus transpoint}
        \PY{n}{transpointquat} \PY{o}{=} \PY{n}{transformationToDQ}\PY{p}{(}\PY{n}{trotTrans}\PY{p}{)}
        \PY{n}{pointquat\PYZus{}new} \PY{o}{=} \PY{n}{transformPose}\PY{p}{(}\PY{n}{pointquat}\PY{p}{,} \PY{n}{transpointquat}\PY{p}{)}
        \PY{n}{ps\PYZus{}new} \PY{o}{=} \PY{n}{DQToTransformation}\PY{p}{(}\PY{n}{pointquat\PYZus{}new}\PY{p}{)}
        \PY{n}{rot} \PY{o}{=} \PY{n}{pointsphere\PYZus{}new}\PY{p}{[}\PY{l+m+mi}{0}\PY{p}{]}\PY{o}{.}\PY{n}{toList}\PY{p}{(}\PY{p}{)}
        \PY{n}{trans} \PY{o}{=} \PY{n}{pointsphere\PYZus{}new}\PY{p}{[}\PY{l+m+mi}{1}\PY{p}{]}\PY{o}{.}\PY{n}{toList}\PY{p}{(}\PY{p}{)}
        \PY{n}{pointsphere\PYZus{}new} \PY{o}{=} \PY{p}{[}\PY{n}{rot}\PY{p}{[}\PY{l+m+mi}{0}\PY{p}{]}\PY{p}{,}\PY{n}{rot}\PY{p}{[}\PY{l+m+mi}{1}\PY{p}{]}\PY{p}{,}\PY{n}{rot}\PY{p}{[}\PY{l+m+mi}{2}\PY{p}{]}\PY{p}{,}\PY{n}{rot}\PY{p}{[}\PY{l+m+mi}{3}\PY{p}{]}\PY{p}{,}\PY{n}{trans}\PY{p}{[}\PY{l+m+mi}{1}\PY{p}{]}\PY{p}{,}\PY{n}{trans}\PY{p}{[}\PY{l+m+mi}{2}\PY{p}{]}\PY{p}{,}
                           \PY{n}{trans}\PY{p}{[}\PY{l+m+mi}{3}\PY{p}{]}\PY{p}{]}
        \PY{n}{point\PYZus{}new} \PY{o}{=} \PY{n}{T\PYZus{}new}\PY{o}{.}\PY{n}{sphereToTangentSpaceCentralProjection}\PY{p}{(}\PY{n}{ps\PYZus{}new}\PY{p}{)}
        \PY{c}{#print point\PYZus{}new}
        \PY{k}{return} \PY{n}{point\PYZus{}new}
\end{Verbatim}

The next section is the code describing the base element.

\makeatletter
\def\PY@reset{\let\PY@it=\relax \let\PY@bf=\relax%
    \let\PY@ul=\relax \let\PY@tc=\relax%
    \let\PY@bc=\relax \let\PY@ff=\relax}
\def\PY@tok#1{\csname PY@tok@#1\endcsname}
\def\PY@toks#1+{\ifx\relax#1\empty\else%
    \PY@tok{#1}\expandafter\PY@toks\fi}
\def\PY@do#1{\PY@bc{\PY@tc{\PY@ul{%
    \PY@it{\PY@bf{\PY@ff{#1}}}}}}}
\def\PY#1#2{\PY@reset\PY@toks#1+\relax+\PY@do{#2}}

\def\PY@tok@gd{\def\PY@tc##1{\textcolor[rgb]{0.63,0.00,0.00}{##1}}}
\def\PY@tok@gu{\let\PY@bf=\textbf\def\PY@tc##1{\textcolor[rgb]{0.50,0.00,0.50}{##1}}}
\def\PY@tok@gt{\def\PY@tc##1{\textcolor[rgb]{0.00,0.25,0.82}{##1}}}
\def\PY@tok@gs{\let\PY@bf=\textbf}
\def\PY@tok@gr{\def\PY@tc##1{\textcolor[rgb]{1.00,0.00,0.00}{##1}}}
\def\PY@tok@cm{\let\PY@it=\textit\def\PY@tc##1{\textcolor[rgb]{0.25,0.50,0.50}{##1}}}
\def\PY@tok@vg{\def\PY@tc##1{\textcolor[rgb]{0.10,0.09,0.49}{##1}}}
\def\PY@tok@m{\def\PY@tc##1{\textcolor[rgb]{0.40,0.40,0.40}{##1}}}
\def\PY@tok@mh{\def\PY@tc##1{\textcolor[rgb]{0.40,0.40,0.40}{##1}}}
\def\PY@tok@go{\def\PY@tc##1{\textcolor[rgb]{0.50,0.50,0.50}{##1}}}
\def\PY@tok@ge{\let\PY@it=\textit}
\def\PY@tok@vc{\def\PY@tc##1{\textcolor[rgb]{0.10,0.09,0.49}{##1}}}
\def\PY@tok@il{\def\PY@tc##1{\textcolor[rgb]{0.40,0.40,0.40}{##1}}}
\def\PY@tok@cs{\let\PY@it=\textit\def\PY@tc##1{\textcolor[rgb]{0.25,0.50,0.50}{##1}}}
\def\PY@tok@cp{\def\PY@tc##1{\textcolor[rgb]{0.74,0.48,0.00}{##1}}}
\def\PY@tok@gi{\def\PY@tc##1{\textcolor[rgb]{0.00,0.63,0.00}{##1}}}
\def\PY@tok@gh{\let\PY@bf=\textbf\def\PY@tc##1{\textcolor[rgb]{0.00,0.00,0.50}{##1}}}
\def\PY@tok@ni{\let\PY@bf=\textbf\def\PY@tc##1{\textcolor[rgb]{0.60,0.60,0.60}{##1}}}
\def\PY@tok@nl{\def\PY@tc##1{\textcolor[rgb]{0.63,0.63,0.00}{##1}}}
\def\PY@tok@nn{\let\PY@bf=\textbf\def\PY@tc##1{\textcolor[rgb]{0.00,0.00,1.00}{##1}}}
\def\PY@tok@no{\def\PY@tc##1{\textcolor[rgb]{0.53,0.00,0.00}{##1}}}
\def\PY@tok@na{\def\PY@tc##1{\textcolor[rgb]{0.49,0.56,0.16}{##1}}}
\def\PY@tok@nb{\def\PY@tc##1{\textcolor[rgb]{0.00,0.50,0.00}{##1}}}
\def\PY@tok@nc{\let\PY@bf=\textbf\def\PY@tc##1{\textcolor[rgb]{0.00,0.00,1.00}{##1}}}
\def\PY@tok@nd{\def\PY@tc##1{\textcolor[rgb]{0.67,0.13,1.00}{##1}}}
\def\PY@tok@ne{\let\PY@bf=\textbf\def\PY@tc##1{\textcolor[rgb]{0.82,0.25,0.23}{##1}}}
\def\PY@tok@nf{\def\PY@tc##1{\textcolor[rgb]{0.00,0.00,1.00}{##1}}}
\def\PY@tok@si{\let\PY@bf=\textbf\def\PY@tc##1{\textcolor[rgb]{0.73,0.40,0.53}{##1}}}
\def\PY@tok@s2{\def\PY@tc##1{\textcolor[rgb]{0.73,0.13,0.13}{##1}}}
\def\PY@tok@vi{\def\PY@tc##1{\textcolor[rgb]{0.10,0.09,0.49}{##1}}}
\def\PY@tok@nt{\let\PY@bf=\textbf\def\PY@tc##1{\textcolor[rgb]{0.00,0.50,0.00}{##1}}}
\def\PY@tok@nv{\def\PY@tc##1{\textcolor[rgb]{0.10,0.09,0.49}{##1}}}
\def\PY@tok@s1{\def\PY@tc##1{\textcolor[rgb]{0.73,0.13,0.13}{##1}}}
\def\PY@tok@sh{\def\PY@tc##1{\textcolor[rgb]{0.73,0.13,0.13}{##1}}}
\def\PY@tok@sc{\def\PY@tc##1{\textcolor[rgb]{0.73,0.13,0.13}{##1}}}
\def\PY@tok@sx{\def\PY@tc##1{\textcolor[rgb]{0.00,0.50,0.00}{##1}}}
\def\PY@tok@bp{\def\PY@tc##1{\textcolor[rgb]{0.00,0.50,0.00}{##1}}}
\def\PY@tok@c1{\let\PY@it=\textit\def\PY@tc##1{\textcolor[rgb]{0.25,0.50,0.50}{##1}}}
\def\PY@tok@kc{\let\PY@bf=\textbf\def\PY@tc##1{\textcolor[rgb]{0.00,0.50,0.00}{##1}}}
\def\PY@tok@c{\let\PY@it=\textit\def\PY@tc##1{\textcolor[rgb]{0.25,0.50,0.50}{##1}}}
\def\PY@tok@mf{\def\PY@tc##1{\textcolor[rgb]{0.40,0.40,0.40}{##1}}}
\def\PY@tok@err{\def\PY@bc##1{\fcolorbox[rgb]{1.00,0.00,0.00}{1,1,1}{##1}}}
\def\PY@tok@kd{\let\PY@bf=\textbf\def\PY@tc##1{\textcolor[rgb]{0.00,0.50,0.00}{##1}}}
\def\PY@tok@ss{\def\PY@tc##1{\textcolor[rgb]{0.10,0.09,0.49}{##1}}}
\def\PY@tok@sr{\def\PY@tc##1{\textcolor[rgb]{0.73,0.40,0.53}{##1}}}
\def\PY@tok@mo{\def\PY@tc##1{\textcolor[rgb]{0.40,0.40,0.40}{##1}}}
\def\PY@tok@kn{\let\PY@bf=\textbf\def\PY@tc##1{\textcolor[rgb]{0.00,0.50,0.00}{##1}}}
\def\PY@tok@mi{\def\PY@tc##1{\textcolor[rgb]{0.40,0.40,0.40}{##1}}}
\def\PY@tok@gp{\let\PY@bf=\textbf\def\PY@tc##1{\textcolor[rgb]{0.00,0.00,0.50}{##1}}}
\def\PY@tok@o{\def\PY@tc##1{\textcolor[rgb]{0.40,0.40,0.40}{##1}}}
\def\PY@tok@kr{\let\PY@bf=\textbf\def\PY@tc##1{\textcolor[rgb]{0.00,0.50,0.00}{##1}}}
\def\PY@tok@s{\def\PY@tc##1{\textcolor[rgb]{0.73,0.13,0.13}{##1}}}
\def\PY@tok@kp{\def\PY@tc##1{\textcolor[rgb]{0.00,0.50,0.00}{##1}}}
\def\PY@tok@w{\def\PY@tc##1{\textcolor[rgb]{0.73,0.73,0.73}{##1}}}
\def\PY@tok@kt{\def\PY@tc##1{\textcolor[rgb]{0.69,0.00,0.25}{##1}}}
\def\PY@tok@ow{\let\PY@bf=\textbf\def\PY@tc##1{\textcolor[rgb]{0.67,0.13,1.00}{##1}}}
\def\PY@tok@sb{\def\PY@tc##1{\textcolor[rgb]{0.73,0.13,0.13}{##1}}}
\def\PY@tok@k{\let\PY@bf=\textbf\def\PY@tc##1{\textcolor[rgb]{0.00,0.50,0.00}{##1}}}
\def\PY@tok@se{\let\PY@bf=\textbf\def\PY@tc##1{\textcolor[rgb]{0.73,0.40,0.13}{##1}}}
\def\PY@tok@sd{\let\PY@it=\textit\def\PY@tc##1{\textcolor[rgb]{0.73,0.13,0.13}{##1}}}

\def\PYZbs{\char`\\}
\def\PYZus{\char`\_}
\def\PYZob{\char`\{}
\def\PYZcb{\char`\}}
\def\PYZca{\char`\^}
\def\PYZat{@}
\def\PYZlb{[}
\def\PYZrb{]}
\makeatother



In the following section mixtures of projected Gaussians are introduced.

\makeatletter
\def\PY@reset{\let\PY@it=\relax \let\PY@bf=\relax%
    \let\PY@ul=\relax \let\PY@tc=\relax%
    \let\PY@bc=\relax \let\PY@ff=\relax}
\def\PY@tok#1{\csname PY@tok@#1\endcsname}
\def\PY@toks#1+{\ifx\relax#1\empty\else%
    \PY@tok{#1}\expandafter\PY@toks\fi}
\def\PY@do#1{\PY@bc{\PY@tc{\PY@ul{%
    \PY@it{\PY@bf{\PY@ff{#1}}}}}}}
\def\PY#1#2{\PY@reset\PY@toks#1+\relax+\PY@do{#2}}

\def\PY@tok@gd{\def\PY@tc##1{\textcolor[rgb]{0.63,0.00,0.00}{##1}}}
\def\PY@tok@gu{\let\PY@bf=\textbf\def\PY@tc##1{\textcolor[rgb]{0.50,0.00,0.50}{##1}}}
\def\PY@tok@gt{\def\PY@tc##1{\textcolor[rgb]{0.00,0.25,0.82}{##1}}}
\def\PY@tok@gs{\let\PY@bf=\textbf}
\def\PY@tok@gr{\def\PY@tc##1{\textcolor[rgb]{1.00,0.00,0.00}{##1}}}
\def\PY@tok@cm{\let\PY@it=\textit\def\PY@tc##1{\textcolor[rgb]{0.25,0.50,0.50}{##1}}}
\def\PY@tok@vg{\def\PY@tc##1{\textcolor[rgb]{0.10,0.09,0.49}{##1}}}
\def\PY@tok@m{\def\PY@tc##1{\textcolor[rgb]{0.40,0.40,0.40}{##1}}}
\def\PY@tok@mh{\def\PY@tc##1{\textcolor[rgb]{0.40,0.40,0.40}{##1}}}
\def\PY@tok@go{\def\PY@tc##1{\textcolor[rgb]{0.50,0.50,0.50}{##1}}}
\def\PY@tok@ge{\let\PY@it=\textit}
\def\PY@tok@vc{\def\PY@tc##1{\textcolor[rgb]{0.10,0.09,0.49}{##1}}}
\def\PY@tok@il{\def\PY@tc##1{\textcolor[rgb]{0.40,0.40,0.40}{##1}}}
\def\PY@tok@cs{\let\PY@it=\textit\def\PY@tc##1{\textcolor[rgb]{0.25,0.50,0.50}{##1}}}
\def\PY@tok@cp{\def\PY@tc##1{\textcolor[rgb]{0.74,0.48,0.00}{##1}}}
\def\PY@tok@gi{\def\PY@tc##1{\textcolor[rgb]{0.00,0.63,0.00}{##1}}}
\def\PY@tok@gh{\let\PY@bf=\textbf\def\PY@tc##1{\textcolor[rgb]{0.00,0.00,0.50}{##1}}}
\def\PY@tok@ni{\let\PY@bf=\textbf\def\PY@tc##1{\textcolor[rgb]{0.60,0.60,0.60}{##1}}}
\def\PY@tok@nl{\def\PY@tc##1{\textcolor[rgb]{0.63,0.63,0.00}{##1}}}
\def\PY@tok@nn{\let\PY@bf=\textbf\def\PY@tc##1{\textcolor[rgb]{0.00,0.00,1.00}{##1}}}
\def\PY@tok@no{\def\PY@tc##1{\textcolor[rgb]{0.53,0.00,0.00}{##1}}}
\def\PY@tok@na{\def\PY@tc##1{\textcolor[rgb]{0.49,0.56,0.16}{##1}}}
\def\PY@tok@nb{\def\PY@tc##1{\textcolor[rgb]{0.00,0.50,0.00}{##1}}}
\def\PY@tok@nc{\let\PY@bf=\textbf\def\PY@tc##1{\textcolor[rgb]{0.00,0.00,1.00}{##1}}}
\def\PY@tok@nd{\def\PY@tc##1{\textcolor[rgb]{0.67,0.13,1.00}{##1}}}
\def\PY@tok@ne{\let\PY@bf=\textbf\def\PY@tc##1{\textcolor[rgb]{0.82,0.25,0.23}{##1}}}
\def\PY@tok@nf{\def\PY@tc##1{\textcolor[rgb]{0.00,0.00,1.00}{##1}}}
\def\PY@tok@si{\let\PY@bf=\textbf\def\PY@tc##1{\textcolor[rgb]{0.73,0.40,0.53}{##1}}}
\def\PY@tok@s2{\def\PY@tc##1{\textcolor[rgb]{0.73,0.13,0.13}{##1}}}
\def\PY@tok@vi{\def\PY@tc##1{\textcolor[rgb]{0.10,0.09,0.49}{##1}}}
\def\PY@tok@nt{\let\PY@bf=\textbf\def\PY@tc##1{\textcolor[rgb]{0.00,0.50,0.00}{##1}}}
\def\PY@tok@nv{\def\PY@tc##1{\textcolor[rgb]{0.10,0.09,0.49}{##1}}}
\def\PY@tok@s1{\def\PY@tc##1{\textcolor[rgb]{0.73,0.13,0.13}{##1}}}
\def\PY@tok@sh{\def\PY@tc##1{\textcolor[rgb]{0.73,0.13,0.13}{##1}}}
\def\PY@tok@sc{\def\PY@tc##1{\textcolor[rgb]{0.73,0.13,0.13}{##1}}}
\def\PY@tok@sx{\def\PY@tc##1{\textcolor[rgb]{0.00,0.50,0.00}{##1}}}
\def\PY@tok@bp{\def\PY@tc##1{\textcolor[rgb]{0.00,0.50,0.00}{##1}}}
\def\PY@tok@c1{\let\PY@it=\textit\def\PY@tc##1{\textcolor[rgb]{0.25,0.50,0.50}{##1}}}
\def\PY@tok@kc{\let\PY@bf=\textbf\def\PY@tc##1{\textcolor[rgb]{0.00,0.50,0.00}{##1}}}
\def\PY@tok@c{\let\PY@it=\textit\def\PY@tc##1{\textcolor[rgb]{0.25,0.50,0.50}{##1}}}
\def\PY@tok@mf{\def\PY@tc##1{\textcolor[rgb]{0.40,0.40,0.40}{##1}}}
\def\PY@tok@err{\def\PY@bc##1{\fcolorbox[rgb]{1.00,0.00,0.00}{1,1,1}{##1}}}
\def\PY@tok@kd{\let\PY@bf=\textbf\def\PY@tc##1{\textcolor[rgb]{0.00,0.50,0.00}{##1}}}
\def\PY@tok@ss{\def\PY@tc##1{\textcolor[rgb]{0.10,0.09,0.49}{##1}}}
\def\PY@tok@sr{\def\PY@tc##1{\textcolor[rgb]{0.73,0.40,0.53}{##1}}}
\def\PY@tok@mo{\def\PY@tc##1{\textcolor[rgb]{0.40,0.40,0.40}{##1}}}
\def\PY@tok@kn{\let\PY@bf=\textbf\def\PY@tc##1{\textcolor[rgb]{0.00,0.50,0.00}{##1}}}
\def\PY@tok@mi{\def\PY@tc##1{\textcolor[rgb]{0.40,0.40,0.40}{##1}}}
\def\PY@tok@gp{\let\PY@bf=\textbf\def\PY@tc##1{\textcolor[rgb]{0.00,0.00,0.50}{##1}}}
\def\PY@tok@o{\def\PY@tc##1{\textcolor[rgb]{0.40,0.40,0.40}{##1}}}
\def\PY@tok@kr{\let\PY@bf=\textbf\def\PY@tc##1{\textcolor[rgb]{0.00,0.50,0.00}{##1}}}
\def\PY@tok@s{\def\PY@tc##1{\textcolor[rgb]{0.73,0.13,0.13}{##1}}}
\def\PY@tok@kp{\def\PY@tc##1{\textcolor[rgb]{0.00,0.50,0.00}{##1}}}
\def\PY@tok@w{\def\PY@tc##1{\textcolor[rgb]{0.73,0.73,0.73}{##1}}}
\def\PY@tok@kt{\def\PY@tc##1{\textcolor[rgb]{0.69,0.00,0.25}{##1}}}
\def\PY@tok@ow{\let\PY@bf=\textbf\def\PY@tc##1{\textcolor[rgb]{0.67,0.13,1.00}{##1}}}
\def\PY@tok@sb{\def\PY@tc##1{\textcolor[rgb]{0.73,0.13,0.13}{##1}}}
\def\PY@tok@k{\let\PY@bf=\textbf\def\PY@tc##1{\textcolor[rgb]{0.00,0.50,0.00}{##1}}}
\def\PY@tok@se{\let\PY@bf=\textbf\def\PY@tc##1{\textcolor[rgb]{0.73,0.40,0.13}{##1}}}
\def\PY@tok@sd{\let\PY@it=\textit\def\PY@tc##1{\textcolor[rgb]{0.73,0.13,0.13}{##1}}}

\def\PYZbs{\char`\\}
\def\PYZus{\char`\_}
\def\PYZob{\char`\{}
\def\PYZcb{\char`\}}
\def\PYZca{\char`\^}
\def\PYZat{@}
\def\PYZlb{[}
\def\PYZrb{]}
\makeatother



From here on the quaternions and their methods are defined.

\makeatletter
\def\PY@reset{\let\PY@it=\relax \let\PY@bf=\relax%
    \let\PY@ul=\relax \let\PY@tc=\relax%
    \let\PY@bc=\relax \let\PY@ff=\relax}
\def\PY@tok#1{\csname PY@tok@#1\endcsname}
\def\PY@toks#1+{\ifx\relax#1\empty\else%
    \PY@tok{#1}\expandafter\PY@toks\fi}
\def\PY@do#1{\PY@bc{\PY@tc{\PY@ul{%
    \PY@it{\PY@bf{\PY@ff{#1}}}}}}}
\def\PY#1#2{\PY@reset\PY@toks#1+\relax+\PY@do{#2}}

\def\PY@tok@gd{\def\PY@tc##1{\textcolor[rgb]{0.63,0.00,0.00}{##1}}}
\def\PY@tok@gu{\let\PY@bf=\textbf\def\PY@tc##1{\textcolor[rgb]{0.50,0.00,0.50}{##1}}}
\def\PY@tok@gt{\def\PY@tc##1{\textcolor[rgb]{0.00,0.25,0.82}{##1}}}
\def\PY@tok@gs{\let\PY@bf=\textbf}
\def\PY@tok@gr{\def\PY@tc##1{\textcolor[rgb]{1.00,0.00,0.00}{##1}}}
\def\PY@tok@cm{\let\PY@it=\textit\def\PY@tc##1{\textcolor[rgb]{0.25,0.50,0.50}{##1}}}
\def\PY@tok@vg{\def\PY@tc##1{\textcolor[rgb]{0.10,0.09,0.49}{##1}}}
\def\PY@tok@m{\def\PY@tc##1{\textcolor[rgb]{0.40,0.40,0.40}{##1}}}
\def\PY@tok@mh{\def\PY@tc##1{\textcolor[rgb]{0.40,0.40,0.40}{##1}}}
\def\PY@tok@go{\def\PY@tc##1{\textcolor[rgb]{0.50,0.50,0.50}{##1}}}
\def\PY@tok@ge{\let\PY@it=\textit}
\def\PY@tok@vc{\def\PY@tc##1{\textcolor[rgb]{0.10,0.09,0.49}{##1}}}
\def\PY@tok@il{\def\PY@tc##1{\textcolor[rgb]{0.40,0.40,0.40}{##1}}}
\def\PY@tok@cs{\let\PY@it=\textit\def\PY@tc##1{\textcolor[rgb]{0.25,0.50,0.50}{##1}}}
\def\PY@tok@cp{\def\PY@tc##1{\textcolor[rgb]{0.74,0.48,0.00}{##1}}}
\def\PY@tok@gi{\def\PY@tc##1{\textcolor[rgb]{0.00,0.63,0.00}{##1}}}
\def\PY@tok@gh{\let\PY@bf=\textbf\def\PY@tc##1{\textcolor[rgb]{0.00,0.00,0.50}{##1}}}
\def\PY@tok@ni{\let\PY@bf=\textbf\def\PY@tc##1{\textcolor[rgb]{0.60,0.60,0.60}{##1}}}
\def\PY@tok@nl{\def\PY@tc##1{\textcolor[rgb]{0.63,0.63,0.00}{##1}}}
\def\PY@tok@nn{\let\PY@bf=\textbf\def\PY@tc##1{\textcolor[rgb]{0.00,0.00,1.00}{##1}}}
\def\PY@tok@no{\def\PY@tc##1{\textcolor[rgb]{0.53,0.00,0.00}{##1}}}
\def\PY@tok@na{\def\PY@tc##1{\textcolor[rgb]{0.49,0.56,0.16}{##1}}}
\def\PY@tok@nb{\def\PY@tc##1{\textcolor[rgb]{0.00,0.50,0.00}{##1}}}
\def\PY@tok@nc{\let\PY@bf=\textbf\def\PY@tc##1{\textcolor[rgb]{0.00,0.00,1.00}{##1}}}
\def\PY@tok@nd{\def\PY@tc##1{\textcolor[rgb]{0.67,0.13,1.00}{##1}}}
\def\PY@tok@ne{\let\PY@bf=\textbf\def\PY@tc##1{\textcolor[rgb]{0.82,0.25,0.23}{##1}}}
\def\PY@tok@nf{\def\PY@tc##1{\textcolor[rgb]{0.00,0.00,1.00}{##1}}}
\def\PY@tok@si{\let\PY@bf=\textbf\def\PY@tc##1{\textcolor[rgb]{0.73,0.40,0.53}{##1}}}
\def\PY@tok@s2{\def\PY@tc##1{\textcolor[rgb]{0.73,0.13,0.13}{##1}}}
\def\PY@tok@vi{\def\PY@tc##1{\textcolor[rgb]{0.10,0.09,0.49}{##1}}}
\def\PY@tok@nt{\let\PY@bf=\textbf\def\PY@tc##1{\textcolor[rgb]{0.00,0.50,0.00}{##1}}}
\def\PY@tok@nv{\def\PY@tc##1{\textcolor[rgb]{0.10,0.09,0.49}{##1}}}
\def\PY@tok@s1{\def\PY@tc##1{\textcolor[rgb]{0.73,0.13,0.13}{##1}}}
\def\PY@tok@sh{\def\PY@tc##1{\textcolor[rgb]{0.73,0.13,0.13}{##1}}}
\def\PY@tok@sc{\def\PY@tc##1{\textcolor[rgb]{0.73,0.13,0.13}{##1}}}
\def\PY@tok@sx{\def\PY@tc##1{\textcolor[rgb]{0.00,0.50,0.00}{##1}}}
\def\PY@tok@bp{\def\PY@tc##1{\textcolor[rgb]{0.00,0.50,0.00}{##1}}}
\def\PY@tok@c1{\let\PY@it=\textit\def\PY@tc##1{\textcolor[rgb]{0.25,0.50,0.50}{##1}}}
\def\PY@tok@kc{\let\PY@bf=\textbf\def\PY@tc##1{\textcolor[rgb]{0.00,0.50,0.00}{##1}}}
\def\PY@tok@c{\let\PY@it=\textit\def\PY@tc##1{\textcolor[rgb]{0.25,0.50,0.50}{##1}}}
\def\PY@tok@mf{\def\PY@tc##1{\textcolor[rgb]{0.40,0.40,0.40}{##1}}}
\def\PY@tok@err{\def\PY@bc##1{\fcolorbox[rgb]{1.00,0.00,0.00}{1,1,1}{##1}}}
\def\PY@tok@kd{\let\PY@bf=\textbf\def\PY@tc##1{\textcolor[rgb]{0.00,0.50,0.00}{##1}}}
\def\PY@tok@ss{\def\PY@tc##1{\textcolor[rgb]{0.10,0.09,0.49}{##1}}}
\def\PY@tok@sr{\def\PY@tc##1{\textcolor[rgb]{0.73,0.40,0.53}{##1}}}
\def\PY@tok@mo{\def\PY@tc##1{\textcolor[rgb]{0.40,0.40,0.40}{##1}}}
\def\PY@tok@kn{\let\PY@bf=\textbf\def\PY@tc##1{\textcolor[rgb]{0.00,0.50,0.00}{##1}}}
\def\PY@tok@mi{\def\PY@tc##1{\textcolor[rgb]{0.40,0.40,0.40}{##1}}}
\def\PY@tok@gp{\let\PY@bf=\textbf\def\PY@tc##1{\textcolor[rgb]{0.00,0.00,0.50}{##1}}}
\def\PY@tok@o{\def\PY@tc##1{\textcolor[rgb]{0.40,0.40,0.40}{##1}}}
\def\PY@tok@kr{\let\PY@bf=\textbf\def\PY@tc##1{\textcolor[rgb]{0.00,0.50,0.00}{##1}}}
\def\PY@tok@s{\def\PY@tc##1{\textcolor[rgb]{0.73,0.13,0.13}{##1}}}
\def\PY@tok@kp{\def\PY@tc##1{\textcolor[rgb]{0.00,0.50,0.00}{##1}}}
\def\PY@tok@w{\def\PY@tc##1{\textcolor[rgb]{0.73,0.73,0.73}{##1}}}
\def\PY@tok@kt{\def\PY@tc##1{\textcolor[rgb]{0.69,0.00,0.25}{##1}}}
\def\PY@tok@ow{\let\PY@bf=\textbf\def\PY@tc##1{\textcolor[rgb]{0.67,0.13,1.00}{##1}}}
\def\PY@tok@sb{\def\PY@tc##1{\textcolor[rgb]{0.73,0.13,0.13}{##1}}}
\def\PY@tok@k{\let\PY@bf=\textbf\def\PY@tc##1{\textcolor[rgb]{0.00,0.50,0.00}{##1}}}
\def\PY@tok@se{\let\PY@bf=\textbf\def\PY@tc##1{\textcolor[rgb]{0.73,0.40,0.13}{##1}}}
\def\PY@tok@sd{\let\PY@it=\textit\def\PY@tc##1{\textcolor[rgb]{0.73,0.13,0.13}{##1}}}

\def\PYZbs{\char`\\}
\def\PYZus{\char`\_}
\def\PYZob{\char`\{}
\def\PYZcb{\char`\}}
\def\PYZca{\char`\^}
\def\PYZat{@}
\def\PYZlb{[}
\def\PYZrb{]}
\makeatother

\begin{Verbatim}[commandchars=\\\{\}, fontsize = \footnotesize]
\PY{k}{class} \PY{n+nc}{quaternion}\PY{p}{:}
    \PY{l+s+sd}{""" definition of the quaternion and its properties}
\PY{l+s+sd}{    """}
    \PY{k}{def} \PY{n+nf}{\PYZus{}\PYZus{}init\PYZus{}\PYZus{}}\PY{p}{(}\PY{n+nb+bp}{self}\PY{p}{,} \PY{n}{a}\PY{p}{,} \PY{n}{b}\PY{p}{,} \PY{n}{c}\PY{p}{,} \PY{n}{d}\PY{p}{)}\PY{p}{:}
        \PY{n+nb+bp}{self}\PY{o}{.}\PY{n}{R} \PY{o}{=} \PY{n+nb}{float}\PY{p}{(}\PY{n}{a}\PY{p}{)}  \PY{c}{# real part}
        \PY{n+nb+bp}{self}\PY{o}{.}\PY{n}{I} \PY{o}{=} \PY{n+nb}{float}\PY{p}{(}\PY{n}{b}\PY{p}{)}  \PY{c}{# first imaginary part}
        \PY{n+nb+bp}{self}\PY{o}{.}\PY{n}{J} \PY{o}{=} \PY{n+nb}{float}\PY{p}{(}\PY{n}{c}\PY{p}{)}  \PY{c}{# second imaginary part}
        \PY{n+nb+bp}{self}\PY{o}{.}\PY{n}{K} \PY{o}{=} \PY{n+nb}{float}\PY{p}{(}\PY{n}{d}\PY{p}{)}  \PY{c}{# third imaginary part}

    \PY{k}{def} \PY{n+nf}{norm}\PY{p}{(}\PY{n+nb+bp}{self}\PY{p}{)}\PY{p}{:}
        \PY{k}{return} \PY{n}{sqrt}\PY{p}{(}\PY{n+nb+bp}{self}\PY{o}{.}\PY{n}{R}\PY{o}{*}\PY{n+nb+bp}{self}\PY{o}{.}\PY{n}{R} \PY{o}{+} \PY{n+nb+bp}{self}\PY{o}{.}\PY{n}{I}\PY{o}{*}\PY{n+nb+bp}{self}\PY{o}{.}\PY{n}{I} \PY{o}{+} \PY{n+nb+bp}{self}\PY{o}{.}\PY{n}{J}\PY{o}{*}\PY{n+nb+bp}{self}\PY{o}{.}\PY{n}{J} \PY{o}{+}
                    \PY{n+nb+bp}{self}\PY{o}{.}\PY{n}{K}\PY{o}{*}\PY{n+nb+bp}{self}\PY{o}{.}\PY{n}{K}\PY{p}{)}

    \PY{k}{def} \PY{n+nf}{norm2}\PY{p}{(}\PY{n+nb+bp}{self}\PY{p}{)}\PY{p}{:}
        \PY{k}{return} \PY{n+nb+bp}{self}\PY{o}{.}\PY{n}{R}\PY{o}{*}\PY{n+nb+bp}{self}\PY{o}{.}\PY{n}{R}\PY{o}{+}\PY{n+nb+bp}{self}\PY{o}{.}\PY{n}{I}\PY{o}{*}\PY{n+nb+bp}{self}\PY{o}{.}\PY{n}{I}\PY{o}{+}\PY{n+nb+bp}{self}\PY{o}{.}\PY{n}{J}\PY{o}{*}\PY{n+nb+bp}{self}\PY{o}{.}\PY{n}{J}\PY{o}{+}\PY{n+nb+bp}{self}\PY{o}{.}\PY{n}{K}\PY{o}{*}\PY{n+nb+bp}{self}\PY{o}{.}\PY{n}{K}

    \PY{k}{def} \PY{n+nf}{normalizeD}\PY{p}{(}\PY{n+nb+bp}{self}\PY{p}{)}\PY{p}{:} \PY{c}{# destructively normalizes the quaternion}
        \PY{n}{n}      \PY{o}{=} \PY{n+nb+bp}{self}\PY{o}{.}\PY{n}{norm}\PY{p}{(}\PY{p}{)}
        \PY{k}{if} \PY{n}{n} \PY{o}{==} \PY{l+m+mi}{0}\PY{p}{:}
            \PY{k}{print} \PY{l+s}{'}\PY{l+s}{cant normalize, is 0}\PY{l+s}{'}
        \PY{k}{else}\PY{p}{:}
            \PY{n+nb+bp}{self}\PY{o}{.}\PY{n}{R} \PY{o}{=} \PY{n+nb+bp}{self}\PY{o}{.}\PY{n}{R}\PY{o}{/}\PY{n}{n}  \PY{c}{# real part}
            \PY{n+nb+bp}{self}\PY{o}{.}\PY{n}{I} \PY{o}{=} \PY{n+nb+bp}{self}\PY{o}{.}\PY{n}{I}\PY{o}{/}\PY{n}{n}  \PY{c}{# first imaginary part}
            \PY{n+nb+bp}{self}\PY{o}{.}\PY{n}{J} \PY{o}{=} \PY{n+nb+bp}{self}\PY{o}{.}\PY{n}{J}\PY{o}{/}\PY{n}{n}  \PY{c}{# second imaginary part}
            \PY{n+nb+bp}{self}\PY{o}{.}\PY{n}{K} \PY{o}{=} \PY{n+nb+bp}{self}\PY{o}{.}\PY{n}{K}\PY{o}{/}\PY{n}{n}  \PY{c}{# third imaginary part}

    \PY{k}{def} \PY{n+nf}{normalize}\PY{p}{(}\PY{n+nb+bp}{self}\PY{p}{)}\PY{p}{:}
        \PY{c}{# non-destructively returns the normalized quaternion}
        \PY{n}{n}      \PY{o}{=} \PY{n+nb+bp}{self}\PY{o}{.}\PY{n}{norm}\PY{p}{(}\PY{p}{)}
        \PY{n}{qret}   \PY{o}{=} \PY{n}{quaternion}\PY{p}{(}\PY{n+nb+bp}{self}\PY{o}{.}\PY{n}{R}\PY{p}{,} \PY{n+nb+bp}{self}\PY{o}{.}\PY{n}{I}\PY{p}{,} \PY{n+nb+bp}{self}\PY{o}{.}\PY{n}{J}\PY{p}{,} \PY{n+nb+bp}{self}\PY{o}{.}\PY{n}{K}\PY{p}{)}
        \PY{k}{if} \PY{n}{n} \PY{o}{==} \PY{l+m+mi}{0}\PY{p}{:}
            \PY{k}{print} \PY{l+s}{'}\PY{l+s}{cant normalize, is 0}\PY{l+s}{'}
        \PY{k}{else}\PY{p}{:}
            \PY{n}{qret}\PY{o}{.}\PY{n}{R} \PY{o}{=} \PY{n+nb+bp}{self}\PY{o}{.}\PY{n}{R}\PY{o}{/}\PY{n}{n}  \PY{c}{# real part}
            \PY{n}{qret}\PY{o}{.}\PY{n}{I} \PY{o}{=} \PY{n+nb+bp}{self}\PY{o}{.}\PY{n}{I}\PY{o}{/}\PY{n}{n}  \PY{c}{# first imaginary part}
            \PY{n}{qret}\PY{o}{.}\PY{n}{J} \PY{o}{=} \PY{n+nb+bp}{self}\PY{o}{.}\PY{n}{J}\PY{o}{/}\PY{n}{n}  \PY{c}{# second imaginary part}
            \PY{n}{qret}\PY{o}{.}\PY{n}{K} \PY{o}{=} \PY{n+nb+bp}{self}\PY{o}{.}\PY{n}{K}\PY{o}{/}\PY{n}{n}  \PY{c}{# third imaginary part}
        \PY{k}{return} \PY{n}{qret}

    \PY{k}{def} \PY{n+nf}{conjD}\PY{p}{(}\PY{n+nb+bp}{self}\PY{p}{)}\PY{p}{:}
        \PY{c}{# destructively turns the quaternion into its conjugate}
        \PY{n+nb+bp}{self}\PY{o}{.}\PY{n}{R} \PY{o}{=}   \PY{n+nb+bp}{self}\PY{o}{.}\PY{n}{R}  \PY{c}{# real part}
        \PY{n+nb+bp}{self}\PY{o}{.}\PY{n}{I} \PY{o}{=} \PY{o}{-} \PY{n+nb+bp}{self}\PY{o}{.}\PY{n}{I}  \PY{c}{# first imaginary part}
        \PY{n+nb+bp}{self}\PY{o}{.}\PY{n}{J} \PY{o}{=} \PY{o}{-} \PY{n+nb+bp}{self}\PY{o}{.}\PY{n}{J}  \PY{c}{# second imaginary part}
        \PY{n+nb+bp}{self}\PY{o}{.}\PY{n}{K} \PY{o}{=} \PY{o}{-} \PY{n+nb+bp}{self}\PY{o}{.}\PY{n}{K}  \PY{c}{# third imaginary part}
        \PY{k}{return} \PY{n+nb+bp}{self}

    \PY{k}{def} \PY{n+nf}{conj}\PY{p}{(}\PY{n+nb+bp}{self}\PY{p}{)}\PY{p}{:} \PY{c}{# non-destructively returns the conjugate quaternion}
        \PY{n}{qret}   \PY{o}{=} \PY{n}{quaternion}\PY{p}{(}\PY{n+nb+bp}{self}\PY{o}{.}\PY{n}{R}\PY{p}{,} \PY{n+nb+bp}{self}\PY{o}{.}\PY{n}{I}\PY{p}{,} \PY{n+nb+bp}{self}\PY{o}{.}\PY{n}{J}\PY{p}{,} \PY{n+nb+bp}{self}\PY{o}{.}\PY{n}{K}\PY{p}{)}
        \PY{n}{qret}\PY{o}{.}\PY{n}{R} \PY{o}{=}   \PY{n+nb+bp}{self}\PY{o}{.}\PY{n}{R}  \PY{c}{# real part}
        \PY{n}{qret}\PY{o}{.}\PY{n}{I} \PY{o}{=} \PY{o}{-} \PY{n+nb+bp}{self}\PY{o}{.}\PY{n}{I}  \PY{c}{# first imaginary part}
        \PY{n}{qret}\PY{o}{.}\PY{n}{J} \PY{o}{=} \PY{o}{-} \PY{n+nb+bp}{self}\PY{o}{.}\PY{n}{J}  \PY{c}{# second imaginary part}
        \PY{n}{qret}\PY{o}{.}\PY{n}{K} \PY{o}{=} \PY{o}{-} \PY{n+nb+bp}{self}\PY{o}{.}\PY{n}{K}  \PY{c}{# third imaginary part}
        \PY{k}{return} \PY{n}{qret}

    \PY{k}{def} \PY{n+nf}{times}\PY{p}{(}\PY{n+nb+bp}{self}\PY{p}{,}\PY{n}{factor}\PY{p}{)}\PY{p}{:} \PY{c}{# multiplication with another quaternion}
        \PY{n}{qret}   \PY{o}{=} \PY{n}{quaternion}\PY{p}{(}\PY{l+m+mi}{0}\PY{p}{,} \PY{l+m+mi}{0}\PY{p}{,} \PY{l+m+mi}{0}\PY{p}{,} \PY{l+m+mi}{0}\PY{p}{)}
        \PY{n}{qret}\PY{o}{.}\PY{n}{R} \PY{o}{=} \PY{n+nb+bp}{self}\PY{o}{.}\PY{n}{R}\PY{o}{*}\PY{n}{factor}\PY{o}{.}\PY{n}{R} \PY{o}{-} \PY{n+nb+bp}{self}\PY{o}{.}\PY{n}{I}\PY{o}{*}\PY{n}{factor}\PY{o}{.}\PY{n}{I} \PY{o}{-} \PY{n+nb+bp}{self}\PY{o}{.}\PY{n}{J}\PY{o}{*}\PY{n}{factor}\PY{o}{.}\PY{n}{J}
                 \PY{o}{-} \PY{n+nb+bp}{self}\PY{o}{.}\PY{n}{K}\PY{o}{*}\PY{n}{factor}\PY{o}{.}\PY{n}{K}
        \PY{n}{qret}\PY{o}{.}\PY{n}{I} \PY{o}{=} \PY{n+nb+bp}{self}\PY{o}{.}\PY{n}{I}\PY{o}{*}\PY{n}{factor}\PY{o}{.}\PY{n}{R} \PY{o}{+} \PY{n+nb+bp}{self}\PY{o}{.}\PY{n}{R}\PY{o}{*}\PY{n}{factor}\PY{o}{.}\PY{n}{I} \PY{o}{-} \PY{n+nb+bp}{self}\PY{o}{.}\PY{n}{K}\PY{o}{*}\PY{n}{factor}\PY{o}{.}\PY{n}{J}
                 \PY{o}{+} \PY{n+nb+bp}{self}\PY{o}{.}\PY{n}{J}\PY{o}{*}\PY{n}{factor}\PY{o}{.}\PY{n}{K}
        \PY{n}{qret}\PY{o}{.}\PY{n}{J} \PY{o}{=} \PY{n+nb+bp}{self}\PY{o}{.}\PY{n}{J}\PY{o}{*}\PY{n}{factor}\PY{o}{.}\PY{n}{R} \PY{o}{+} \PY{n+nb+bp}{self}\PY{o}{.}\PY{n}{K}\PY{o}{*}\PY{n}{factor}\PY{o}{.}\PY{n}{I} \PY{o}{+} \PY{n+nb+bp}{self}\PY{o}{.}\PY{n}{R}\PY{o}{*}\PY{n}{factor}\PY{o}{.}\PY{n}{J}
                 \PY{o}{-} \PY{n+nb+bp}{self}\PY{o}{.}\PY{n}{I}\PY{o}{*}\PY{n}{factor}\PY{o}{.}\PY{n}{K}
        \PY{n}{qret}\PY{o}{.}\PY{n}{K} \PY{o}{=} \PY{n+nb+bp}{self}\PY{o}{.}\PY{n}{K}\PY{o}{*}\PY{n}{factor}\PY{o}{.}\PY{n}{R} \PY{o}{-} \PY{n+nb+bp}{self}\PY{o}{.}\PY{n}{J}\PY{o}{*}\PY{n}{factor}\PY{o}{.}\PY{n}{I} \PY{o}{+} \PY{n+nb+bp}{self}\PY{o}{.}\PY{n}{I}\PY{o}{*}\PY{n}{factor}\PY{o}{.}\PY{n}{J}
                 \PY{o}{+} \PY{n+nb+bp}{self}\PY{o}{.}\PY{n}{R}\PY{o}{*}\PY{n}{factor}\PY{o}{.}\PY{n}{K}  \PY{c}{# third imaginary part}
        \PY{k}{return} \PY{n}{qret}

    \PY{k}{def} \PY{n+nf}{toList}\PY{p}{(}\PY{n+nb+bp}{self}\PY{p}{)}\PY{p}{:}
        \PY{k}{return}\PY{p}{(}\PY{p}{[}\PY{n+nb+bp}{self}\PY{o}{.}\PY{n}{R}\PY{p}{,} \PY{n+nb+bp}{self}\PY{o}{.}\PY{n}{I}\PY{p}{,} \PY{n+nb+bp}{self}\PY{o}{.}\PY{n}{J}\PY{p}{,} \PY{n+nb+bp}{self}\PY{o}{.}\PY{n}{K}\PY{p}{]}\PY{p}{)}

    \PY{k}{def} \PY{n+nf}{invD}\PY{p}{(}\PY{n+nb+bp}{self}\PY{p}{)}\PY{p}{:}
        \PY{n}{n2} \PY{o}{=} \PY{n+nb+bp}{self}\PY{o}{.}\PY{n}{norm2}\PY{p}{(}\PY{p}{)}
        \PY{k}{if} \PY{n}{n2} \PY{o}{==} \PY{l+m+mi}{0}\PY{p}{:}
            \PY{k}{print} \PY{l+s}{'}\PY{l+s}{cant invert, is 0}\PY{l+s}{'}
        \PY{k}{else}\PY{p}{:}
            \PY{n+nb+bp}{self}\PY{o}{.}\PY{n}{R} \PY{o}{=}   \PY{n+nb+bp}{self}\PY{o}{.}\PY{n}{R}\PY{o}{/}\PY{n}{n2}  \PY{c}{# real part}
            \PY{n+nb+bp}{self}\PY{o}{.}\PY{n}{I} \PY{o}{=} \PY{o}{-} \PY{n+nb+bp}{self}\PY{o}{.}\PY{n}{I}\PY{o}{/}\PY{n}{n2}  \PY{c}{# first imaginary part}
            \PY{n+nb+bp}{self}\PY{o}{.}\PY{n}{J} \PY{o}{=} \PY{o}{-} \PY{n+nb+bp}{self}\PY{o}{.}\PY{n}{J}\PY{o}{/}\PY{n}{n2}  \PY{c}{# second imaginary part}
            \PY{n+nb+bp}{self}\PY{o}{.}\PY{n}{K} \PY{o}{=} \PY{o}{-} \PY{n+nb+bp}{self}\PY{o}{.}\PY{n}{K}\PY{o}{/}\PY{n}{n2}  \PY{c}{# third imaginary part}

    \PY{k}{def} \PY{n+nf}{inv}\PY{p}{(}\PY{n+nb+bp}{self}\PY{p}{)}\PY{p}{:}
        \PY{n}{qret}   \PY{o}{=} \PY{n}{quaternion}\PY{p}{(}\PY{l+m+mi}{0}\PY{p}{,}\PY{l+m+mi}{0}\PY{p}{,}\PY{l+m+mi}{0}\PY{p}{,}\PY{l+m+mi}{0}\PY{p}{)}
        \PY{n}{n2}     \PY{o}{=} \PY{n+nb+bp}{self}\PY{o}{.}\PY{n}{norm2}\PY{p}{(}\PY{p}{)}
        \PY{k}{if} \PY{n}{n2} \PY{o}{==} \PY{l+m+mi}{0}\PY{p}{:}
            \PY{k}{print} \PY{l+s}{'}\PY{l+s}{cant invert, is 0}\PY{l+s}{'}
        \PY{k}{else}\PY{p}{:}
            \PY{n}{qret}\PY{o}{.}\PY{n}{R} \PY{o}{=}   \PY{n+nb+bp}{self}\PY{o}{.}\PY{n}{R}\PY{o}{/}\PY{n}{n2}  \PY{c}{# real part}
            \PY{n}{qret}\PY{o}{.}\PY{n}{I} \PY{o}{=} \PY{o}{-} \PY{n+nb+bp}{self}\PY{o}{.}\PY{n}{I}\PY{o}{/}\PY{n}{n2}  \PY{c}{# first imaginary part}
            \PY{n}{qret}\PY{o}{.}\PY{n}{J} \PY{o}{=} \PY{o}{-} \PY{n+nb+bp}{self}\PY{o}{.}\PY{n}{J}\PY{o}{/}\PY{n}{n2}  \PY{c}{# second imaginary part}
            \PY{n}{qret}\PY{o}{.}\PY{n}{K} \PY{o}{=} \PY{o}{-} \PY{n+nb+bp}{self}\PY{o}{.}\PY{n}{K}\PY{o}{/}\PY{n}{n2}  \PY{c}{# third imaginary part}
            \PY{c}{#print qret}
            \PY{k}{return} \PY{n}{qret}

    \PY{k}{def} \PY{n+nf}{equal}\PY{p}{(}\PY{n+nb+bp}{self}\PY{p}{,}\PY{n}{same}\PY{p}{)}\PY{p}{:} \PY{c}{#tests whether is equal to another quaternion}
        \PY{k}{if} \PY{p}{(}\PY{n+nb+bp}{self}\PY{o}{.}\PY{n}{R} \PY{o}{==} \PY{n}{same}\PY{o}{.}\PY{n}{R}\PY{p}{)} \PY{o}{&} \PY{p}{(}\PY{n+nb+bp}{self}\PY{o}{.}\PY{n}{I} \PY{o}{==} \PY{n}{same}\PY{o}{.}\PY{n}{I}\PY{p}{)} \PY{o}{&} \PY{p}{(}\PY{n+nb+bp}{self}\PY{o}{.}\PY{n}{J} \PY{o}{==} \PY{n}{same}\PY{o}{.}\PY{n}{J}\PY{p}{)}
           \PY{o}{&} \PY{p}{(}\PY{n+nb+bp}{self}\PY{o}{.}\PY{n}{K} \PY{o}{==} \PY{n}{same}\PY{o}{.}\PY{n}{K}\PY{p}{)}\PY{p}{:}
            \PY{k}{return} \PY{n+nb+bp}{True}
        \PY{k}{else}\PY{p}{:}
            \PY{k}{return} \PY{n+nb+bp}{False}

    \PY{k}{def} \PY{n+nf}{copy}\PY{p}{(}\PY{n+nb+bp}{self}\PY{p}{)}\PY{p}{:}
        \PY{n}{qret}   \PY{o}{=} \PY{n}{quaternion}\PY{p}{(}\PY{l+m+mi}{0}\PY{p}{,}\PY{l+m+mi}{0}\PY{p}{,}\PY{l+m+mi}{0}\PY{p}{,}\PY{l+m+mi}{0}\PY{p}{)}
        \PY{n}{qret}\PY{o}{.}\PY{n}{R} \PY{o}{=}   \PY{n+nb+bp}{self}\PY{o}{.}\PY{n}{R}  \PY{c}{# real part}
        \PY{n}{qret}\PY{o}{.}\PY{n}{I} \PY{o}{=}   \PY{n+nb+bp}{self}\PY{o}{.}\PY{n}{I}  \PY{c}{# first imaginary part}
        \PY{n}{qret}\PY{o}{.}\PY{n}{J} \PY{o}{=}   \PY{n+nb+bp}{self}\PY{o}{.}\PY{n}{J}  \PY{c}{# second imaginary part}
        \PY{n}{qret}\PY{o}{.}\PY{n}{K} \PY{o}{=}   \PY{n+nb+bp}{self}\PY{o}{.}\PY{n}{K}  \PY{c}{# third imaginary part}
        \PY{k}{return} \PY{n}{qret}

    \PY{k}{def} \PY{n+nf}{plus}\PY{p}{(}\PY{n+nb+bp}{self}\PY{p}{,}\PY{n}{otherq}\PY{p}{)}\PY{p}{:}
        \PY{n}{qret} \PY{o}{=} \PY{n}{quaternion}\PY{p}{(}\PY{l+m+mi}{0}\PY{p}{,}\PY{l+m+mi}{0}\PY{p}{,}\PY{l+m+mi}{0}\PY{p}{,}\PY{l+m+mi}{0}\PY{p}{)}
        \PY{n}{qret}\PY{o}{.}\PY{n}{R} \PY{o}{=} \PY{n+nb+bp}{self}\PY{o}{.}\PY{n}{R} \PY{o}{+} \PY{n}{otherq}\PY{o}{.}\PY{n}{R}
        \PY{n}{qret}\PY{o}{.}\PY{n}{I} \PY{o}{=} \PY{n+nb+bp}{self}\PY{o}{.}\PY{n}{I} \PY{o}{+} \PY{n}{otherq}\PY{o}{.}\PY{n}{I}
        \PY{n}{qret}\PY{o}{.}\PY{n}{J} \PY{o}{=} \PY{n+nb+bp}{self}\PY{o}{.}\PY{n}{J} \PY{o}{+} \PY{n}{otherq}\PY{o}{.}\PY{n}{J}
        \PY{n}{qret}\PY{o}{.}\PY{n}{K} \PY{o}{=} \PY{n+nb+bp}{self}\PY{o}{.}\PY{n}{K} \PY{o}{+} \PY{n}{otherq}\PY{o}{.}\PY{n}{K}
        \PY{k}{return} \PY{n}{qret}

    \PY{k}{def} \PY{n+nf}{scalar}\PY{p}{(}\PY{n+nb+bp}{self}\PY{p}{,}\PY{n}{num}\PY{p}{)}\PY{p}{:}
        \PY{n}{qret} \PY{o}{=} \PY{n}{quaternion}\PY{p}{(}\PY{l+m+mi}{0}\PY{p}{,}\PY{l+m+mi}{0}\PY{p}{,}\PY{l+m+mi}{0}\PY{p}{,}\PY{l+m+mi}{0}\PY{p}{)}
        \PY{n}{qret}\PY{o}{.}\PY{n}{R} \PY{o}{=} \PY{n}{num}\PY{o}{*}\PY{n+nb+bp}{self}\PY{o}{.}\PY{n}{R}
        \PY{n}{qret}\PY{o}{.}\PY{n}{I} \PY{o}{=} \PY{n}{num}\PY{o}{*}\PY{n+nb+bp}{self}\PY{o}{.}\PY{n}{I}
        \PY{n}{qret}\PY{o}{.}\PY{n}{J} \PY{o}{=} \PY{n}{num}\PY{o}{*}\PY{n+nb+bp}{self}\PY{o}{.}\PY{n}{J}
        \PY{n}{qret}\PY{o}{.}\PY{n}{K} \PY{o}{=} \PY{n}{num}\PY{o}{*}\PY{n+nb+bp}{self}\PY{o}{.}\PY{n}{K}
        \PY{k}{return} \PY{n}{qret}

    \PY{k}{def} \PY{n+nf}{toMatrix}\PY{p}{(}\PY{n+nb+bp}{self}\PY{p}{)}\PY{p}{:}
        \PY{n}{a} \PY{o}{=} \PY{n+nb+bp}{self}\PY{o}{.}\PY{n}{R}
        \PY{n}{b} \PY{o}{=} \PY{n+nb+bp}{self}\PY{o}{.}\PY{n}{I}
        \PY{n}{c} \PY{o}{=} \PY{n+nb+bp}{self}\PY{o}{.}\PY{n}{J}
        \PY{n}{d} \PY{o}{=} \PY{n+nb+bp}{self}\PY{o}{.}\PY{n}{K}
        \PY{n}{mat}\PY{o}{=}\PY{n}{matrix}\PY{p}{(}\PY{p}{[}\PY{p}{[}\PY{l+m+mi}{1} \PY{o}{-} \PY{l+m+mi}{2}\PY{o}{*}\PY{n}{c}\PY{o}{*}\PY{n}{c} \PY{o}{-} \PY{l+m+mi}{2}\PY{o}{*}\PY{n}{d}\PY{o}{*}\PY{n}{d}\PY{p}{,} \PY{l+m+mi}{2}\PY{o}{*}\PY{n}{b}\PY{o}{*}\PY{n}{c} \PY{o}{-} \PY{l+m+mi}{2}\PY{o}{*}\PY{n}{a}\PY{o}{*}\PY{n}{d}\PY{p}{,} \PY{l+m+mi}{2}\PY{o}{*}\PY{p}{(}\PY{n}{a}\PY{o}{*}\PY{n}{c} \PY{o}{+} \PY{n}{b}\PY{o}{*}\PY{n}{d}\PY{p}{)}\PY{p}{]}\PY{p}{,} \PYZbs{}
                    \PY{p}{[}\PY{l+m+mi}{2}\PY{o}{*}\PY{p}{(}\PY{n}{b}\PY{o}{*}\PY{n}{c} \PY{o}{+} \PY{n}{a}\PY{o}{*}\PY{n}{d}\PY{p}{)}\PY{p}{,} \PY{l+m+mi}{1} \PY{o}{-} \PY{l+m+mi}{2}\PY{o}{*}\PY{n}{b}\PY{o}{*}\PY{n}{b} \PY{o}{-} \PY{l+m+mi}{2}\PY{o}{*}\PY{n}{d}\PY{o}{*}\PY{n}{d}\PY{p}{,} \PY{o}{-}\PY{l+m+mi}{2}\PY{o}{*}\PY{n}{a}\PY{o}{*}\PY{n}{b} \PY{o}{+} \PY{l+m+mi}{2}\PY{o}{*}\PY{n}{c}\PY{o}{*}\PY{n}{d}\PY{p}{]}\PY{p}{,}\PYZbs{}
                    \PY{p}{[}\PY{o}{-}\PY{l+m+mi}{2}\PY{o}{*}\PY{n}{a}\PY{o}{*}\PY{n}{c} \PY{o}{+} \PY{l+m+mi}{2}\PY{o}{*}\PY{n}{b}\PY{o}{*}\PY{n}{d}\PY{p}{,} \PY{l+m+mi}{2}\PY{o}{*}\PY{p}{(}\PY{n}{a}\PY{o}{*}\PY{n}{b} \PY{o}{+} \PY{n}{c}\PY{o}{*}\PY{n}{d}\PY{p}{)}\PY{p}{,} \PY{l+m+mi}{1} \PY{o}{-} \PY{l+m+mi}{2}\PY{o}{*}\PY{n}{b}\PY{o}{*}\PY{n}{b} \PY{o}{-} \PY{l+m+mi}{2}\PY{o}{*}\PY{n}{c}\PY{o}{*}\PY{n}{c}\PY{p}{]}\PY{p}{]}\PY{p}{)}
        \PY{k}{return} \PY{n}{mat}

    \PY{k}{def} \PY{n+nf}{randomUnit}\PY{p}{(}\PY{n+nb+bp}{self}\PY{p}{)}\PY{p}{:}
        \PY{l+s+sd}{"""this method returns a random unit quaternion from an equal}
\PY{l+s+sd}{        distribution on S3"""}
        \PY{n}{qret} \PY{o}{=} \PY{n}{quaternion}\PY{p}{(}\PY{l+m+mi}{1}\PY{p}{,}\PY{l+m+mi}{1}\PY{p}{,}\PY{l+m+mi}{1}\PY{p}{,}\PY{l+m+mi}{1}\PY{p}{)}
        \PY{k}{while} \PY{n}{qret}\PY{o}{.}\PY{n}{norm}\PY{p}{(}\PY{p}{)} \PY{o}{>} \PY{l+m+mf}{1.0}\PY{p}{:}
            \PY{n}{qret}\PY{o}{.}\PY{n}{R} \PY{o}{=} \PY{n}{random}\PY{o}{.}\PY{n}{uniform}\PY{p}{(}\PY{o}{-}\PY{l+m+mf}{1.0}\PY{p}{,} \PY{l+m+mf}{1.0}\PY{p}{)}
            \PY{n}{qret}\PY{o}{.}\PY{n}{I} \PY{o}{=} \PY{n}{random}\PY{o}{.}\PY{n}{uniform}\PY{p}{(}\PY{o}{-}\PY{l+m+mf}{1.0}\PY{p}{,} \PY{l+m+mf}{1.0}\PY{p}{)}
            \PY{n}{qret}\PY{o}{.}\PY{n}{J} \PY{o}{=} \PY{n}{random}\PY{o}{.}\PY{n}{uniform}\PY{p}{(}\PY{o}{-}\PY{l+m+mf}{1.0}\PY{p}{,} \PY{l+m+mf}{1.0}\PY{p}{)}
            \PY{n}{qret}\PY{o}{.}\PY{n}{K} \PY{o}{=} \PY{n}{random}\PY{o}{.}\PY{n}{uniform}\PY{p}{(}\PY{o}{-}\PY{l+m+mf}{1.0}\PY{p}{,} \PY{l+m+mf}{1.0}\PY{p}{)}
        \PY{n}{qret} \PY{o}{=} \PY{n}{qret}\PY{o}{.}\PY{n}{normalize}\PY{p}{(}\PY{p}{)}
        \PY{k}{return} \PY{n}{qret}

    \PY{k}{def} \PY{n+nf}{randomImaginary}\PY{p}{(}\PY{n+nb+bp}{self}\PY{p}{,}\PY{n+nb}{min} \PY{o}{=} \PY{p}{[}\PY{o}{-}\PY{l+m+mi}{5}\PY{p}{,}\PY{o}{-}\PY{l+m+mi}{5}\PY{p}{,}\PY{o}{-}\PY{l+m+mi}{5}\PY{p}{]}\PY{p}{,}\PY{n+nb}{max} \PY{o}{=} \PY{p}{[}\PY{l+m+mi}{5}\PY{p}{,}\PY{l+m+mi}{5}\PY{p}{,}\PY{l+m+mi}{5}\PY{p}{]}\PY{p}{)}\PY{p}{:}
        \PY{l+s+sd}{"""this method returns a random imaginary quaternion from an}
\PY{l+s+sd}{        equal distribution on S3"""}
        \PY{n}{qret} \PY{o}{=} \PY{n}{quaternion}\PY{p}{(}\PY{l+m+mi}{0}\PY{p}{,}\PY{l+m+mi}{0}\PY{p}{,}\PY{l+m+mi}{0}\PY{p}{,}\PY{l+m+mi}{0}\PY{p}{)}
        \PY{n}{qret}\PY{o}{.}\PY{n}{I} \PY{o}{=} \PY{n}{random}\PY{o}{.}\PY{n}{uniform}\PY{p}{(}\PY{n+nb}{min}\PY{p}{[}\PY{l+m+mi}{0}\PY{p}{]}\PY{p}{,}\PY{n+nb}{max}\PY{p}{[}\PY{l+m+mi}{0}\PY{p}{]}\PY{p}{)}
        \PY{n}{qret}\PY{o}{.}\PY{n}{J} \PY{o}{=} \PY{n}{random}\PY{o}{.}\PY{n}{uniform}\PY{p}{(}\PY{n+nb}{min}\PY{p}{[}\PY{l+m+mi}{1}\PY{p}{]}\PY{p}{,}\PY{n+nb}{max}\PY{p}{[}\PY{l+m+mi}{1}\PY{p}{]}\PY{p}{)}
        \PY{n}{qret}\PY{o}{.}\PY{n}{K} \PY{o}{=} \PY{n}{random}\PY{o}{.}\PY{n}{uniform}\PY{p}{(}\PY{n+nb}{min}\PY{p}{[}\PY{l+m+mi}{2}\PY{p}{]}\PY{p}{,}\PY{n+nb}{max}\PY{p}{[}\PY{l+m+mi}{2}\PY{p}{]}\PY{p}{)}
        \PY{k}{return} \PY{n}{qret}

    \PY{k}{def} \PY{n+nf}{radAxis}\PY{p}{(}\PY{n+nb+bp}{self}\PY{p}{,} \PY{n}{rad}\PY{p}{,} \PY{n}{axis}\PY{p}{)}\PY{p}{:}
        \PY{l+s+sd}{"""}
\PY{l+s+sd}{        this method returns a unit quaternion corresponding to a}
\PY{l+s+sd}{        rotation by rad around axis}
\PY{l+s+sd}{        """}
        \PY{c}{# axis need not be normalized, we'll take care of that}
        \PY{n}{qret} \PY{o}{=} \PY{n}{quaternion}\PY{p}{(}\PY{l+m+mi}{1}\PY{p}{,}\PY{l+m+mi}{0}\PY{p}{,}\PY{l+m+mi}{0}\PY{p}{,}\PY{l+m+mi}{0}\PY{p}{)}
        \PY{n}{a1}   \PY{o}{=} \PY{n+nb}{float}\PY{p}{(}\PY{n}{axis}\PY{p}{[}\PY{l+m+mi}{0}\PY{p}{]}\PY{p}{)}
        \PY{n}{a2}   \PY{o}{=} \PY{n+nb}{float}\PY{p}{(}\PY{n}{axis}\PY{p}{[}\PY{l+m+mi}{1}\PY{p}{]}\PY{p}{)}
        \PY{n}{a3}   \PY{o}{=} \PY{n+nb}{float}\PY{p}{(}\PY{n}{axis}\PY{p}{[}\PY{l+m+mi}{2}\PY{p}{]}\PY{p}{)}
        \PY{n}{axisnorm} \PY{o}{=} \PY{n}{sqrt}\PY{p}{(}\PY{n}{a1}\PY{o}{*}\PY{n}{a1} \PY{o}{+} \PY{n}{a2}\PY{o}{*}\PY{n}{a2} \PY{o}{+} \PY{n}{a3}\PY{o}{*}\PY{n}{a3}\PY{p}{)}
        \PY{k}{if} \PY{n}{axisnorm} \PY{o}{>} \PY{l+m+mi}{0}\PY{p}{:}
            \PY{n}{qret}\PY{o}{.}\PY{n}{R} \PY{o}{=} \PY{n}{cos}\PY{p}{(}\PY{n}{rad}\PY{o}{/}\PY{l+m+mf}{2.}\PY{p}{)}
            \PY{n}{sr2}    \PY{o}{=} \PY{n}{sin}\PY{p}{(}\PY{n}{rad}\PY{o}{/}\PY{l+m+mf}{2.}\PY{p}{)}\PY{o}{/}\PY{n}{axisnorm}
            \PY{n}{qret}\PY{o}{.}\PY{n}{I} \PY{o}{=} \PY{n}{sr2}\PY{o}{*}\PY{n}{a1}
            \PY{n}{qret}\PY{o}{.}\PY{n}{J} \PY{o}{=} \PY{n}{sr2}\PY{o}{*}\PY{n}{a2}
            \PY{n}{qret}\PY{o}{.}\PY{n}{K} \PY{o}{=} \PY{n}{sr2}\PY{o}{*}\PY{n}{a3}
        \PY{k}{return} \PY{n}{qret}

    \PY{k}{def} \PY{n+nf}{degreeAxis}\PY{p}{(}\PY{n+nb+bp}{self}\PY{p}{,} \PY{n}{deg}\PY{p}{,} \PY{n}{axis}\PY{p}{)}\PY{p}{:}
        \PY{l+s+sd}{"""this method returns a unit quaternion corresponding to a}
\PY{l+s+sd}{        rotation by deg around axis"""}
        \PY{c}{# axis need not be normalized, we'll take care of that}
        \PY{n}{rad} \PY{o}{=} \PY{n}{deg} \PY{o}{*} \PY{n}{pi}\PY{o}{/}\PY{l+m+mf}{180.}
        \PY{n}{qret} \PY{o}{=} \PY{n+nb+bp}{self}\PY{o}{.}\PY{n}{radAxis}\PY{p}{(}\PY{n}{rad}\PY{p}{,} \PY{n}{axis}\PY{p}{)}
        \PY{k}{return} \PY{n}{qret}

    \PY{k}{def} \PY{n+nf}{display}\PY{p}{(}\PY{n+nb+bp}{self}\PY{p}{)}\PY{p}{:}
        \PY{k}{print} \PY{p}{[}\PY{n+nb+bp}{self}\PY{o}{.}\PY{n}{R}\PY{p}{,} \PY{n+nb+bp}{self}\PY{o}{.}\PY{n}{I}\PY{p}{,} \PY{n+nb+bp}{self}\PY{o}{.}\PY{n}{J}\PY{p}{,} \PY{n+nb+bp}{self}\PY{o}{.}\PY{n}{K}\PY{p}{]}
\end{Verbatim}

The final section is concerned with dual quaternions.

\makeatletter
\def\PY@reset{\let\PY@it=\relax \let\PY@bf=\relax%
    \let\PY@ul=\relax \let\PY@tc=\relax%
    \let\PY@bc=\relax \let\PY@ff=\relax}
\def\PY@tok#1{\csname PY@tok@#1\endcsname}
\def\PY@toks#1+{\ifx\relax#1\empty\else%
    \PY@tok{#1}\expandafter\PY@toks\fi}
\def\PY@do#1{\PY@bc{\PY@tc{\PY@ul{%
    \PY@it{\PY@bf{\PY@ff{#1}}}}}}}
\def\PY#1#2{\PY@reset\PY@toks#1+\relax+\PY@do{#2}}

\def\PY@tok@gd{\def\PY@tc##1{\textcolor[rgb]{0.63,0.00,0.00}{##1}}}
\def\PY@tok@gu{\let\PY@bf=\textbf\def\PY@tc##1{\textcolor[rgb]{0.50,0.00,0.50}{##1}}}
\def\PY@tok@gt{\def\PY@tc##1{\textcolor[rgb]{0.00,0.25,0.82}{##1}}}
\def\PY@tok@gs{\let\PY@bf=\textbf}
\def\PY@tok@gr{\def\PY@tc##1{\textcolor[rgb]{1.00,0.00,0.00}{##1}}}
\def\PY@tok@cm{\let\PY@it=\textit\def\PY@tc##1{\textcolor[rgb]{0.25,0.50,0.50}{##1}}}
\def\PY@tok@vg{\def\PY@tc##1{\textcolor[rgb]{0.10,0.09,0.49}{##1}}}
\def\PY@tok@m{\def\PY@tc##1{\textcolor[rgb]{0.40,0.40,0.40}{##1}}}
\def\PY@tok@mh{\def\PY@tc##1{\textcolor[rgb]{0.40,0.40,0.40}{##1}}}
\def\PY@tok@go{\def\PY@tc##1{\textcolor[rgb]{0.50,0.50,0.50}{##1}}}
\def\PY@tok@ge{\let\PY@it=\textit}
\def\PY@tok@vc{\def\PY@tc##1{\textcolor[rgb]{0.10,0.09,0.49}{##1}}}
\def\PY@tok@il{\def\PY@tc##1{\textcolor[rgb]{0.40,0.40,0.40}{##1}}}
\def\PY@tok@cs{\let\PY@it=\textit\def\PY@tc##1{\textcolor[rgb]{0.25,0.50,0.50}{##1}}}
\def\PY@tok@cp{\def\PY@tc##1{\textcolor[rgb]{0.74,0.48,0.00}{##1}}}
\def\PY@tok@gi{\def\PY@tc##1{\textcolor[rgb]{0.00,0.63,0.00}{##1}}}
\def\PY@tok@gh{\let\PY@bf=\textbf\def\PY@tc##1{\textcolor[rgb]{0.00,0.00,0.50}{##1}}}
\def\PY@tok@ni{\let\PY@bf=\textbf\def\PY@tc##1{\textcolor[rgb]{0.60,0.60,0.60}{##1}}}
\def\PY@tok@nl{\def\PY@tc##1{\textcolor[rgb]{0.63,0.63,0.00}{##1}}}
\def\PY@tok@nn{\let\PY@bf=\textbf\def\PY@tc##1{\textcolor[rgb]{0.00,0.00,1.00}{##1}}}
\def\PY@tok@no{\def\PY@tc##1{\textcolor[rgb]{0.53,0.00,0.00}{##1}}}
\def\PY@tok@na{\def\PY@tc##1{\textcolor[rgb]{0.49,0.56,0.16}{##1}}}
\def\PY@tok@nb{\def\PY@tc##1{\textcolor[rgb]{0.00,0.50,0.00}{##1}}}
\def\PY@tok@nc{\let\PY@bf=\textbf\def\PY@tc##1{\textcolor[rgb]{0.00,0.00,1.00}{##1}}}
\def\PY@tok@nd{\def\PY@tc##1{\textcolor[rgb]{0.67,0.13,1.00}{##1}}}
\def\PY@tok@ne{\let\PY@bf=\textbf\def\PY@tc##1{\textcolor[rgb]{0.82,0.25,0.23}{##1}}}
\def\PY@tok@nf{\def\PY@tc##1{\textcolor[rgb]{0.00,0.00,1.00}{##1}}}
\def\PY@tok@si{\let\PY@bf=\textbf\def\PY@tc##1{\textcolor[rgb]{0.73,0.40,0.53}{##1}}}
\def\PY@tok@s2{\def\PY@tc##1{\textcolor[rgb]{0.73,0.13,0.13}{##1}}}
\def\PY@tok@vi{\def\PY@tc##1{\textcolor[rgb]{0.10,0.09,0.49}{##1}}}
\def\PY@tok@nt{\let\PY@bf=\textbf\def\PY@tc##1{\textcolor[rgb]{0.00,0.50,0.00}{##1}}}
\def\PY@tok@nv{\def\PY@tc##1{\textcolor[rgb]{0.10,0.09,0.49}{##1}}}
\def\PY@tok@s1{\def\PY@tc##1{\textcolor[rgb]{0.73,0.13,0.13}{##1}}}
\def\PY@tok@sh{\def\PY@tc##1{\textcolor[rgb]{0.73,0.13,0.13}{##1}}}
\def\PY@tok@sc{\def\PY@tc##1{\textcolor[rgb]{0.73,0.13,0.13}{##1}}}
\def\PY@tok@sx{\def\PY@tc##1{\textcolor[rgb]{0.00,0.50,0.00}{##1}}}
\def\PY@tok@bp{\def\PY@tc##1{\textcolor[rgb]{0.00,0.50,0.00}{##1}}}
\def\PY@tok@c1{\let\PY@it=\textit\def\PY@tc##1{\textcolor[rgb]{0.25,0.50,0.50}{##1}}}
\def\PY@tok@kc{\let\PY@bf=\textbf\def\PY@tc##1{\textcolor[rgb]{0.00,0.50,0.00}{##1}}}
\def\PY@tok@c{\let\PY@it=\textit\def\PY@tc##1{\textcolor[rgb]{0.25,0.50,0.50}{##1}}}
\def\PY@tok@mf{\def\PY@tc##1{\textcolor[rgb]{0.40,0.40,0.40}{##1}}}
\def\PY@tok@err{\def\PY@bc##1{\fcolorbox[rgb]{1.00,0.00,0.00}{1,1,1}{##1}}}
\def\PY@tok@kd{\let\PY@bf=\textbf\def\PY@tc##1{\textcolor[rgb]{0.00,0.50,0.00}{##1}}}
\def\PY@tok@ss{\def\PY@tc##1{\textcolor[rgb]{0.10,0.09,0.49}{##1}}}
\def\PY@tok@sr{\def\PY@tc##1{\textcolor[rgb]{0.73,0.40,0.53}{##1}}}
\def\PY@tok@mo{\def\PY@tc##1{\textcolor[rgb]{0.40,0.40,0.40}{##1}}}
\def\PY@tok@kn{\let\PY@bf=\textbf\def\PY@tc##1{\textcolor[rgb]{0.00,0.50,0.00}{##1}}}
\def\PY@tok@mi{\def\PY@tc##1{\textcolor[rgb]{0.40,0.40,0.40}{##1}}}
\def\PY@tok@gp{\let\PY@bf=\textbf\def\PY@tc##1{\textcolor[rgb]{0.00,0.00,0.50}{##1}}}
\def\PY@tok@o{\def\PY@tc##1{\textcolor[rgb]{0.40,0.40,0.40}{##1}}}
\def\PY@tok@kr{\let\PY@bf=\textbf\def\PY@tc##1{\textcolor[rgb]{0.00,0.50,0.00}{##1}}}
\def\PY@tok@s{\def\PY@tc##1{\textcolor[rgb]{0.73,0.13,0.13}{##1}}}
\def\PY@tok@kp{\def\PY@tc##1{\textcolor[rgb]{0.00,0.50,0.00}{##1}}}
\def\PY@tok@w{\def\PY@tc##1{\textcolor[rgb]{0.73,0.73,0.73}{##1}}}
\def\PY@tok@kt{\def\PY@tc##1{\textcolor[rgb]{0.69,0.00,0.25}{##1}}}
\def\PY@tok@ow{\let\PY@bf=\textbf\def\PY@tc##1{\textcolor[rgb]{0.67,0.13,1.00}{##1}}}
\def\PY@tok@sb{\def\PY@tc##1{\textcolor[rgb]{0.73,0.13,0.13}{##1}}}
\def\PY@tok@k{\let\PY@bf=\textbf\def\PY@tc##1{\textcolor[rgb]{0.00,0.50,0.00}{##1}}}
\def\PY@tok@se{\let\PY@bf=\textbf\def\PY@tc##1{\textcolor[rgb]{0.73,0.40,0.13}{##1}}}
\def\PY@tok@sd{\let\PY@it=\textit\def\PY@tc##1{\textcolor[rgb]{0.73,0.13,0.13}{##1}}}

\def\PYZbs{\char`\\}
\def\PYZus{\char`\_}
\def\PYZob{\char`\{}
\def\PYZcb{\char`\}}
\def\PYZca{\char`\^}
\def\PYZat{@}
\def\PYZlb{[}
\def\PYZrb{]}
\makeatother

\begin{Verbatim}[commandchars=\\\{\}, fontsize = \footnotesize]
\PY{k}{class} \PY{n+nc}{dualQuaternion}\PY{p}{:}
    \PY{l+s+sd}{"""Quaternions have 4 entries. Quat = [q1,q2,q3,q4] the first one is}
\PY{l+s+sd}{    real, the others are imaginary."""}
    \PY{k}{def} \PY{n+nf}{\PYZus{}\PYZus{}init\PYZus{}\PYZus{}}\PY{p}{(}\PY{n+nb+bp}{self}\PY{p}{,}\PY{n}{Quat1}\PY{p}{,}\PY{n}{Quat2}\PY{p}{)}\PY{p}{:}
        \PY{k}{assert} \PY{n+nb}{isinstance}\PY{p}{(}\PY{n}{Quat1}\PY{p}{,}\PY{n}{quaternion}\PY{p}{)}\PY{p}{,} \PY{l+s}{'}\PY{l+s}{1st argument no quaterion}\PY{l+s}{'}
        \PY{k}{assert} \PY{n+nb}{isinstance}\PY{p}{(}\PY{n}{Quat2}\PY{p}{,}\PY{n}{quaternion}\PY{p}{)}\PY{p}{,} \PY{l+s}{'}\PY{l+s}{2nd argument no quaternion}\PY{l+s}{'}
        \PY{n+nb+bp}{self}\PY{o}{.}\PY{n}{Real} \PY{o}{=} \PY{n}{Quat1}\PY{o}{.}\PY{n}{copy}\PY{p}{(}\PY{p}{)} \PY{c}{#Realteil des dualen Quaternions}
        \PY{n+nb+bp}{self}\PY{o}{.}\PY{n}{Dual} \PY{o}{=} \PY{n}{Quat2}\PY{o}{.}\PY{n}{copy}\PY{p}{(}\PY{p}{)} \PY{c}{#Dualteil des dualen Quaternions}
        \PY{c}{#ein duales Quaternion hat dann die Form: Quat1 + E*Quat2}

    \PY{k}{def} \PY{n+nf}{plus}\PY{p}{(}\PY{n+nb+bp}{self}\PY{p}{,}\PY{n}{otherDuQu}\PY{p}{)}\PY{p}{:}
        \PY{n}{dqret} \PY{o}{=} \PY{n}{dualQuaternion}\PY{p}{(}\PY{n}{quaternion}\PY{p}{(}\PY{l+m+mi}{0}\PY{p}{,}\PY{l+m+mi}{0}\PY{p}{,}\PY{l+m+mi}{0}\PY{p}{,}\PY{l+m+mi}{0}\PY{p}{)}\PY{p}{,}\PY{n}{quaternion}\PY{p}{(}\PY{l+m+mi}{0}\PY{p}{,}\PY{l+m+mi}{0}\PY{p}{,}\PY{l+m+mi}{0}\PY{p}{,}\PY{l+m+mi}{0}\PY{p}{)}\PY{p}{)}
        \PY{n}{sR} \PY{o}{=} \PY{n+nb+bp}{self}\PY{o}{.}\PY{n}{Real}
        \PY{n}{oR} \PY{o}{=} \PY{n}{otherDuQu}\PY{o}{.}\PY{n}{Real}
        \PY{n}{sD} \PY{o}{=} \PY{n+nb+bp}{self}\PY{o}{.}\PY{n}{Dual}
        \PY{n}{oD} \PY{o}{=} \PY{n}{otherDuQu}\PY{o}{.}\PY{n}{Dual}
        \PY{n}{dqret}\PY{o}{.}\PY{n}{Real} \PY{o}{=} \PY{n}{sR}\PY{o}{.}\PY{n}{plus}\PY{p}{(}\PY{n}{oR}\PY{p}{)}
        \PY{n}{dqret}\PY{o}{.}\PY{n}{Dual} \PY{o}{=} \PY{n}{sD}\PY{o}{.}\PY{n}{plus}\PY{p}{(}\PY{n}{oD}\PY{p}{)}
        \PY{k}{return} \PY{n}{dqret}

    \PY{k}{def} \PY{n+nf}{equal}\PY{p}{(}\PY{n+nb+bp}{self}\PY{p}{,}\PY{n}{otherDuQu}\PY{p}{)}\PY{p}{:}
        \PY{c}{#tests whether is equal to another quaternion}
        \PY{n}{sR} \PY{o}{=} \PY{n+nb+bp}{self}\PY{o}{.}\PY{n}{Real}
        \PY{n}{oR} \PY{o}{=} \PY{n}{otherDuQu}\PY{o}{.}\PY{n}{Real}
        \PY{n}{sD} \PY{o}{=} \PY{n+nb+bp}{self}\PY{o}{.}\PY{n}{Dual}
        \PY{n}{oD} \PY{o}{=} \PY{n}{otherDuQu}\PY{o}{.}\PY{n}{Dual}
        \PY{k}{if} \PY{p}{(}\PY{n}{sR}\PY{o}{.}\PY{n}{equal}\PY{p}{(}\PY{n}{oR}\PY{p}{)}\PY{p}{)} \PY{o}{&} \PY{p}{(}\PY{n}{sD}\PY{o}{.}\PY{n}{equal}\PY{p}{(}\PY{n}{oD}\PY{p}{)}\PY{p}{)}\PY{p}{:}
            \PY{k}{return} \PY{n+nb+bp}{True}
        \PY{k}{else}\PY{p}{:}
            \PY{k}{return} \PY{n+nb+bp}{False}

    \PY{k}{def} \PY{n+nf}{copy}\PY{p}{(}\PY{n+nb+bp}{self}\PY{p}{)}\PY{p}{:}
        \PY{n}{dqret} \PY{o}{=} \PY{n}{dualQuaternion}\PY{p}{(}\PY{n}{quaternion}\PY{p}{(}\PY{l+m+mi}{0}\PY{p}{,}\PY{l+m+mi}{0}\PY{p}{,}\PY{l+m+mi}{0}\PY{p}{,}\PY{l+m+mi}{0}\PY{p}{)}\PY{p}{,}\PY{n}{quaternion}\PY{p}{(}\PY{l+m+mi}{0}\PY{p}{,}\PY{l+m+mi}{0}\PY{p}{,}\PY{l+m+mi}{0}\PY{p}{,}\PY{l+m+mi}{0}\PY{p}{)}\PY{p}{)}
        \PY{n}{dqret}\PY{o}{.}\PY{n}{Real} \PY{o}{=} \PY{n+nb+bp}{self}\PY{o}{.}\PY{n}{Real}  \PY{c}{# real part}
        \PY{n}{dqret}\PY{o}{.}\PY{n}{Dual} \PY{o}{=} \PY{n+nb+bp}{self}\PY{o}{.}\PY{n}{Dual}  \PY{c}{# dual part}
        \PY{k}{return} \PY{n}{dqret}

    \PY{k}{def} \PY{n+nf}{toList}\PY{p}{(}\PY{n+nb+bp}{self}\PY{p}{)}\PY{p}{:}
        \PY{n}{Q1} \PY{o}{=} \PY{n+nb+bp}{self}\PY{o}{.}\PY{n}{Real}
        \PY{n}{Q2} \PY{o}{=} \PY{n+nb+bp}{self}\PY{o}{.}\PY{n}{Dual}
        \PY{k}{return} \PY{p}{[}\PY{p}{[}\PY{n}{Q1}\PY{o}{.}\PY{n}{R}\PY{p}{,} \PY{n}{Q1}\PY{o}{.}\PY{n}{I}\PY{p}{,} \PY{n}{Q1}\PY{o}{.}\PY{n}{J}\PY{p}{,} \PY{n}{Q1}\PY{o}{.}\PY{n}{K}\PY{p}{]}\PY{p}{,}\PY{p}{[}\PY{n}{Q2}\PY{o}{.}\PY{n}{R}\PY{p}{,} \PY{n}{Q2}\PY{o}{.}\PY{n}{I}\PY{p}{,} \PY{n}{Q2}\PY{o}{.}\PY{n}{J}\PY{p}{,} \PY{n}{Q2}\PY{o}{.}\PY{n}{K}\PY{p}{]}\PY{p}{]}

    \PY{k}{def} \PY{n+nf}{times}\PY{p}{(}\PY{n+nb+bp}{self}\PY{p}{,}\PY{n}{otherDuQu}\PY{p}{)}\PY{p}{:} \PY{c}{# multiplication with another quaternion}
        \PY{n}{dqret} \PY{o}{=} \PY{n}{dualQuaternion}\PY{p}{(}\PY{n}{quaternion}\PY{p}{(}\PY{l+m+mi}{0}\PY{p}{,}\PY{l+m+mi}{0}\PY{p}{,}\PY{l+m+mi}{0}\PY{p}{,}\PY{l+m+mi}{0}\PY{p}{)}\PY{p}{,}\PY{n}{quaternion}\PY{p}{(}\PY{l+m+mi}{0}\PY{p}{,}\PY{l+m+mi}{0}\PY{p}{,}\PY{l+m+mi}{0}\PY{p}{,}\PY{l+m+mi}{0}\PY{p}{)}\PY{p}{)}
        \PY{n}{qR1} \PY{o}{=} \PY{n+nb+bp}{self}\PY{o}{.}\PY{n}{Real}
        \PY{n}{qR2} \PY{o}{=} \PY{n}{otherDuQu}\PY{o}{.}\PY{n}{Real}
        \PY{n}{qD1} \PY{o}{=} \PY{n+nb+bp}{self}\PY{o}{.}\PY{n}{Dual}
        \PY{n}{qD2} \PY{o}{=} \PY{n}{otherDuQu}\PY{o}{.}\PY{n}{Dual}
        \PY{n}{qu1} \PY{o}{=} \PY{n}{qR1}\PY{o}{.}\PY{n}{times}\PY{p}{(}\PY{n}{qD2}\PY{p}{)}
        \PY{n}{qu2} \PY{o}{=} \PY{n}{qD1}\PY{o}{.}\PY{n}{times}\PY{p}{(}\PY{n}{qR2}\PY{p}{)}
        \PY{n}{dqret}\PY{o}{.}\PY{n}{Real} \PY{o}{=} \PY{n}{qR1}\PY{o}{.}\PY{n}{times}\PY{p}{(}\PY{n}{qR2}\PY{p}{)}
        \PY{n}{dqret}\PY{o}{.}\PY{n}{Dual} \PY{o}{=} \PY{n}{qu1}\PY{o}{.}\PY{n}{plus}\PY{p}{(}\PY{n}{qu2}\PY{p}{)}
        \PY{k}{return} \PY{n}{dqret}

    \PY{k}{def} \PY{n+nf}{conjTotal}\PY{p}{(}\PY{n+nb+bp}{self}\PY{p}{)}\PY{p}{:}
        \PY{n}{dqret} \PY{o}{=} \PY{n}{dualQuaternion}\PY{p}{(}\PY{n}{quaternion}\PY{p}{(}\PY{l+m+mi}{0}\PY{p}{,}\PY{l+m+mi}{0}\PY{p}{,}\PY{l+m+mi}{0}\PY{p}{,}\PY{l+m+mi}{0}\PY{p}{)}\PY{p}{,}\PY{n}{quaternion}\PY{p}{(}\PY{l+m+mi}{0}\PY{p}{,}\PY{l+m+mi}{0}\PY{p}{,}\PY{l+m+mi}{0}\PY{p}{,}\PY{l+m+mi}{0}\PY{p}{)}\PY{p}{)}
        \PY{n}{sR} \PY{o}{=} \PY{n+nb+bp}{self}\PY{o}{.}\PY{n}{Real}
        \PY{n}{sD} \PY{o}{=} \PY{n+nb+bp}{self}\PY{o}{.}\PY{n}{Dual}
        \PY{n}{dqret}\PY{o}{.}\PY{n}{Real} \PY{o}{=} \PY{n}{sR}\PY{o}{.}\PY{n}{conj}\PY{p}{(}\PY{p}{)}
        \PY{n}{cD} \PY{o}{=} \PY{n}{sD}\PY{o}{.}\PY{n}{conj}\PY{p}{(}\PY{p}{)}
        \PY{n}{dqret}\PY{o}{.}\PY{n}{Dual} \PY{o}{=} \PY{n}{cD}\PY{o}{.}\PY{n}{scalar}\PY{p}{(}\PY{o}{-}\PY{l+m+mi}{1}\PY{p}{)}
        \PY{k}{return} \PY{n}{dqret}
    \PY{k}{def} \PY{n+nf}{conjQuat}\PY{p}{(}\PY{n+nb+bp}{self}\PY{p}{)}\PY{p}{:}
        \PY{n}{dqret} \PY{o}{=} \PY{n}{dualQuaternion}\PY{p}{(}\PY{n}{quaternion}\PY{p}{(}\PY{l+m+mi}{0}\PY{p}{,}\PY{l+m+mi}{0}\PY{p}{,}\PY{l+m+mi}{0}\PY{p}{,}\PY{l+m+mi}{0}\PY{p}{)}\PY{p}{,}\PY{n}{quaternion}\PY{p}{(}\PY{l+m+mi}{0}\PY{p}{,}\PY{l+m+mi}{0}\PY{p}{,}\PY{l+m+mi}{0}\PY{p}{,}\PY{l+m+mi}{0}\PY{p}{)}\PY{p}{)}
        \PY{n}{sR} \PY{o}{=} \PY{n+nb+bp}{self}\PY{o}{.}\PY{n}{Real}
        \PY{n}{sD} \PY{o}{=} \PY{n+nb+bp}{self}\PY{o}{.}\PY{n}{Dual}
        \PY{n}{dqret}\PY{o}{.}\PY{n}{Real} \PY{o}{=} \PY{n}{sR}\PY{o}{.}\PY{n}{conj}\PY{p}{(}\PY{p}{)}
        \PY{n}{dqret}\PY{o}{.}\PY{n}{Dual} \PY{o}{=} \PY{n}{sD}\PY{o}{.}\PY{n}{conj}\PY{p}{(}\PY{p}{)}
        \PY{k}{return} \PY{n}{dqret}
    \PY{k}{def} \PY{n+nf}{conjDual}\PY{p}{(}\PY{n+nb+bp}{self}\PY{p}{)}\PY{p}{:}
        \PY{n}{dqret} \PY{o}{=} \PY{n}{dualQuaternion}\PY{p}{(}\PY{n}{quaternion}\PY{p}{(}\PY{l+m+mi}{0}\PY{p}{,}\PY{l+m+mi}{0}\PY{p}{,}\PY{l+m+mi}{0}\PY{p}{,}\PY{l+m+mi}{0}\PY{p}{)}\PY{p}{,}\PY{n}{quaternion}\PY{p}{(}\PY{l+m+mi}{0}\PY{p}{,}\PY{l+m+mi}{0}\PY{p}{,}\PY{l+m+mi}{0}\PY{p}{,}\PY{l+m+mi}{0}\PY{p}{)}\PY{p}{)}
        \PY{n}{sR} \PY{o}{=} \PY{n+nb+bp}{self}\PY{o}{.}\PY{n}{Real}
        \PY{n}{du} \PY{o}{=} \PY{n+nb+bp}{self}\PY{o}{.}\PY{n}{Dual}
        \PY{n}{dqret}\PY{o}{.}\PY{n}{Real} \PY{o}{=} \PY{n}{sR}
        \PY{n}{dqret}\PY{o}{.}\PY{n}{Dual} \PY{o}{=} \PY{n}{du}\PY{o}{.}\PY{n}{scalar}\PY{p}{(}\PY{o}{-}\PY{l+m+mi}{1}\PY{p}{)}
        \PY{k}{return} \PY{n}{dqret}

    \PY{k}{def} \PY{n+nf}{transformationMatrix}\PY{p}{(}\PY{n+nb+bp}{self}\PY{p}{)}\PY{p}{:}
        \PY{n}{R} \PY{o}{=} \PY{n+nb+bp}{self}\PY{o}{.}\PY{n}{Real}
        \PY{n}{D} \PY{o}{=} \PY{n+nb+bp}{self}\PY{o}{.}\PY{n}{Dual}
        \PY{n}{rotmat} \PY{o}{=} \PY{n}{R}\PY{o}{.}\PY{n}{toMatrix}\PY{p}{(}\PY{p}{)} \PY{c}{#matrix}
        \PY{n}{DD} \PY{o}{=} \PY{n}{D}\PY{o}{.}\PY{n}{scalar}\PY{p}{(}\PY{l+m+mi}{2}\PY{p}{)}
        \PY{n}{DD} \PY{o}{=} \PY{n}{DD}\PY{o}{.}\PY{n}{times}\PY{p}{(}\PY{n}{R}\PY{o}{.}\PY{n}{conj}\PY{p}{(}\PY{p}{)}\PY{p}{)}
        \PY{n}{DDlist} \PY{o}{=} \PY{n}{DD}\PY{o}{.}\PY{n}{toList}\PY{p}{(}\PY{p}{)}
        \PY{n}{transvec} \PY{o}{=} \PY{n}{DDlist}\PY{p}{[}\PY{l+m+mi}{1}\PY{p}{:}\PY{p}{]}
        \PY{n}{transvec} \PY{o}{=} \PY{n}{makeVector}\PY{p}{(}\PY{n}{transvec}\PY{p}{)}
        \PY{c}{#print rotmat}
        \PY{c}{#print transvec}
        \PY{k}{return} \PY{p}{[}\PY{n}{rotmat}\PY{p}{,}\PY{n}{transvec}\PY{p}{]}

    \PY{k}{def} \PY{n+nf}{inv}\PY{p}{(}\PY{n+nb+bp}{self}\PY{p}{)}\PY{p}{:}
        \PY{n}{ret}     \PY{o}{=} \PY{n}{dualQuaternion}\PY{p}{(}\PY{n}{quaternion}\PY{p}{(}\PY{l+m+mi}{0}\PY{p}{,}\PY{l+m+mi}{0}\PY{p}{,}\PY{l+m+mi}{0}\PY{p}{,}\PY{l+m+mi}{0}\PY{p}{)}\PY{p}{,}\PY{n}{quaternion}\PY{p}{(}\PY{l+m+mi}{0}\PY{p}{,}\PY{l+m+mi}{0}\PY{p}{,}\PY{l+m+mi}{0}\PY{p}{,}\PY{l+m+mi}{0}\PY{p}{)}\PY{p}{)}
        \PY{n}{real}    \PY{o}{=} \PY{n+nb+bp}{self}\PY{o}{.}\PY{n}{Real}
        \PY{n}{realc}   \PY{o}{=} \PY{n}{real}\PY{o}{.}\PY{n}{conj}\PY{p}{(}\PY{p}{)}
        \PY{n}{ret}\PY{o}{.}\PY{n}{Real}\PY{o}{=} \PY{n}{realc}
        \PY{n}{dual}    \PY{o}{=} \PY{n+nb+bp}{self}\PY{o}{.}\PY{n}{Dual}
        \PY{n}{retd}    \PY{o}{=} \PY{n}{dual}\PY{o}{.}\PY{n}{times}\PY{p}{(}\PY{n}{realc}\PY{p}{)}
        \PY{n}{retd}    \PY{o}{=} \PY{n}{realc}\PY{o}{.}\PY{n}{times}\PY{p}{(}\PY{n}{retd}\PY{p}{)}
        \PY{n}{retd}    \PY{o}{=} \PY{n}{retd}\PY{o}{.}\PY{n}{scalar}\PY{p}{(}\PY{o}{-}\PY{l+m+mf}{1.0}\PY{p}{)}
        \PY{n}{ret}\PY{o}{.}\PY{n}{Dual}\PY{o}{=} \PY{n}{retd}
        \PY{k}{return} \PY{n}{ret}

    \PY{k}{def} \PY{n+nf}{display}\PY{p}{(}\PY{n+nb+bp}{self}\PY{p}{)}\PY{p}{:}
        \PY{k}{print} \PY{n+nb+bp}{self}\PY{o}{.}\PY{n}{Real}\PY{o}{.}\PY{n}{toList}\PY{p}{(}\PY{p}{)} \PY{o}{+} \PY{n+nb+bp}{self}\PY{o}{.}\PY{n}{Dual}\PY{o}{.}\PY{n}{toList}\PY{p}{(}\PY{p}{)}
\end{Verbatim}

\newpage
\section{}

\begin{figure}[bht]
\begin{center}
 \includegraphics[width=1\textwidth]{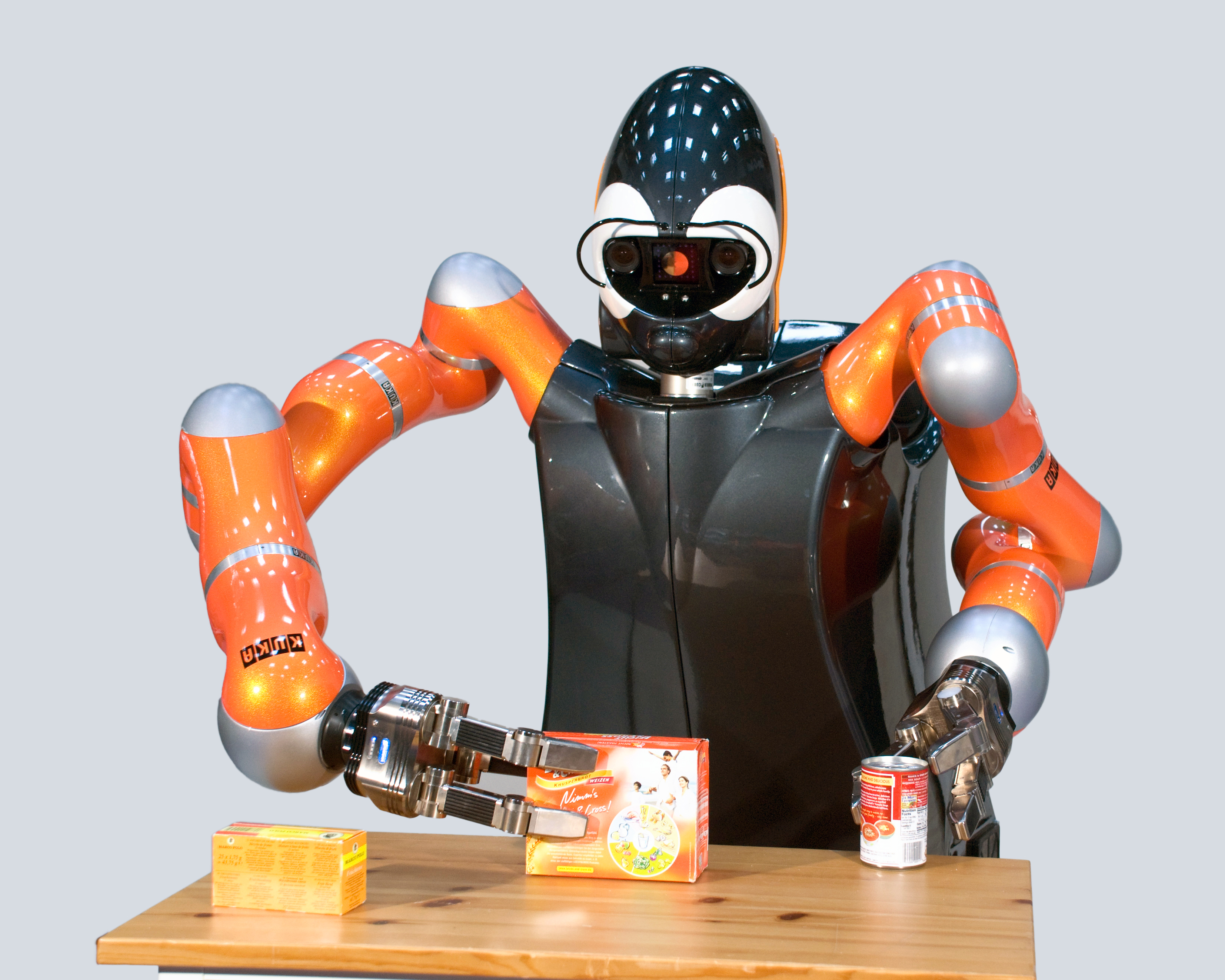}
 \end{center}
\caption[DESIRE robot grasping two objects]{The picture shows the DESIRE robot on picking up two objects of a simple scenario at the same time.}
 \label{desire3}
\end{figure}

\begin{figure}[bht]
\begin{center}
 \includegraphics[width=1\textwidth]{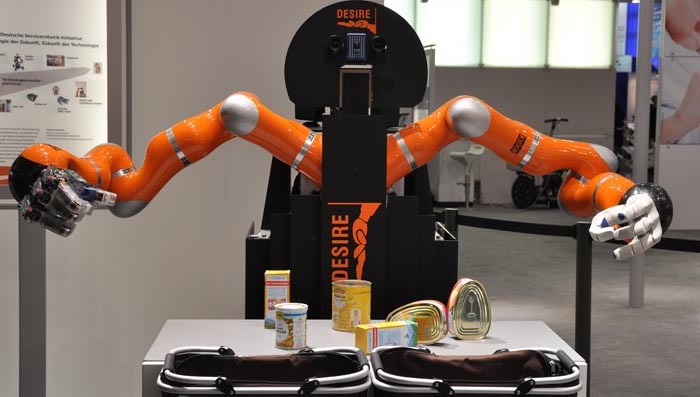}
 \end{center}
\caption[DESIRE robot]{The picture shows an older model of the DESIRE robot. This one was presented on the CeBIT 2009. }
 \label{desire4}
\end{figure}

\begin{figure}[bht]
\begin{center}
 \includegraphics[width=1\textwidth]{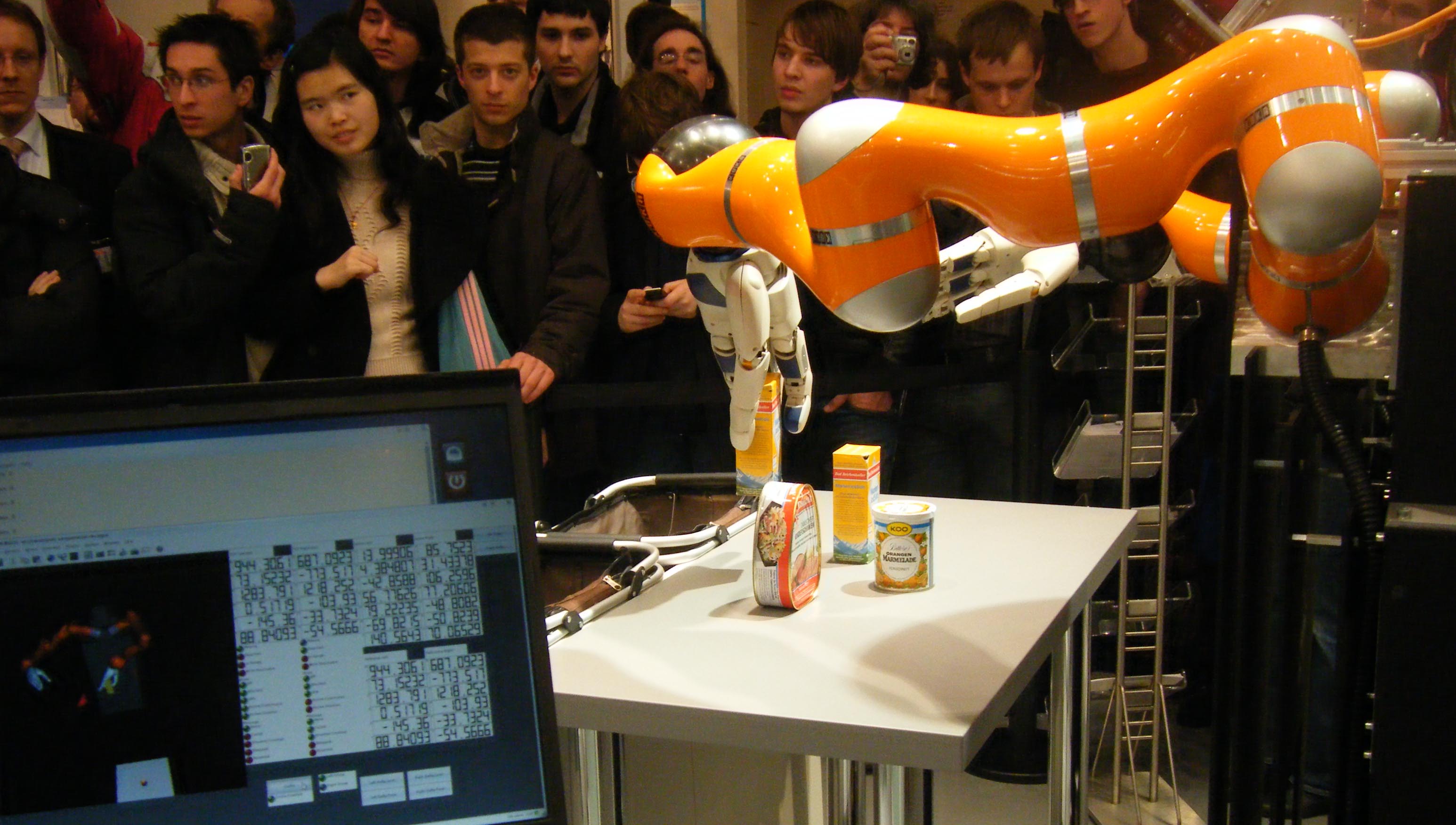}
 \end{center}
\caption[DESIRE robot sorting trash]{The picture shows the DESIRE robot sorting trash. The robot picks the objects up and throws them into one of the bags on the back side of the table. Which bag is chosen depends on whether the object is empty or not. }
 \label{desire5}
\end{figure}

\end{appendix}


\newpage
\bibliographystyle{plain}
\bibliography{references}

\newpage
\listoffigures
\addcontentsline{toc}{chapter}{Table of Contents}
\newpage
\renewcommand \thesection {\Alph{section}}
\setcounter{section}{1}

\newpage
\thispagestyle{empty}
\addcontentsline{toc}{chapter}{Declaration}
\vspace*{2cm}
\begin{center}{\large Declaration}\end{center}
\vspace{2cm}
\begin{quote}
Hiermit erkl\"{a}re ich ehrenw\"{o}rtlich, dass ich die hier vorliegende Diplomarbeit selbst\"{a}ndig verfasst und nur die ausgewiesenen Quellen verwendet habe.\\
\vspace{3cm}

\dots\dots\dots\dots\dots\dots\dots\dots\dots\\
\\
Muriel Lang, M\"{u}nchen, den 24.01.2011
\end{quote}

\end{document}